\documentclass[preprint,12pt,authoryear]{elsarticle}

\usepackage{amssymb}
\usepackage{amsmath}
\newcommand{\ar}{\text{AR}}
\newcommand{\argmax}{\operatornamewithlimits{argmax}}
\newcommand{\softmax}{\operatornamewithlimits{softmax}}

\usepackage{amsfonts}
\usepackage{booktabs}
\usepackage{pdfpages}
\usepackage{graphicx}
\usepackage{array}
\usepackage{float}
\usepackage{algorithm}
\usepackage{algpseudocode}
\usepackage{caption}
\usepackage{tablefootnote}
\usepackage{url}

\journal{Speech Communication}

\begin{document}

\begin{frontmatter}



\title{Towards Unsupervised Speech Recognition Without Pronunciation Models}


\author[uiucece]{Junrui Ni\corref{aaa}}\ead{junruin2@illinois.edu}
\author[mitcsail]{Liming Wang}
\author[mitibm]{Yang Zhang}
\author[mitibm]{Kaizhi Qian}
\author[uiucece]{Heting Gao}
\author[uiucece]{Mark Hasegawa-Johnson\corref{aaa}}\ead{jhasegaw@illinois.edu}
\author[kaistee]{Chang D. Yoo}

\cortext[aaa]{Corresponding Author}
\affiliation[uiucece]{organization={Department of Electrical and Computer Engineering, University of Illinois Urbana-Champaign},
            country={United States}}

\affiliation[mitcsail]{organization={Computer Science and Artificial Intelligence Laboratory, Massachusetts Institute of Technology (MIT)},
            country={United States}}

\affiliation[mitibm]{organization={MIT-IBM Watson AI Lab},
            country={United States}}

\affiliation[kaistee]{organization={Electrical Engineering, Korea Advanced Institute of Science and Technology (KAIST)},
            country={Republic of Korea}}

\begin{abstract}
Recent advancements in supervised automatic speech recognition (ASR) have achieved remarkable performance, largely due to the growing availability of large transcribed speech corpora. However, most languages lack sufficient paired speech and text data to effectively train these systems. In this article, we tackle the challenge of developing ASR systems without paired speech and text corpora by proposing the removal of reliance on a phoneme lexicon. We explore a new research direction: word-level unsupervised ASR, and experimentally demonstrate that an unsupervised speech recognizer can emerge from joint speech-to-speech and text-to-text masked token-infilling. Using a curated speech corpus containing a fixed number of English words, our system iteratively refines the word segmentation structure and achieves a word error rate of between 20-23\%, depending on the vocabulary size, without parallel transcripts, oracle word boundaries, or a pronunciation lexicon. This innovative model surpasses the performance of previous unsupervised ASR models under the lexicon-free setting.
\end{abstract}

\begin{graphicalabstract}
\includegraphics[width=\linewidth]{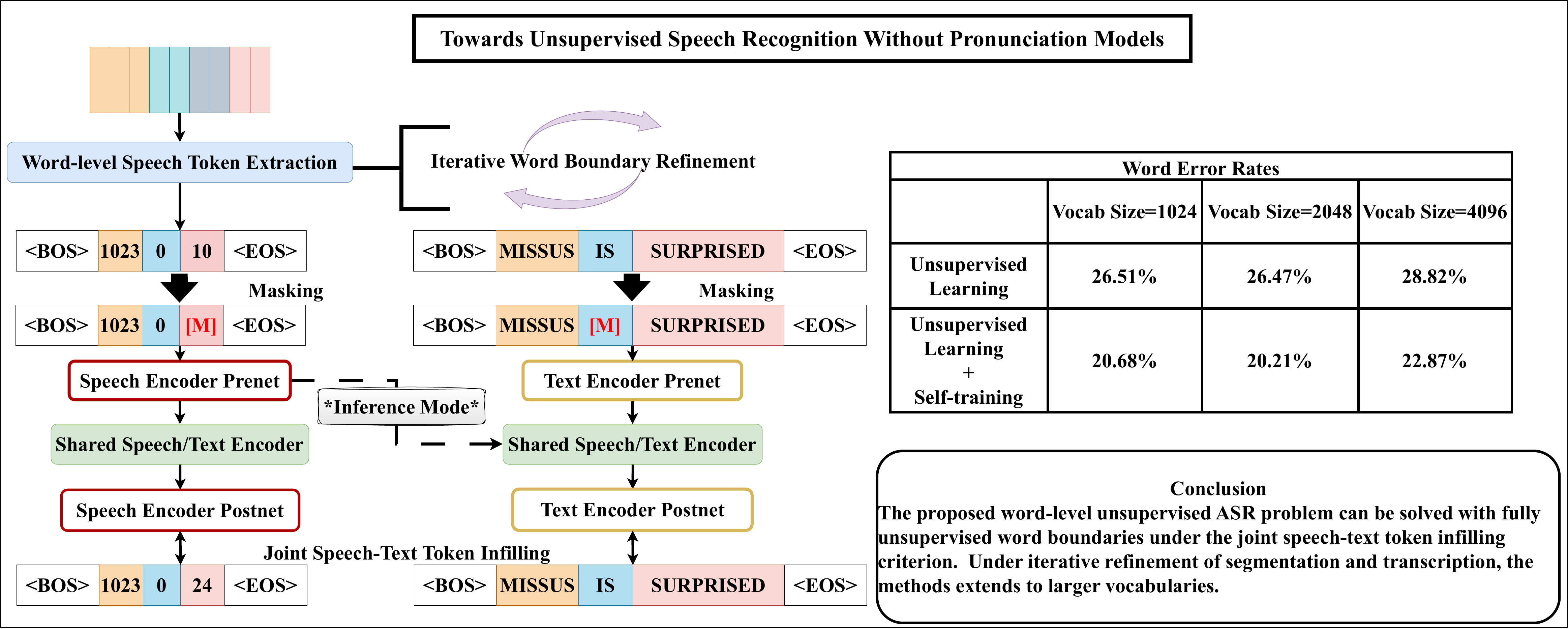}
\end{graphicalabstract}

\begin{highlights}
    \item Proposes word-level unsupervised ASR learned with unsupervised word boundaries.
    \item Proposes joint speech-text token infilling for learning word-level unsupervised ASR.
    \item Develops an iterative pipeline refining word-level segmentation and transcription.
    \item Shows that proposed method beats strong baselines and extends to larger vocabularies.
\end{highlights}


\end{frontmatter}



\section{Introduction}
Recent advancements in speech pre-training methods and models have significantly improved supervised speech recognition, achieving impressive phone and word error rates for several high-resource languages \citep{Baevski2020-wav2vec2,Hsu2022-hubert,Babu2022-xlsrv2}. However, these systems rely on abundant transcribed speech, making them impractical for low-resource languages. In extreme cases, there may be no parallel speech and text data available for a language. To address this, many studies have focused on using non-parallel speech and text corpora within the same language, tackling the challenge known as the unsupervised speech recognition (unsupervised ASR) problem \citep{Liu18-mapping-relation-gan,Yeh2019-unsup-asr,Chen2019,Baevski2021-wav2vec-u,Liu2022-wav2vecu2.0,wang-etal-2023-unsup-speech2sign,wang2024unsupervised,Tseng2024-reborn}.


The unsupervised ASR problem optimizes the following objective:
\begin{align} \label{uns_criterion}
    \min_{\theta}D_{f}\left[\mathbf{P^{\prime}} \left(Y\right) \lVert \mathbb{E}_{X}\left[\mathbf{Q}_{\theta} \left(Y \mid X \right)\right] \right]
\end{align}
under the assumption that
\begin{equation} \label{data_assumption}
    D_{g}\left[\mathbf{P} \left(Y\right) \lVert \mathbf{P^{\prime}} \left(Y\right)\right] < \epsilon
\end{equation}
in an appropriate sense, where $\mathbf{Q}_{\theta} \left(Y \mid X \right)$ is the output distribution given input speech $X$ over generated text $Y$ by the recognition model with parameters $\theta$, $\mathbf{P} \left(Y \mid X\right)$ is the ground-truth distribution over transcript $Y$ given speech $X$, $\mathbf{P^{\prime}} \left(Y\right)$ is the distribution over the unpaired text used for unsupervised learning, $\mathbf{P} \left(Y\right)$ is the distribution over ground-truth speech transcripts, and $D_f$ and $D_g$ are some reasonable divergence criteria to measure the distribution mismatch.
Note that Eq. (\ref{data_assumption}) ensures that the domain difference between the speech transcript hidden from us and the text data used for unsupervised learning is small. 
The above formulation shows three design considerations:
\begin{itemize}
    \item What is a reasonable granularity for the text space $Y$;
    \item What is a reasonable representation for speech $X$ so that we can cast $X$ into the predicted text space $Y$ more efficiently;
    \item What model and optimization algorithm serve as good candidates for minimizing $D_{f}\left[\mathbf{P^{\prime}} \left(Y \right) \lVert \mathbb{E}_{X}\left[\mathbf{Q}_{\theta} \left(Y \mid X \right)\right] \right]$.
\end{itemize}

For the representation/granularity of the raw and predicted text space, most previous unsupervised ASR systems  \citep{Liu18-mapping-relation-gan,Yeh2019-unsup-asr,Chen2019,Baevski2021-wav2vec-u,Liu2022-wav2vecu2.0,wang2024unsupervised} chose to work with phone-level speech features and phone-level text. While phone-level unsupervised ASR does not require paired speech and text to train, converting raw text to the phone sequences requires a grapheme-to-phone converter (G2P), and comes with the extra cost of developing a G2P converter or collecting a lexicon. While there are methods to build low-resource G2Ps with minimal resources \citep{MHJ-languagenet-g2p}, errors in the G2P system could propagate to the phone-level unsupervised ASR. Further, an unsupervised ASR system designed for the phone-level unsupervised ASR task cannot easily adapt to a G2P-free setting, even for letter-based languages with letter-based unsupervised training \citep{Liu2022-wav2vecu2.0,Ni-unsuptts-interspeech2022}. To reduce the reliance of unsupervised ASR systems on G2P converters, we propose solving a new task: word-level unsupervised ASR. We propose training an unsupervised ASR system that directly matches the distribution between predicted text from word-level speech features and word-level text. To constrain the problem as an early exploration, we curate synthetic corpora of different vocabulary sizes using fixed numbers of high-frequency word types from the base corpus; by synthetically varying the number of word types, we are able to propose and systematically study extensions to our algorithm that learn whole-word models despite heavy-tailed word frequency distributions.

As we choose the text space to be word-level sequences, we need to extract word-level information from speech to efficiently achieve Eq. (\ref{uns_criterion}). Studies have revealed that the speech representations extracted from speech foundation models pre-trained on speech-only data correlate well with underlying linguistic structures, such as phones and words \citep{pasad2022-comp-layer-analysis,Pasad2024-what-do-ssl-speech-words}. Previous work shows that phone-level unsupervised ASR benefits greatly from the phone-level features extracted with these foundation models \citep{Baevski2021-wav2vec-u,Liu2022-wav2vecu2.0,Gao2023-euro}. In our work for word-level unsupervised ASR, we pool discretized word-level features from the speech representations extracted from a HuBERT-Large model \citep{Hsu2022-hubert}.

Now that we have chosen the granularity of the text space and the representations of input speech, we design the model and optimization algorithm for minimizing $D_{f}\left[\mathbf{P^{\prime}} \left(Y\right) \lVert \mathbb{E}_X[\mathbf{Q}_{\theta} \left(Y \mid X \right)] \right]$. We propose a novel model for word-level unsupervised ASR based on \emph{joint speech-text token-infilling (JSTTI)} with the Transformer architecture. Inspired by the recent theme in unsupervised phone-level ASR and unsupervised speech segmentation that learns the segmental structure of speech in an end-to-end fashion \citep{Bhati2022-scpc,wang2024unsupervised,Tseng2024-reborn}, we add a \emph{differentiable boundary soft-pooler} to the JSTTI Transformer model and leverage an \emph{external boundary refinement routine} \citep{Algayres23-xlsr-wav2bnd} to improve the segmental structure of discretized word-level speech features. Finally, we leverage \emph{pseudo-text self-training} to boost the performance. Every step of our pipeline reduces the word error rate, and the resulting JSTTI system significantly outperforms two existing strong baselines for G2P-free unsupervised ASRs.

The contributions of the paper can be summarized as follows: 
\begin{itemize}
    \item We propose word-level unsupervised ASR learned completely with unsupervised word boundaries.
    \item We propose the JSTTI criterion for training word-level unsupervised ASR, and demonstrate that this criterion outperforms strong baselines.
    \item We develop a full pipeline that iteratively refines the word-level segmental structure, including an end-to-end differentiable boundary pooler built into the JSTTI task and an external boundary self-training routine leveraging speech foundation models.
    \item We show that our proposed JSTTI-based word-level unsupervised ASR remains performant across different vocabulary sizes. 
\end{itemize}

The structure of this paper is as follows.
The proposed whole-word unsupervised ASR builds on a large number of foundation models and published algorithms, which are described in Section \ref{sec:related_work}.
Section \ref{sec:models} describes the JSTTI model of word-level unsupervised ASR, including the segmentation pipeline, the JSTTI training criterion, a proposed method of end-to-end boundary refinement, and a student-teacher self-training method that considerably reduces error rates in most unsupervised ASR methods including JSTTI.  Section \ref{sec:method} describes the synthetically curated datasets, experimental settings, and baselines.  Section \ref{sec:result} describes the key result of this paper: whole-word unsupervised ASR is possible using iteratively refined segmentation, on top of which models are trained using a JSTTI training criterion.  The performance of these models depends on segmentation accuracy, therefore the dependence is explored in more detail in Section \ref{sec:segmentation}.  Dependence of results on vocabulary size is explored in Section \ref{sec:extend_vocab}, and it 
is demonstrated that iterative expansion of the vocabulary size can be performed with minimal increase in word error rate.  Finally, Section \ref{sec:conclusion} concludes.

\section{Background}\label{sec:related_work}
We discuss relevant background on unsupervised ASR, self-supervised learning, and unsupervised speech segmentation, which our proposed whole-word unsupervised ASR builds upon or compares against.

\subsection{Unsupervised Speech Recognition}
All previous studies of unsupervised ASR work with phoneme-level segmentation and phoneme-level text. Earlier models work with traditional speech features and solve the distribution matching problem in Eq. (\ref{uns_criterion}) with empirical output distribution matching \citep{Yeh2019-unsup-asr} or GAN training \citep{Liu18-mapping-relation-gan,Chen2019}. Recently, wav2vec-U \citep{Baevski2021-wav2vec-u} leverages acoustic features from speech foundation models \citep{Baevski2020-wav2vec2,conneau2020-xlsr53} and achieves competitive unsupervised ASR performance with GAN training and self-training. Later modifications remove the reliance on feature extraction heuristics \citep{Liu2022-wav2vecu2.0}, extend the choice of foundation models and decoding methods \citep{Gao2023-euro}, and incorporate an iterative method that refines the phoneme boundary segmental structure with reinforcement learning \citep{Tseng2024-reborn}. Among prior work, boundary refinement has been found useful to improve the results of the unsupervised systems \citep{Yeh2019-unsup-asr,Chen2019,Tseng2024-reborn}, which we also adopt into our system pipeline for word-level unsupervised ASR.

Our work uses wav2vec-U \citep{Baevski2021-wav2vec-u} together with REBORN \citep{Tseng2024-reborn} to establish baselines for G2P-dependent and G2P-free unsupervised ASRs. Our work also utilizes a recent approach called position-unigram and skipgram matching (PUSM) \citep{wang-etal-2023-unsup-speech2sign} for establishing word-level unsupervised ASR baselines. We briefly introduce these related backgrounds in the following sections.

\subsubsection{wav2vec-U and REBORN}\label{background:wav2vecu_REBORN}
\emph{wav2vec-U} \citep{Baevski2021-wav2vec-u} is a recent unsupervised automatic speech recognition (ASR) system that learns to recognize phones by minimizing the discriminability between recognized phone sequences and the phone sequences from an unpaired text corpus. They first extract speech representations from a self-supervised model pre-trained on thousands of hours of unlabelled speech \citep{Baevski2020-wav2vec2}, reduce its dimensionality using principal components analysis (PCA), and apply k-means pooling to build up the segment-level representations. The k-means pooling step, consisting of pooling consecutive frames with the same k-means label into cluster segments and a further step of mean-pooling between adjacent cluster segments, can be considered coarse phone-level segmentation. Then, they use a generative adversarial network (GAN) to learn the mapping from speech features to phone transcripts in an unsupervised manner, where the generator $G$ takes the segment representations and outputs phoneme distribution sequences, and the discriminator $D$ is trained against the generator to distinguish which source (real or generated) the input phoneme sequence is from. 
The learning algorithm randomly samples $n$ speech samples $\{X^{(1)},\ldots,X^{(n)}\}$ and $n$ unpaired text samples $\{Y^{(1)},\ldots,Y^{(n)}\}$, then adjusts the parameters of the discriminator to minimize $\mathcal{L}_{D}(D)$, and those of the generator to minimize $\mathcal{L}_{G}(G)$, where:
\begin{align}
\mathcal{L}_{D}(D)&=\sum_{i=1}^{n}-\left[\log \left(1-D\left(Y^{(i)}\right)\right) +  \log \left(D\left(G\left(X^{(i)}\right)\right)\right)\right]
\end{align} 
\begin{align}
\mathcal{L}_{G}(G)&=\sum_{i=1}^{n}-\log \left(1- D\left(G\left(X^{(i)}\right)\right)\right)
\end{align}

Several auxiliary losses for the generator or the discriminator help the model converge, including a gradient penalty loss, a phoneme entropy loss, and a smoothness loss between consecutive phoneme logits. As speech utterances may contain silences initially and finally and between consecutive words, they insert silence tokens into the text randomly at those positions to allow better distribution matching during GAN training. After GAN training, they apply greedy decoding to the generator's output over the training set, which they use as targets to train a phoneme HMM. The outputs from the phoneme HMM are decoded into words, and the decoded results are used as targets to fine-tune a wav2vec 2.0 checkpoint pre-trained on unlabeled speech under the character-level CTC loss \citep{Graves06-CTC}.

The REBORN framework \citep{Tseng2024-reborn} is the most recent follow-up for wav2vec-U. It starts with a converged wav2vec-U model and a convolutional neural network (CNN) segmentation model trained to clone the initial k-means-based segment boundaries. REBORN then iterates between two stages. In each iteration, the first stage uses policy gradient methods to refine a CNN-based phone-level segmentation model, given the fixed phone predictor in the wav2vec-U model from the previous iteration, and the second stage re-trains the wav2vec-U model using the re-segmented features obtained from the segmentation model in stage one. In more detail, each iteration repeats the following two stages:
\begin{itemize}
    \item In stage one, the segmentation model is trained with policy gradient algorithms based on REINFORCE \citep{williams1992-reinforce}, where the utterance-wise reward $R$ is a weighted combination of the perplexity difference reward $R_{ppl}$, edit-distance reward $R_{edit}$, and length difference reward $R_{len}$, and each separate reward is defined as:
    \begin{align}
        R_{ppl} = \operatorname{PPL}\left(Y^{'}_{\theta -1}\right) - \operatorname{PPL}\left(Y^{'}_{\theta}\right)
    \end{align}
    \begin{align}
        R_{edit} =  - \frac{d_{Lev}\left(Y^{'}_{\theta -1}, Y^{'}_{\theta}\right)}{\left |Y^{'}_{\theta -1}\right |}
    \end{align}
    \begin{align}
        R_{len} = 1 - \frac{\left|\left |Y^{'}_{\theta }\right | - \left | Y^{'}_{\theta - 1}\right | \right |}{\left |Y^{'}_{\theta -1}\right |}
    \end{align}
    where $Y^{'}_{\theta - 1}$ is the de-duplicated output obtained with the segmentation model $\pi_{\theta -1}$ and the phone predictor (i.e., the wav2vec-U generator) $G_{\theta -1}$ trained in the previous REBORN iteration $\theta - 1$,  $Y^{'}_{\theta}$ is obtained similarly but with $\pi_{\theta}$ and $G_{\theta -1}$, $\operatorname{PPL}\left(\cdot \right)$ is the language model (LM) perplexity score from a 4-gram phone LM, $d_{Lev}\left(\cdot,\cdot \right)$ is the Levenshtein distance and $\left|Y\right |$ denotes the length of text sequence $Y$.
    \item In stage two, the learned segmenter in stage one is used to obtain phone-level segmentation. The raw segmentation is optionally post-processed by merging consecutive segments that yield the same phone prediction from the phone predictor $G_{\theta -1}$. After that, the frame-level PCA features are re-segmented into phone-level features using the newly obtained segmentation, and a new phone predictor $G_{\theta}$ is trained under the wav2vec-U framework.
\end{itemize}

The REBORN framework allows iterative refinement of the phone-level segmental structures and outperforms wav2vec-U \citep{Baevski2021-wav2vec-u} and wav2vec-U 2.0 \citep{Liu2022-wav2vecu2.0} on several English and non-English corpora. However, wav2vec-U and REBORN have only been studied extensively in a G2P-dependent setting, and performance usually degrades significantly under a G2P-free setting with raw writing units such as characters \citep{Liu2022-wav2vecu2.0,Ni-unsuptts-interspeech2022}.

\subsubsection{Position Unigram and Skipgram Matching}\label{background:pusm}
The position-unigram and skipgram matching loss, or the \emph{PUSM} loss, was proposed in a recent work towards word-level unsupervised speech-to-sign-language translation \citep{wang-etal-2023-unsup-speech2sign}. Similar to our word-level unsupervised ASR task, they clustered word-level speech features obtained from forced alignments and trained a linear generator 
$G \in \mathbb{R}^{\left|\mathcal{X}\right| \times \left|\mathcal{Y}\right|}$
defined as 
\begin{equation}\label{eq:linear_generator}
G_{X_t,Y_t}:=\frac{\exp \left(W_{X_t, Y_t}\right)}{\sum_{Y^{\prime} \in \mathcal{Y}} \exp \left(W_{X_t,Y^{\prime}}\right)},
\end{equation}
where $Y_t$ denotes a visual sign cluster index, $X_t$ denotes the speech cluster index and $W_{X_t,Y_t}$ is the $(X_t,Y_t)^{\text{th}}$ element of a learned parameter matrix $W\in \mathbb{R}^{\left|\mathcal{X}\right| \times \left|\mathcal{Y}\right|}$.

During inference, the generator takes each discrete speech token in the sequence as input and converts it to text or sign-language clusters by taking $\argmax(\cdot)$ over the output probability vector.  During training, the generator 
is optimized using a weighted sum of two distribution matching losses. The first loss is the position-dependent unigram loss $\mathcal{L}_{\mathrm{pos}}(G)$, defined as: 
\begin{equation}\label{eqn:pos_unigram}
\mathcal{L}_{\mathrm{pos}}(G)=\sum_{t=1}^T\left\lVert\hat{P}_{X_t} G-\hat{P}_{Y_t}\right\rVert_1
\end{equation}
where $\hat{P}_{X_t}$ is the empirical unigram distribution of speech cluster indices at position $t$, and $\hat{P}_{Y_t}$ is the empirical unigram distribution of textual words at position $t$. The second loss is the skipgram loss:
\begin{equation}\label{eqn:skipgram}
\mathcal{L}_{\mathrm{skip}}(G)=\sum_{k=1}^K\left\lVert G^{\top} \hat{P}_k^{X X^{\prime}} G-\hat{P}_k^{Y Y^{\prime}}\right\rVert_1
\end{equation}
where $\hat{P}_k^{X X^{\prime}}$ and $\hat{P}_k^{Y Y^{\prime}}$ are the empirical distributions of position-independent lag-$k$ skipgrams aggregated over speech and text token sequences, respectively. The skipgram distribution ${P}_k^{Z Z^{\prime}}$ itself is defined as the joint distribution between two tokens that are $k$ positions apart, as:
\begin{equation}\label{eqn:biskipgram}
{P}_k^{Z Z^{\prime}}:= \operatorname{Pr}\left[Z_1=z, Z_{k+1}=z^{\prime}\right] =\frac{\sum_{i=1}^{L-k} P_{Z_i Z_{i+k}}\left(z, z^{\prime}\right)}{L-k}
\end{equation}
The word-level PUSM models for unsupervised ASR and unsupervised speech-to-sign require forced alignments to segment and pool speech features, which is an ideal assumption in the unsupervised scenario. The space complexity of the skipgram matching loss in Eq. (\ref{eqn:skipgram}) is of $\mathcal{O}(|V|^2)$, where $|V|$ is the vocabulary size, making the PUSM loss expensive to compute in large-vocabulary settings.

\subsection{Speech Representation Learning}
Speech foundation models pre-trained on large amounts of untranscribed speech audio \citep{Baevski2020-wav2vec2,Hsu2022-hubert,chen2022-wavlm} have been found to provide expressive speech representation for phone-level unsupervised ASR \citep{Baevski2021-wav2vec-u,Gao2023-euro}. Our work on word-level unsupervised ASR utilizes the features from a pre-trained HuBERT-Large model \citep{Hsu2022-hubert}, owing to recent probing analyses of word-level information learned in speech foundation models \citep{pasad2022-comp-layer-analysis,Pasad2024-what-do-ssl-speech-words}. The HuBERT approach is optimized under a loss similar to the Masked Language Modeling (MLM) loss over masked discrete tokens provided by an acoustic discovery model. We also leverage VG-HuBERT \citep{peng2022-vghubert}, a visually grounded, self-supervised speech model for word discovery, trained with paired speech captions and images on top of representations from the audio and image encoders. We utilize the unsupervised word boundaries inherent within the VG-HuBERT representations to obtain word-level cluster centroids for speech quantization.

There have been several successful attempts at joint speech-text pre-training \citep{Ao2022-speecht5,zhang2022-speechut,zhang2024-speechlm,bapna2021-slam,chen2022-maestro}. Our JSTTI-based unsupervised ASR model is inspired by the earlier SpeechT5 approach \citep{Ao2022-speecht5} that does not use paired data, which is a self-supervised pre-training framework that performs joint speech/text representation learning under a shared encoder-decoder architecture to reconstruct masked speech representations and masked text spans. We have adopted many of its design choices into our JSTTI model for word-level unsupervised ASR, including the masking function and the cross-modal vector quantizer. Our work is among the first to show that joint masked reconstruction on discrete speech and text leads to an unsupervised ASR model, similar to earlier work in unsupervised machine translation \citep{Artetxe18-unsupnmt,Lample2018-unsupmtmono}.

\subsubsection{HuBERT}
HuBERT is a self-supervised learning approach trained on unlabeled speech data. Given a speech utterance during training, a CNN encoder $G$ downsamples and converts raw speech $S$ into feature vectors $X = x_1, \cdots, x_T$ operating at 50Hz, and a proportion of the temporal indices $M \subset [T]$ are selected to be masked (with the corresponding feature vectors replaced with a learnable masked embedding $\tilde{x}$). The masked representation $r(X, M)$ is then fed into a Transformer encoder and a prediction head, jointly denoted as the prediction model $f$. For each frame at $t$, an acoustic unit discovery model provides the $C$-class categorical target $l_t$, and the model parameters are jointly optimized under the cross-entropy loss for all frames $t \in M$:
\begin{equation}
    \mathcal{L} \left(G, f; S, M, l_1, \cdots, l_T \right)= - \sum_{t \in M} \log p_f\left(l_t \mid r(G(S), M), t\right)
\end{equation}
where $p_f\left(\cdot \mid r(G(S), M), t\right)$ denotes the C-class categorical probability at the prediction head, given a masked version of the speech utterance as input.
For the first iteration of HuBERT training, a k-means model trained on top of 39-dimensional MFCC features acts as the acoustic unit discovery model for providing $l_t$'s, while for later iterations, k-means clustering is run on the latent features extracted from the HuBERT model pre-trained in the previous iteration. Fine-tuning a pre-trained HuBERT model for ASR allows one to obtain state-of-the-art error rates over various data scenarios, showing that the pre-trained representation encodes rich and meaningful speech structures and information.

\subsubsection{VG-HuBERT}\label{background:vghubert}
Unlike its speech-only pre-training variants, VG-HuBERT is trained on a paired speech-image corpus for the visual grounding task. The audio encoder is initialized as a pre-trained HuBERT encoder, and the image encoder is initialized as a pre-trained DINO Vision Transformer. 

The model takes, as input, an image $I_{i}$ and its spoken description $A_{i}$. A [CLS\_A] token is concatenated with audio features, and a [CLS\_I] token is concatenated with image patches before feeding into their respective transformer layers. A similarity score $S_{i,j}$ is calculated between the hidden embeddings of [CLS\_A] and [CLS\_I], which should be large when the speech faithfully describes the content of the image and small otherwise. The image encoder, audio encoder, and the two CLS token embeddings are jointly optimized with noise contrastive estimation, under the sum of $\mathcal{L}_{A \rightarrow I}$ and $\mathcal{L}_{I \rightarrow A}$:
\begin{align} \mathcal{L}_{A \rightarrow I} & =-\frac{1}{B} \sum_{i=1}^B \log \frac{e^{S_{i, i}-\delta}}{e^{S_{i, i}-\delta}+\sum_{j=1:j\ne i}^B 
e^{S_{i, j}}} \\ \mathcal{L}_{I \rightarrow A} & =-\frac{1}{B} \sum_{i=1}^B \log \frac{e^{S_{i, i}-\delta}}{e^{S_{i, i}-\delta}+\sum_{j=1:j\ne i}^B 
e^{S_{j, i}}}\end{align}
where $B$ is the batch size, and $\delta$ is a margin hyperparameter.

One significant discovery of \cite{peng2022-vghubert} is that word boundaries and segments appear naturally from visual grounding learning. After training the VG-HuBERT model, they perform word discovery by calculating the self-attention weight of the [CLS\_A] token across all speech frames at a specific layer. Each consecutive span of collected attention weights above a certain percentage is treated as a detected word.

\subsubsection{SpeechT5}
The SpeechT5 \citep{Ao2022-speecht5} framework is a self-supervised encoder-decoder pre-training framework incorporating unlabeled speech and text for joint speech/text representation learning. The two modalities share the same encoder-decoder part of the architecture. In contrast, each modality has its own encoder pre-net/post-net and decoder pre-net/post-net. The joint pre-training task is iterated between a speech pre-training step and a text pre-training step.

For the speech pre-training step, a masked speech representation from the speech encoder pre-net serves as the input into the encoder, and the decoder autoregressively reconstructs the spectrogram of the original speech waveform under the loss $\mathcal{L}_{\ar}^s$. At the same time, given the masked contextualized representation from the speech encoder pre-net, a masked language modeling loss $\mathcal{L}_{\text{MLM}}^s$ is calculated at the end of the encoder from a speech encoder post-net, with acoustic token targets provided by clustering the HuBERT model's latent representation:
\begin{equation}
\mathcal{L}_{\text{MLM}}^s=-\sum_{n \in \mathcal{M}} \log p_{E_s}\left(l_n \mid \tilde{\mathbf{X}}, n\right)
\end{equation}
\begin{equation}
\mathcal{L}_{\ar}^s=\sum_{n=1}^{N}\left\lVert\mathbf{X}_n-\hat{\mathbf{X}}_n\right\rVert_1
\end{equation}
where $\tilde{\mathbf{X}}$ denotes the masked speech representation from the shared encoder, $p_{E_s}$ denotes the probability distribution modeled by the encoder and the speech encoder post-net, $l_n$ denotes the acoustic token target at the $n^\text{th}$ frame, $\mathbf{X}_n$ denotes the $n^\text{th}$ frame of the target mel-spectrogram, and $\hat{\mathbf{X}}_n$ denotes the $n^\text{th}$ frame of the predicted spectrogram, obtained from the shared decoder together with the speech decoder post-net by conditioning on $\tilde{\mathbf{X}}$ and the previous teacher spectrogram frames $\mathbf{X}_{< t}$.

The text pre-training step follows a similar text-infilling approach of BART \citep{lewis-etal-2020-bart}, where they replace random text spans in the input text sentence with a single mask token. The text encoder pre-net takes a masked version of the text input and feed it into the shared encoder. The decoder is trained to reconstruct the original unmasked text sentence, conditioned on the previous tokens:
\begin{equation}
\mathcal{L}_{\ar}^t=\sum_{n=1}^{N} -\log p_{D_t}\left(\mathbf{Y}_n \mid \mathbf{Y}_{<n}, \tilde{\mathbf{Y}}\right),
\end{equation}
where $\tilde{\mathbf{Y}}$ is masked text representation from the shared encoder, $\mathbf{Y}_n$ denotes the text token at the $n^\text{th}$ position, $\mathbf{Y}_{<n}$ denotes all teacher text tokens before the $n^\text{th}$ position, and $p_{D_t}$ denotes the text distribution modeled by the shared decoder and the text decoder post-net, condition on $\tilde{\mathbf{Y}}$ and $\mathbf{Y}_{<n}$.

To learn the cross-modal representation without paired data, they add a vector quantizer after the encoder output, where a small percentage of the speech contextualized representation $\tilde{\mathbf{X}}$ and the text contextualized representation $\tilde{\mathbf{Y}}$ from the encoder across all time steps are replaced with their closest codebook vectors (in $l_2$ distance) from this shared quantizer. The losses during the speech pre-training and text pre-training steps are then augmented with a diversity loss to encourage sharing code vectors between speech and text representations:
\begin{equation}
\mathcal{L}_d=\frac{1}{K} \sum_{k=1}^K p_k \log p_k
\end{equation}
where $p_k$ is the average probability that the quantizer chooses the $k^\text{th}$ code vector.

The pre-trained architecture, when fine-tuned on multiple downstream speech-processing tasks, including automatic speech recognition, text-to-speech synthesis, speech-to-text translation, voice conversion, speech enhancement, and speaker identification, beat all comparable baselines at that time, demonstrating the power and wide range of applications for joint speech-text pre-training.

\subsection{Unsupervised Word Segmentation for Speech}
We require unsupervised word boundaries for pooling word-level information. Many recent unsupervised word segmentation algorithms follow either a top-down multi-level approach by incorporating differentiable segmenters and performing contrastive predictive coding \citep{Bhati2022-scpc,Cuervo2022-macpc, CuervoL2022-hcpc} or a bottom-up approach that discovers phone-like units and word units in separate steps \citep{Kamper2023-wordsegdp}. There have also been statistical approaches based on the Dirichlet process on top of an instance lexicon \citep{algayres-etal-2022-dp}. 

In addition to the VG-HuBERT word segmentation mentioned previously, our work borrows two additional word segmenters: GradSeg \citep{fuchs2023-gradseg} and XLS-R fine-tuning on unsupervised word boundaries \citep{Algayres23-xlsr-wav2bnd}. Our JSTTI-based end-to-end boundary refinement routine can be considered as a combination of the differentiable segmenter in Segmental-CPC \citep{Bhati2022-scpc} and the differentiable soft-pooler in segmental PUSM \citep{wang2024unsupervised}, but has been developed to include the features of both.

\subsubsection{GradSeg}
GradSeg \citep{fuchs2023-gradseg} is an efficient model that performs unsupervised word segmentation for speech utterances. Let $\left\{f_1, \cdots, f_T\right\}$ denote the features of an utterance extracted from a speech foundation model, say, a wav2vec-2.0 \citep{Baevski2020-wav2vec2} model pre-trained on untranscribed speech. The GradSeg module relies on one key finding: while simply applying a peak detection algorithm on top of the temporal gradient magnitude $m_t$ of the features, denoted as $ m_t=\left\lVert \frac{\mathbf{f}_{t+1}-\mathbf{f}_{t-1}}{2}\right\rVert^2$, yields mediocre word boundary detection, low values of $m_t$ are excellent indicators of far-from-boundary regions. Hence, the GradSeg method trains an efficient ridge regression model $c\left(f_t\right)$. The inputs $f_t$ are extracted from the training utterances, with their corresponding targets being $p_t=1_{\left[m_t>\theta\right]}$. The threshold $\theta$ is set to the lowest $p^\text{th}$ percentile of the gradient values over the training utterances. Hence, according to the previous observation, a negative label indicates non-boundary regions. After training, they apply a greedy non-maxima-suppression method on top of the regression scores $s_t = c\left(f_t\right)$ by specifying an average word duration and the minimum separation of consecutive word boundaries. This simple GradSeg method performs better than multiple prior statistical and neural word segmentation methods and only requires 100 utterances to train. 

\subsubsection{XLS-R Fine-tuning on Noisy Word Boundaries}
The XLS-R noisy boundary fine-tuning method is based on XLS-R \citep{Babu2022-xlsrv2},
a multilingual speech foundation model pre-trained on untranscribed speech from 128 languages.
XLS-R is fine-tuned to replicate noisy word boundaries obtained (without human annotation) by applying an off-the-shelf unsupervised word segmenter.  During fine-tuning, the boundaries
detected by the off-the-shelf unsupervised word segmenter 
are treated as ground truth, thus every boundary frame
and its two adjacent frames are given the labels
$b_t=1$, while all other frames are given the label 
$b_t=0$.
For each utterance in the fine-tuning corpus,
the XLS-R model 
produces a sequence of learnable embeddings $\{f_1, \cdots, f_T\}$, which are fed into a randomly initialized sigmoid layer;
all parameters of XLS-R and of the sigmoid layer are fine-tuned to minimize binary cross-entropy of the unsupervised boundary labels $\{b_1,\ldots,b_T\}$.
Due to label imbalance, 
binary cross-entropy is only backpropagated from a predetermined fraction of frames that have the highest loss. During inference, the fine-tuned XLS-R model predicts framewise word boundary probabilities $\{p_1,\ldots,p_T\}$, and a peak detection algorithm finds the temporal local maxima, which are interpreted as word boundaries. By collectively training across five multilingual corpora, this noisy boundary XLS-R fine-tuning method can consistently improve the segmentation performance of multiple off-the-shelf, unsupervised word segmenters and establishes the state-of-the-art unsupervised word segmentation results over multiple languages, even in some zero-shot settings.

\subsubsection{Segmental-CPC and Segmental PUSM}
Segmental Contrastive Predictive Coding \citep{Bhati2022-scpc} (SCPC) is an extension of Contrastive Predictive Coding \citep{vandenoord2018-cpc} (CPC) for speech self-supervised learning. The SCPC method builds a two-level CPC network that solves two contrastive learning tasks, one at the frame level and another at the segment level. 

The frame-level model learns a CNN, $f_{enc}$, to extract frame-level representation $\mathbf{z}$ from raw audio under the next frame classification loss:
\begin{equation}
\mathcal{L}_{\mathrm{NFC}}=-\log \frac{\exp \left(\operatorname{sim}\left(\mathbf{z}_t, \mathbf{z}_{t+1}\right)\right)}{\sum_{\tilde{\mathbf{z}} \in \mathcal{Z}_t} \exp \left(\operatorname{sim}\left(\mathbf{z}_t, \tilde{\mathbf{z}}\right)\right)}
\end{equation}
where the set $\mathcal{Z}_t$ contains $\mathbf{z}_{t+1}$ and $K$ distractor frames from the same utterance, and $\operatorname{sim}(x,y)$ is the cosine similarity between $x$ and $y$.

After obtaining the frame-level representation, SCPC constructs a differentiable boundary detector to pool segment-level features. The segment boundaries are obtained from the frame-level dis-similarity scores $\mathbf{d} = \left(d_1, d_2,\dots d_L\right)$ via peak finding:
\begin{align} 
p_t^{(1)} & =\min \left(\max \left(d_t-d_{t+1}, 0\right), \max \left(d_t-d_{t-1}, 0\right)\right) \nonumber \\ p_t^{(2)} & =\min \left(\max \left(d_t-d_{t+2}, 0\right), \max \left(d_t-d_{t-2}, 0\right)\right) \nonumber \\ p_t & =\min \left(\max \left(\max \left(p_t^{(1)}, p_t^{(2)}\right)-\text { thres, } 0\right), p_t^{(1)}\right)
\end{align}
where $p_t^{(2)}$ is used to smooth $p_t^{(1)}$, and $p_t$ is zero when the $t^\text{th}$ frame is not a boundary and non-zero if it is a boundary. To obtain a differentiable version of the boundary signals $b_t$ without saturating the gradients, SCPC calculates:
\begin{align} 
\mathbf{b}_{\text {soft }} & =\tanh (10 \mathbf{p}) \nonumber \\ \mathbf{b}_{\text {hard }} & =\tanh (1000 \mathbf{p}) \nonumber \\ \mathbf{b} & =\mathbf{b}_{\text {soft }}+\operatorname{sg}\left(\mathbf{b}_{\text {hard }}-\mathbf{b}_{\text {soft }}\right)
\end{align}
where $\operatorname{sg}\left(\cdot\right)$ is the stop-gradient operator.

The input to the segment-level model is a set of features $\{\overline{\mathbf{z}}_1,\ldots,\overline{\mathbf{z}}_L\}$, where $\overline{\mathbf{z}}_l$ is the weighted mean of frame-level features in the $l^{\text{th}}$ soft-boundary segment, thus $\overline{\mathbf{z}}_l=\sum_{t \in S_l} h_{l, t}\mathbf{z}_t$, where $S_l$ denotes the set of frames that are segmented into the $l^{\text{th}}$ segment and $h_{l, t} \in [0,1]$ is the weight assigned to the $t^{\text{th}}$ frame for the $l^{\text{th}}$ segment. 
SCPC learns a CNN segment encoder $s_{enc}$ to produce segment features $\mathbf{s}$ and an additional gated recurrent unit (GRU) context network $s_{ar}$ on top of $\mathbf{s}$ to produce the context representation $\mathbf{c}_t = s_{ar}\left(s_t\right)$.  The segment-level model is trained under the next segment classification loss:
\begin{equation}
\mathcal{L}_{\mathrm{NSC}}=-\log \frac{\exp \left(\operatorname{sim}\left(\mathbf{c}_t, \mathbf{s}_{t+1}\right)\right)}{\sum_{\tilde{\mathbf{s}} \in \mathcal{S}_t} \exp \left(\operatorname{sim}\left(\mathbf{c}_t, \tilde{\mathbf{s}}\right)\right)}
\end{equation}
where the set $\mathcal{S}_t$ contains segment $\mathbf{s}_{t+1}$ and $K$ distractor segments from the same utterance.

During inference, SCPC runs a peak detection algorithm on dis-similarity scores of the frame-level features $\mathbf{z}$ for phoneme segmentation. Similarly, running the peak detection algorithm on the dis-similarity scores between context representation $\mathbf{c}$ and segment representation $\mathbf{s}$ gives unsupervised word boundaries. SCPC performs better than other neural and non-neural baselines for phoneme and word segmentation before its publication.

The segmental-PUSM model \citep{wang2024unsupervised} provides an alternative method for phone-level unsupervised ASR and unsupervised phone segmentation based on a differentiable version of the position-unigram and N-skipgram loss applied to unpaired speech and phonemicized text. The system trains a generator $G(x)$ that tries to match the position-unigram and N-skipgram distributions between the text corpus $\mathcal{Y}$ and the distribution of $G\left(\mathcal{X}\right)$, where $\mathcal{X}$ denotes the speech corpus. The generator $G$ takes frame-level features from the wav2vec 2.0 \citep{Baevski2020-wav2vec2} model and the frame-level discrete cluster sequence obtained from the continuous features as inputs. The generator is composed of several parts: 1) a CNN modality converter that takes frame-level discrete cluster sequences $X_1, \cdots ,X_T$, and outputs the predicted frame-level phoneme probabilities $\hat{\mathbf{y}}_1,\cdots, \hat{\mathbf{y}}_T$, 2) a CNN segmenter that takes continuous, frame-level features $\mathbf{x}_1,\cdots, \mathbf{x}_T$ from the wav2vec 2.0 model as input and outputs phoneme boundary probabilities $p_1,\cdots, p_T$, and 3) a soft-pooler that takes in the predicted frame-level phoneme boundary probabilities $\hat{\mathbf{y}}_1,\cdots, \hat{\mathbf{y}}_T$ and obtains segment-level predicted phoneme probabilities $\overline{\mathbf{y}}_1,\cdots, \overline{\mathbf{y}}_L$, where $\overline{\mathbf{y}}_l=\sum_t\mathbf{p}_t\softmax_t\left(-\left|l-\sum_{\tau=0}^Tp_\tau\right|\right)$. The position unigram and N-skipgram loss are used for distribution matching between the phone segment probabilities from the generator $G$. The position-unigram loss is the same as described in Eq. (\ref{eqn:pos_unigram}), while the N-skipgram loss simply extends the bi-skipgram loss in Eq. (\ref{eqn:skipgram}) to N-skipgrams, which is defined as the joint distribution over $N$ tokens with skip-sizes $\mathbf{k} = (k_1, \cdots k_{N-1})$ between consecutive tokens (instead of just two tokens as in the bi-skipgram, i.e., Eq. (\ref{eqn:biskipgram})):
\begin{equation}
    {P}^{N}_{Y, \mathbf{k}} \left(y_{1\cdots N}\right)= P_{Y_1, Y_{1 + k_1}, \cdots, Y_{1 + \sum_{j=1}^{N-1} k_j}}\left(y_{1\cdots N}\right)
\end{equation}

While the segmental-PUSM method lags behind wav2vec-U \citep{Baevski2021-wav2vec-u} in terms of phone-level unsupervised ASR, by leveraging the unpaired phone-level text, it outperforms several strong baselines in unsupervised phone segmentation.
\section{Model Architecture, Training, and Refinement}\label{sec:models}

The proposed unsupervised ASR is an end-to-end differentiable
model of segmentation and feature extraction, trained to 
minimize discriminability between word sequences extracted
from speech versus those extracted from text.  
Section~\ref{subsec:model_training} 
describes the unsupervised ASR training criterion with discretized speech tokens at the word level and unpaired text of whole words.
Section~\ref{subsec:extract_word_level_features} describes
unsupervised word segmentation
and feature extraction algorithms to obtain word-level speech tokens. 
The methods of Section~\ref{subsec:unsup_bnd_and_refinement} are used to refine unsupervised word segmentation and allow our proposed unsupervised ASR training criterion in Section~\ref{subsec:model_training} to be backpropagated into the segmentation model. Finally, 
Section~\ref{subsec:self_train_trans} describes iterative
refinement of the model using pseudo-text self-training.




\subsection{Joint Speech-Text Token-infilling with Transformer}\label{subsec:model_training}
Inspired by SpeechT5 \citep{Ao2022-speecht5} and prior work in unsupervised NMT \citep{Artetxe18-unsupnmt,Lample2018-unsupmtmono}, we develop a \emph{joint speech-text token-infilling (JSTTI) approach} for word-level unsupervised ASR. The base version (i.e., without end-to-end speech-text boundary refinement introduced in Section \ref{subsec:unsup_bnd_and_refinement}) in our main results is a Transformer encoder shared between the two modalities. The Transformer takes masked speech or text token sequences as inputs and predicts the masked and unmasked regions of the original sequence. The speech tokens are considered to be word-level, as the frame-level speech features are mean-pooled within consecutive unsupervised word boundaries before applying quantization.

\begin{figure}[H]
  \centering
  \includegraphics[width=\linewidth, trim={0 0 10 0}, clip]{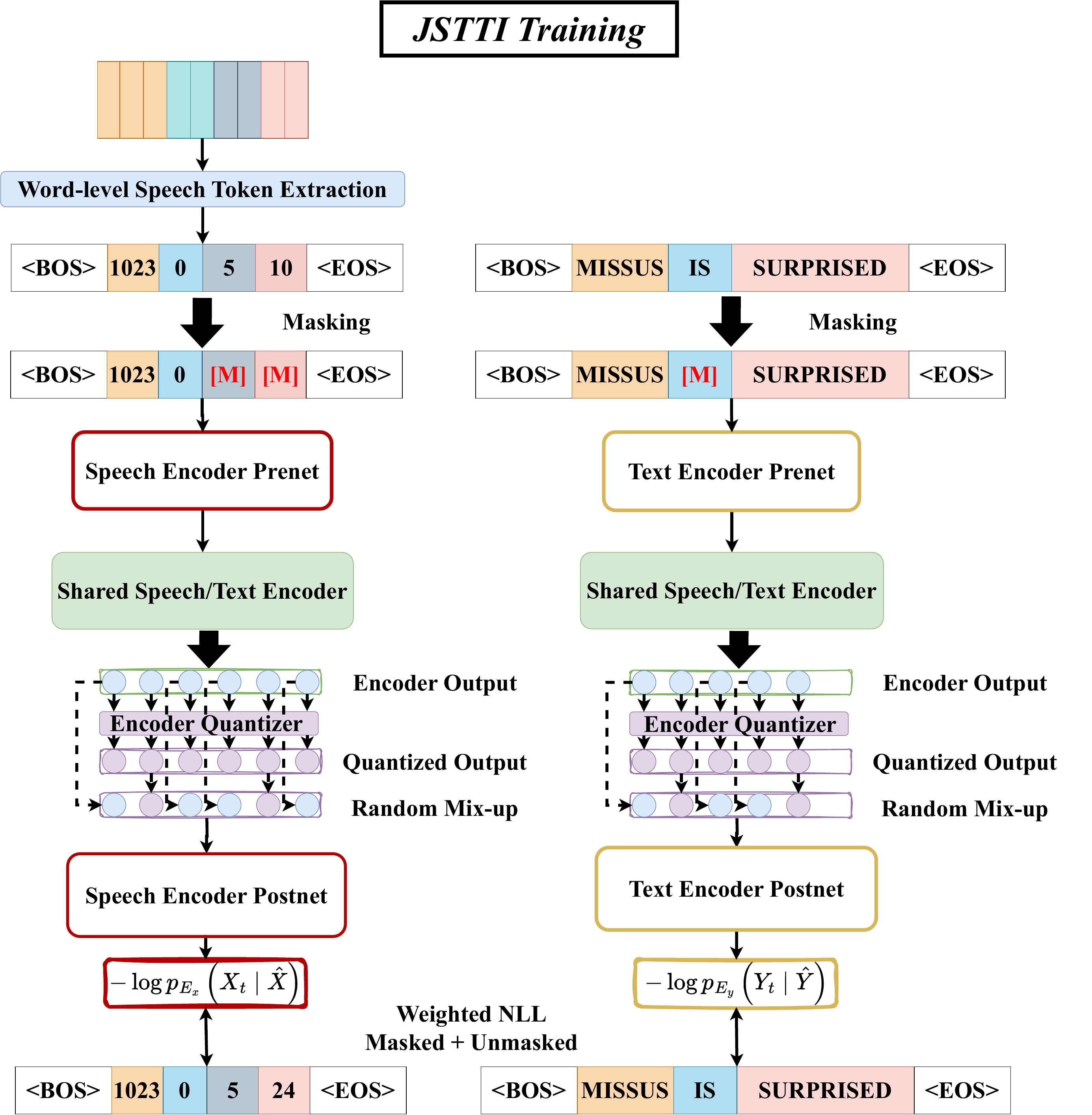}
  \caption{Illustration of Joint Speech-Text Token Infilling during training. The encoder-only model is shared between speech-to-speech token infilling (left branch) and text-to-text token infilling (right branch). If $N$ consecutive tokens are masked on the input side, they are replaced with $N$ $<\!\text{MASK}\!>$ tokens (denoted as [M] in the figure) or with samples randomly drawn from the modality-specific vocabulary. The word-level speech token extraction process is detailed in Section \ref{subsec:extract_word_level_features}.}
  \label{fig:justspeechwordt5_enc}
\end{figure}

We train the JSTTI encoder model with a training criterion based on the text-infilling approach of BART \citep{lewis-etal-2020-bart}, with hyperparameters based on the 
values used in SpeechT5 \citep{Ao2022-speecht5}. Specifically, we first sample spans of discrete tokens following a Poisson distribution with a mean length of $3.5$ tokens, and the total length of the sampled spans is limited to less than $30\%$ of the input sequence length. We then replace each token within $90\%$ of the sampled spans (chosen at random) with a $<\!\text{MASK}\!>$ token, and each token within the remaining $10\%$ of the sampled spans with a random token from the modality-specific vocabulary set. 
Because we use an encoder-only architecture, we do not change the length of the masked sequence as in BART or SpeechT5. 
In a default setting that we call ``random mix-up,''
some of the speech and text encoder outputs (selected
uniformly at random with $p=0.3$) are further 
randomized using a shared Gumbel-softmax vector quantizer.
Because of random mix-up, the model is usefully trained
using a weighted combination of both the masked and unmasked
portions of its target sequence, thus the 
training criterion is given by:
\begin{align}\label{eqn:speech_mle_enc}
\mathcal{L}_{\text {wNLL}}^\mathbf{X}&=-\sum_{t\in \mathcal{T}_m} \log p_{E_x}(X_t \mid \hat{X}) -\lambda \sum_{t\in \mathcal{T}_u} \log p_{E_x}(X_t \mid \hat{X}) \\
\label{eqn:text_mle_enc}
\mathcal{L}_{\text{wNLL}}^\mathbf{Y}&=-\sum_{t\in \mathcal{T}_m} \log p_{E_y}(Y_t \mid \hat{Y})
-\lambda \sum_{t\in \mathcal{T}_u} \log p_{E_y}(Y_t \mid \hat{Y})
\end{align}
where``wNLL'' stands for ``weighted negative log likelihood loss,'' $\mathcal{T}_m$/$\mathcal{T}_u$ denotes the set of masked/unmasked positions, $p_{E_x}(X_t\mid\hat{X})$ is 
the probability of speech cluster index $X_t$ given masked speech sequence $\hat{X}$,
$p_{E_y}(Y_t|\hat{Y})$ is the probability of text word $Y_t$ given masked text sequence $\hat{Y}$, and $\lambda$ is a
hyperparameter.
Figure~\ref{fig:justspeechwordt5_enc} shows the overall architecture.

\begin{figure}[H]
  \centering
  \includegraphics[width=0.7\linewidth, trim={0 0 0 0}, clip]{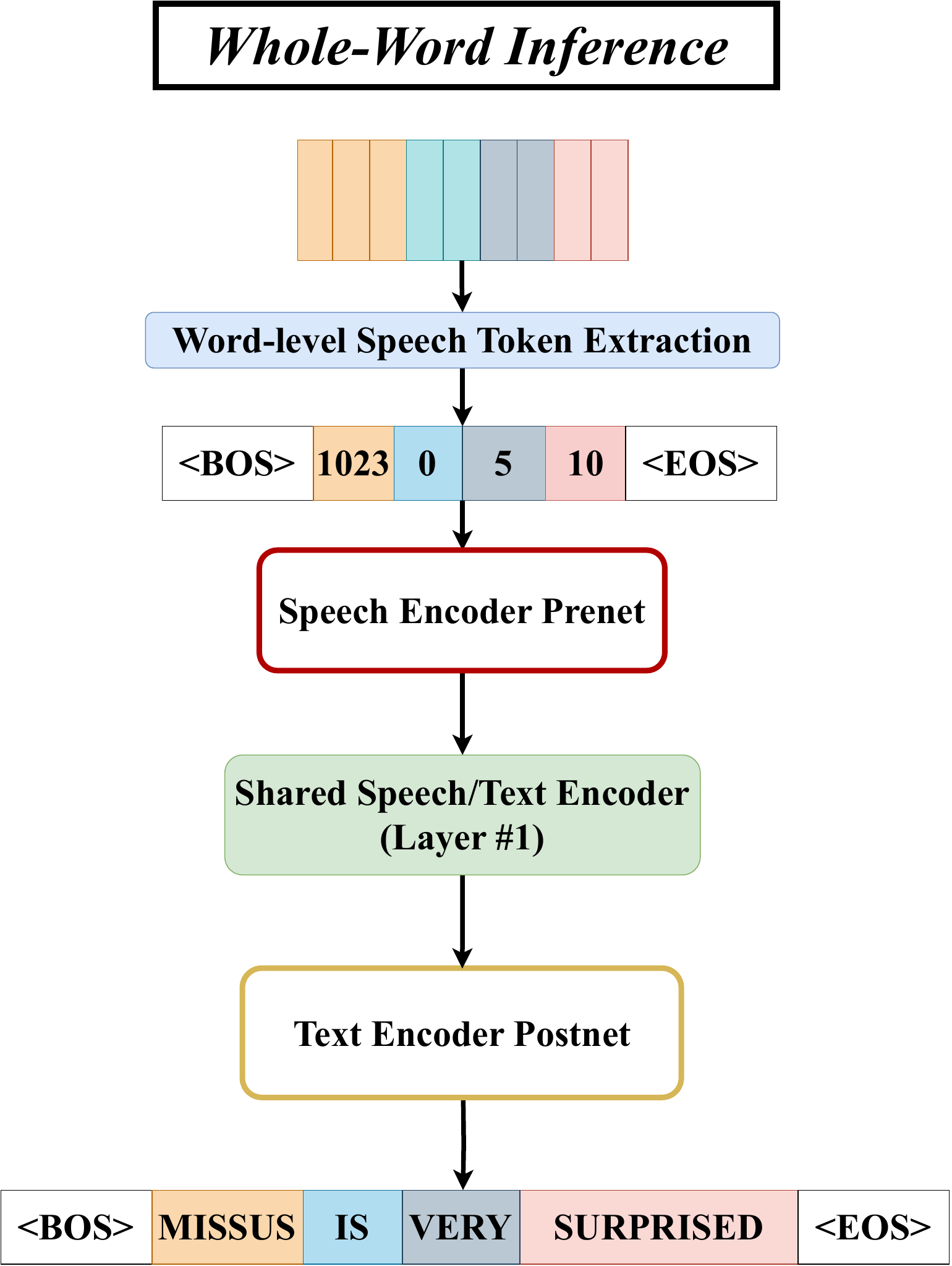}
  \caption{Whole-word ASR inference process after JSTTI training. The encoder takes speech tokens as input and predicts a text token for each token representation.}
  \label{fig:inference_justspeechwordt5_enc}
\end{figure}

Figure~\ref{fig:inference_justspeechwordt5_enc} shows the whole-word ASR inference process after JSTTI training. During inference, the speech encoder prenet takes the speech token input, feeds it through the speech/text encoder, strips off the last trained layer, and feeds the second-to-last layer embedding to the text encoder postnet to get the text hypothesis. We obtain much lower WER if 
the text-modality postnet takes inputs from the 
second-to-last trained layer of the speech encoder, rather than the
last layer. 
Early work on unsupervised and transfer learning \citep{Bengio2012-deeptransfer}
suggests that stripping off the last layer of a learned
encoder is a general technique for improving cross-modality 
transferability of a learned representation.
We explore this specific design choice in \ref{subsec:model_layerwise}.

\subsection{Extracting Word-level Speech Representations}\label{subsec:extract_word_level_features}
\begin{figure}[H]
  \centering
  \includegraphics[width=\linewidth]{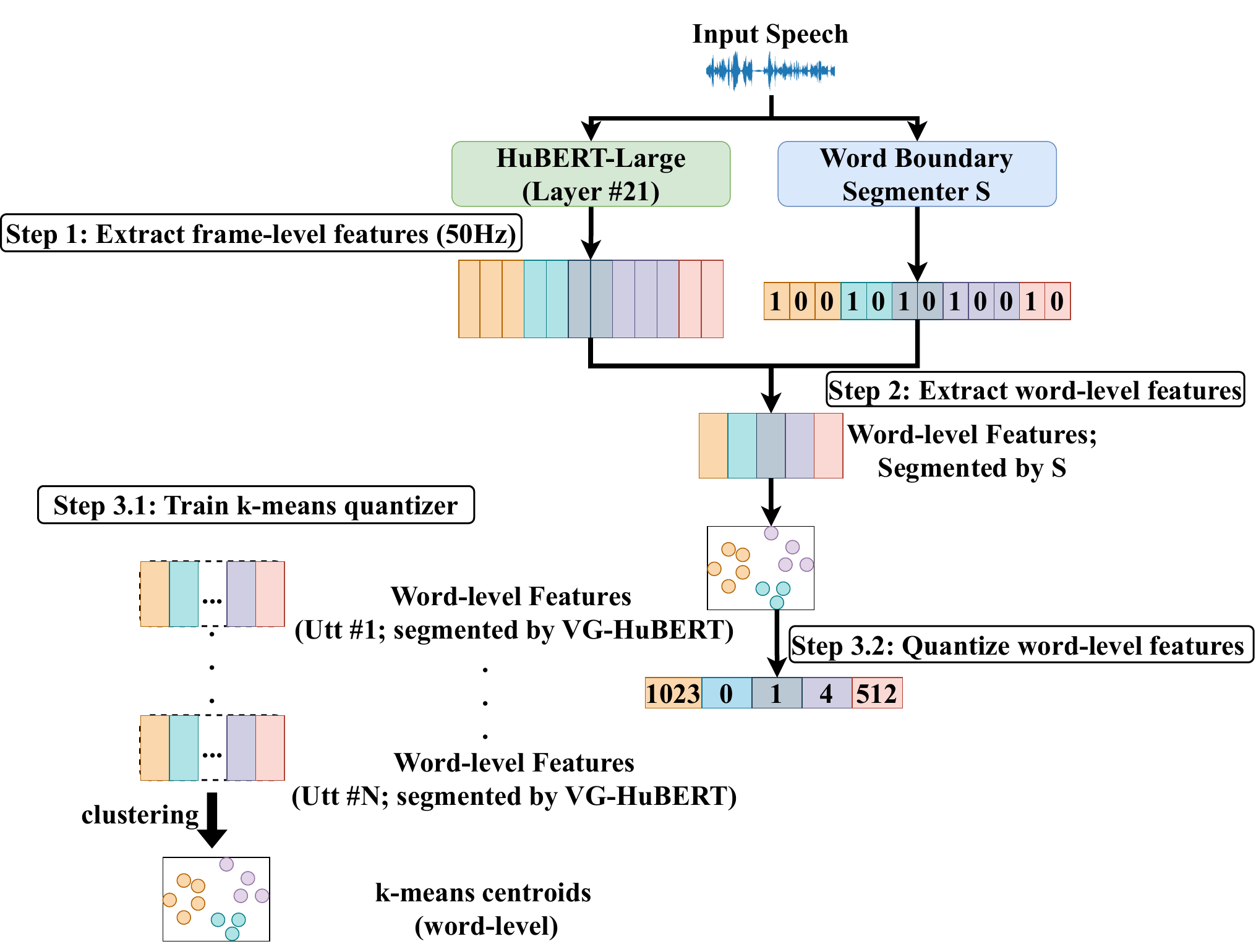}
  \caption{Word-level speech token extraction. The three-step speech feature extraction routine converts continuous speech into discrete, word-level tokens. Step 3.1 only uses word-level features extracted from VG-HuBERT for training the k-means quantizer, even if the segmentation model $S$ used in Step 2 is not VG-HuBERT.}
  \label{fig:extract_features_steps1to3}
\end{figure}

A speech foundation model carries word-level information in its intermediate-layer representations \citep{pasad2022-comp-layer-analysis, Pasad2024-what-do-ssl-speech-words}. 
In order to focus experiments on the task of learning whole-word representations, we assume the availability of a 
HuBERT-Large model pre-trained on 60k hours of untranscribed English speech \citep{Hsu2022-hubert} to provide acoustic features $X$. The overall process is displayed in Figure \ref{fig:extract_features_steps1to3}.

As the first step, we \emph{extract frame-level continuous features} from the 21st layer of the HuBERT-Large model, as the speech features from this specific layer are strongly correlated with word-level information \citep{pasad2022-comp-layer-analysis}.

The frame-level speech features from the HuBERT-Large model operate at 50Hz, while the frequency of individual spoken words may only average around 4Hz. To further compress the frame-level features, we \emph{mean-pool the frame-level features} between consecutive pairs of word boundaries to extract word-level speech features. Many different types of word boundary segmenters are tested in Section~\ref{sec:result}, including boundaries derived from a force-aligned reference transcript (an oracle setting included as a top-line), 
off-the-shelf unsupervised word boundary detectors such as \emph{VG-HuBERT} \citep{peng2022-vghubert} and \emph{GradSeg} \citep{fuchs2023-gradseg}, and a method based on XLS-R unsupervised boundary self-training \citep{Algayres23-xlsr-wav2bnd} that we call \emph{wav2bnd}.
The \emph{wav2bnd} algorithm starts with a pre-trained 0.3B XLS-R checkpoint \citep{Babu2022-xlsrv2} with raw speech as input, and is then fine-tuned to reproduce pseudo-ground-truth boundary labels generated by one of the other unsupervised word segmentation algorithms (e.g., \emph{GradSeg}), following the word boundary fine-tuning heuristics in \cite{Algayres23-xlsr-wav2bnd}. During inference, \emph{wav2bnd} applies the same peak detection algorithm in \cite{Algayres23-xlsr-wav2bnd} on the frame-level boundary probabilities from the fine-tuned XLS-R model, with detection parameters tuned so that the average number of detected words per utterance roughly aligns with the prior temporal frequency of spoken words. We denote the resulting boundaries as \emph{GradSeg+wav2bnd}, as the training labels are created by \emph{GradSeg}.

For the third step, we \emph{quantize} the continuous word-level speech feature vectors into discrete tokens using a fixed set of cluster centroids. This step is divided into two substeps: obtaining the quantization model and obtaining the word-level quantized sequence. To obtain the quantization model, we run the k-means clustering algorithm on top of the word-level features extracted with word boundaries provided by VG-HuBERT. In the second substep, we simply \emph{fix} this k-means quantization model and apply it on top of the word-level continuous feature vectors pooled with different word segmenters. In other words, we do not re-cluster the word-level feature vectors to obtain new centroids for quantization, even if the word-level continuous feature vectors are obtained with a word segmenter other than VG-HuBERT. In \ref{subsec:choice_word_clusters}, we show that this specific choice is reasonable.

From now on, we will formally denote the discrete word-level speech cluster sequence as $X$ and the discrete word-level text sequence as $Y$. The unsupervised word-level ASR is a model $G(X)$ learned in order to minimize the distortion between the distributions $P(Y)$ and $P(G(X))$.


\subsection{Unsupervised Word Boundary Refinement}\label{subsec:unsup_bnd_and_refinement}
The base JSTTI models trained with discrete word-level speech tokens extracted with \emph{VG-HuBERT} or \emph{GradSeg} boundaries do not achieve good WER.  Boundaries inferred using \emph{GradSeg+wav2bnd} are quite good, but further improvement is possible by adding an end-to-end differentiable boundary soft-pooler to the JSTTI model for improved joint speech-text modeling.

\begin{figure}[H]
  \centering
  \includegraphics[width=0.8\linewidth,trim={0 0 0 0},clip]{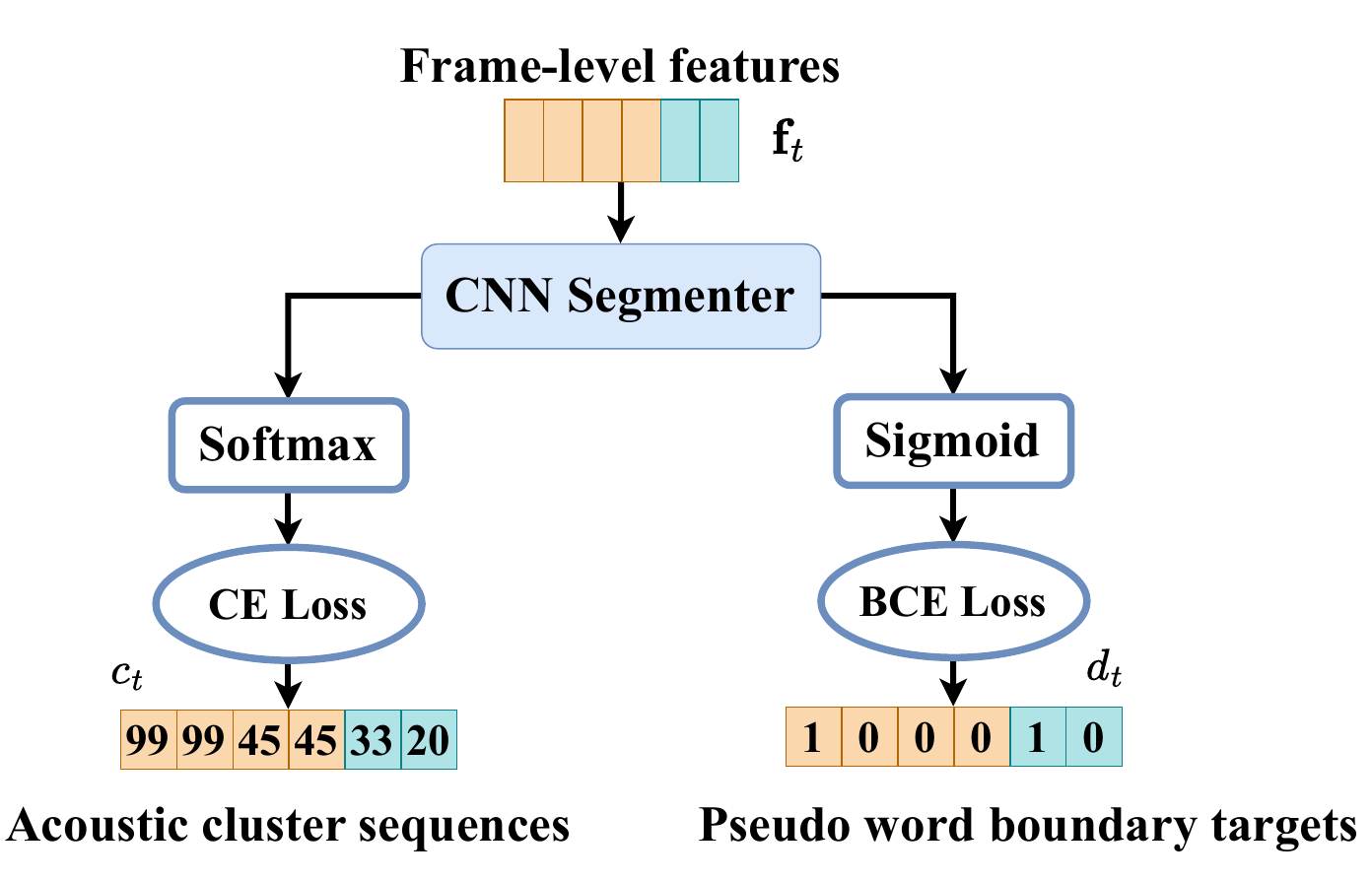}
  \caption{Behavior-cloning stage of the CNN segmeter. We train a simple CNN classifier that takes frame-level speech features as input and jointly predicts the frame-level \textit{Gradseg + wav2bnd} unsupervised word boundary targets and the frame-level acoustic cluster labels as outputs. In the figure, CE stands for cross-entropy loss and BCE stands for binary cross-entropy loss.}
  \label{fig:justcnnsegmenter}
\end{figure}

The differentiable soft-pooling algorithm is initialized by training a small CNN with two output layers: a binary output classifier that mimics the word boundary predictions from \textit{GradSeg+wav2bnd}, and a multi-class output layer that predicts frame-level acoustic cluster indices (Figure \ref{fig:justcnnsegmenter}). The primary purpose of the acoustic clustering output layer is to prevent the CNN from overfitting to the \emph{GradSeg+wav2bnd} boundary prediction task.  Prevention of overfitting is adequately supported by the use of a multi-class output layer that predicts frame-wise segment labels generated by k-means clustering of the  6$^\text{th}$ layer of the HuBERT-base model. We choose the number of frame-level acoustic clusters to be 100 to match previous settings for unit-based speech resynthesis \citep{polyak21-speech-resynthesis}. 
Figure \ref{fig:justcnnsegmenter} shows how this additional CNN segmenter works during this behavior-cloning stage.

\begin{figure}[H]
\centering
\includegraphics[width=\linewidth,trim={0 0 0 0},clip]{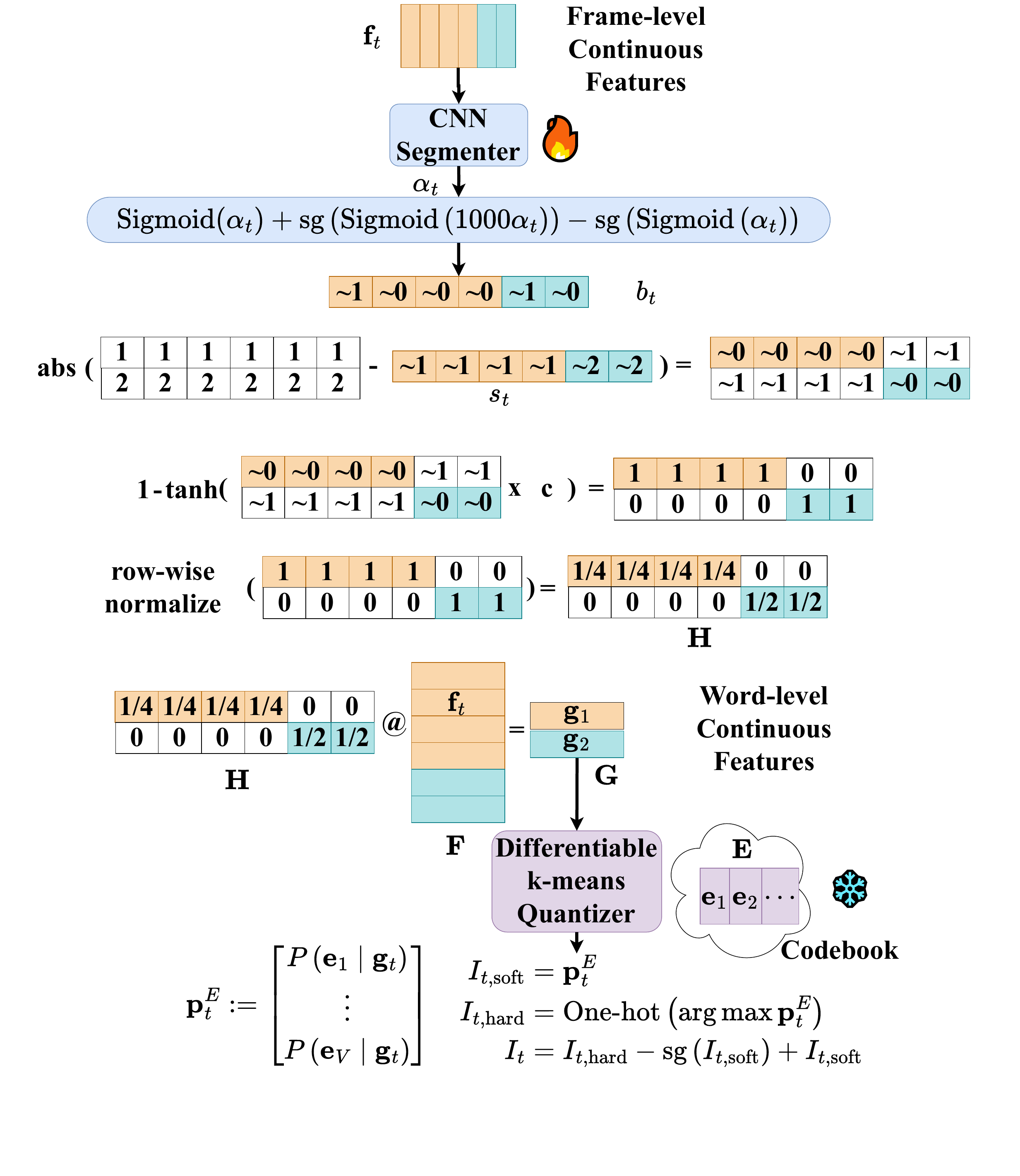}
  \caption{Illustraining example of the differentiable soft-pooler. This example shows how the differentiable soft-pooler obtains word-level features from the boundary probability predictions of the CNN segmenter and how the speech discrete cluster sequence is obtained in a differentiable manner. The variables in the figure are referenced in the description. The symbol ``$\sim$’’ means ``a real number close to,’’ e.g., ``$\sim2$’’ denotes a real number between 1.5 and 2.5. Note that the CNN segmenter is trainable, while the differentiable k-means quantizer is not.}
\label{fig:mean_pool_diff_bnd}
\end{figure}

After both the CNN segmenter and the JSTTI model converge, we combine them for joint fine-tuning. Figure \ref{fig:mean_pool_diff_bnd} shows how the CNN segmenter converts continuous frame-level speech features into discrete speech cluster sequences in a differentiable manner. We first describe the process we use to mean-pool within consecutive word boundaries predicted by the CNN segmenter in a differentiable manner, which we denote as the \emph{differentiable boundary soft-pooler} \citep{Bhati2022-scpc,wang2024unsupervised}. Let $\mathbf{f}_t$ denote the frame-level features at frame $t$, and $\mathbf{g}_m$ denote the word-level features for the $m^{\text{th}}$ segment. We further denote $\mathbf{F} = \begin{bmatrix}\mathbf{f}_1 &\cdots & \mathbf{f}_T \end{bmatrix}^{\top}$,  $\mathbf{G} = \begin{bmatrix}\mathbf{g}_1 & \cdots & \mathbf{g}_M \end{bmatrix}^{\top}$, $\mathbf{H}\in\Re^{M\times T}$ whose
$(m,t)^{\text{th}}$ element is $h_{m,t}$, 
and let the logits generated
by the CNN segmenter be $\{\alpha_1,\ldots,\alpha_T\}$,
where $\alpha_t\in\Re$, then
\begin{align}\label{eqn:diff_bnd_whole_ops}
b_t&=\sigma(\alpha_t)+
\text{sg}\left(\sigma(1000\alpha_t)-\sigma(\alpha_t)\right)\\
s_t&=\sum_{\tau=1}^t b_\tau\\
h_{m,t}&=
\frac{1-\text{tanh}(c|m-s_t|)}{\sum_{\tau=1}^T\left(1-\text{tanh}(c|m-s_\tau|)\right)}\\
\mathbf{G}&=\mathbf{H}\mathbf{F}
\end{align}
where $\sigma(x)=\frac{1}{1+\exp(-x)}$ is the sigmoid function, $\text{tanh}(x)=2\sigma(2x)-1$ is the hyperbolic tangent, $\text{sg}(\cdot)$ is the stop-gradient operator, and $c$ is a positive real hyperparameter. This sequence of operations is depicted graphically in Figure \ref{fig:mean_pool_diff_bnd}.
We further describe the process denoted as the \emph{differentiable k-means quantizer}, where word-level features $\mathbf{g}_t$ go through a k-means vector quantizer with a fixed codebook $\mathbf{E} = \begin{bmatrix}\mathbf{e}_1 & \cdots & \mathbf{e}_V\end{bmatrix}$ to obtain word-level discrete speech tokens in a differentiable manner. We output a differentiable discrete index $I_t$ for $g_t$ with the straight-through estimator trick as
\begin{align}
    I_t &= I_{t, \mathrm{soft}} + \operatorname{sg}\left(I_{t,\mathrm{hard}} - I_{t, \mathrm{soft}}\right)
\end{align}
where
\begin{align}
    I_{t,\mathrm{soft}} &= \begin{bmatrix}P\left(\mathbf{e}_1 \mid \mathbf{g}_t\right) & \cdots & P\left(\mathbf{e}_V \mid \mathbf{g}_t\right)\end{bmatrix}^{\top}  \nonumber \\
    I_{t,\mathrm{hard}} &= \operatorname{One-hot}\left(\operatorname{argmax}_{v}P(\mathbf{e}_v|\mathbf{g}_t)\right)\nonumber \\
P\left(\mathbf{e}_v \mid \mathbf{g}_t\right) & =\frac{\exp \left(-\left\lVert\mathbf{g}_t-\mathbf{e}_v\right\rVert^2_2\right)}{\sum_{k \in V} \exp \left(-\left\lVert\mathbf{g}_t-\mathbf{e}_k\right\rVert^2_2\right)}
\end{align}

We further restrict the differentiable soft-pooler's allowed policy space to mitigate instability and avoid under-segmentation and over-segmentation. The first regularization loss is the word count loss $\mathcal{L_{\text{wc}}}$, which restricts the total number of boundary predictions from the CNN segmenter to be close to the pseudo-ground-truth labels, $d_t$, which are the outputs of the \emph{GradSeg+wav2bnd} segmenter used to train the initial CNN. The second loss is the word frequency loss $\mathcal{L_{\text{wf}}}$, which assumes that on average, there is one boundary frame per every $R$ consecutive speech frames. To calculate the two losses, we use hard counts and straight-through estimators:

\begin{align}
    \mathcal{L}_{\text{wc}}\left(b_1,\cdots, b_T; d_1 \cdots d_T\right) &= \left |\sum_{t = 1}^T b_t- \sum_{t = 1}^T d_t\right| \label{eqn:wc_loss} \\
    \mathcal{L}_{\text{wf}}\left(b_1,\cdots, b_T\right) &= \sum_{j=1}^{\left\lfloor \frac{T}{R}\right\rfloor} \left|\sum_{t = jR + 1}^{(j+1)R} b_t- 1\right|\label{eqn:wf_loss}
\end{align}

\begin{figure}[H]
  \centering
  \includegraphics[width=\linewidth,trim={0 0 0 0}, clip]{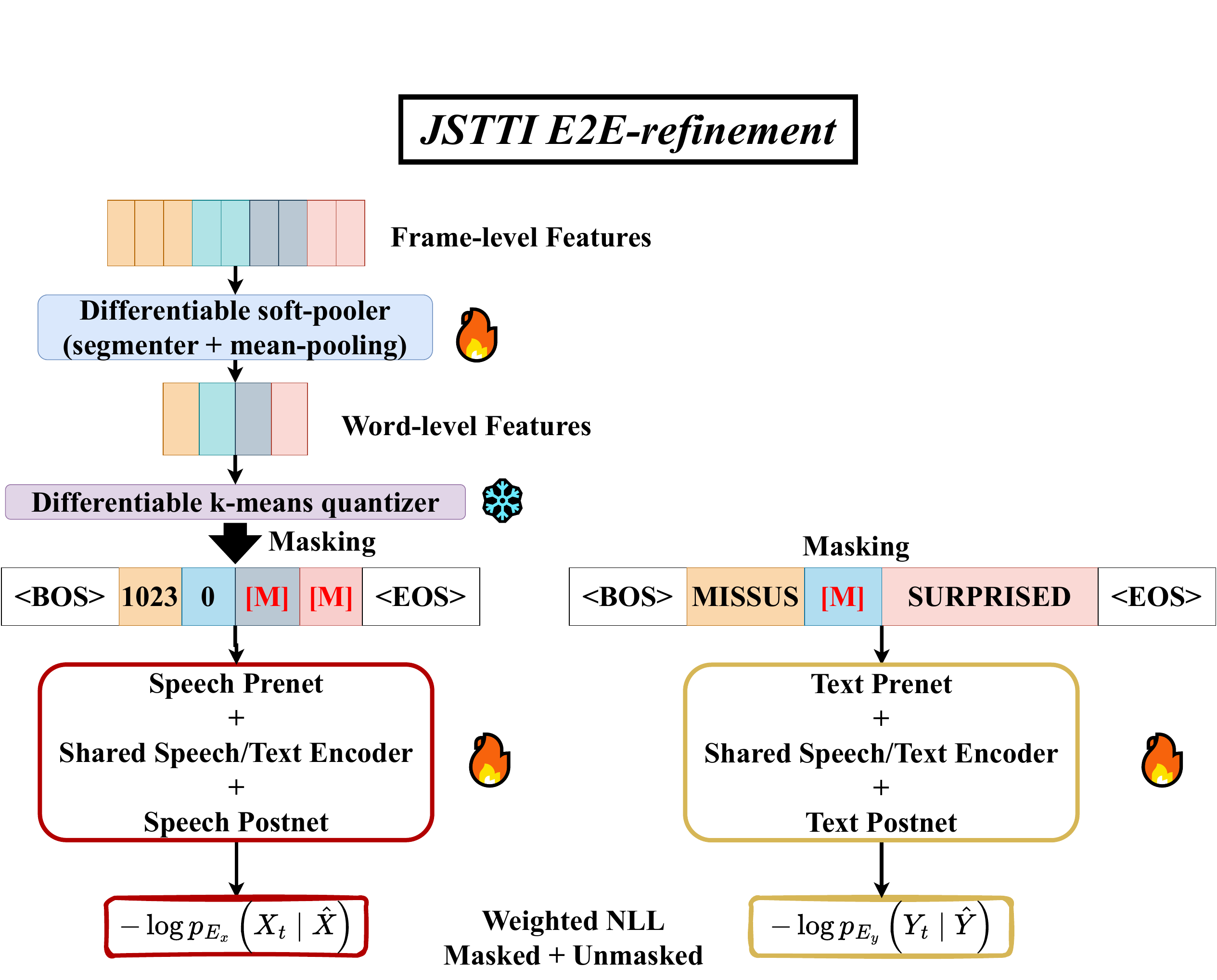}
  \caption{Illustration of JSTTI E2E-refinement. Combining the differentiable soft-pooler (CNN segmenter and differentiable mean-pooling) and the differentiable k-means quantizer in Figure \ref{fig:mean_pool_diff_bnd} with the original JSTTI Transformer encoder in Figure \ref{fig:justspeechwordt5_enc} gives the JSTTI E2E-refinement routine. The trainable CNN segmenter is initialized from cloning the word segmentation outputs of \textit{GradSeg + wav2bnd}, the frozen k-means quantizer is initialized with cluster centroids obtained in the data pre-processing step (Step 3.1 of Section \ref{subsec:extract_word_level_features}), and the parameters of the trainable JSTTI Transformer encoder are initialized from a non-E2E JSTTI encoder previously trained with speech token sequences directly extracted using \textit{GradSeg + wav2bnd} boundaries. The outputs of the differentiable k-means quantizer, before masking, serve as the target for calculating the speech loss $\mathcal{L}_{\text {wNLL}}^\mathbf{X}$.}
  \label{fig:e2e_speechwordt5}
\end{figure}

After obtaining a differentiable version of the discrete speech tokens from the vector quantizer, we mask the tokens following the masking strategy described in Section \ref{subsec:model_training}, and the unmasked version of the token sequence (with each token's gradient detached) serves as the target for the speech token-infilling loss (c.f. Eq. (\ref{eqn:speech_mle_enc})). We also increase the weight of the text token-infilling loss so that the shared Transformer cannot optimize the speech loss by excessively reducing the entropy of the k-means quantizer output without making the text loss much higher. Everything else as described in Section \ref{subsec:model_training} for the basic JSTTI model is left unchanged, and we denote this JSTTI-dependent word boundary refinement method on top of \textit{GradSeg + wav2bnd} boundaries as \emph{GradSeg + wav2bnd + JSTTI E2E-refinement}. Figure \ref{fig:e2e_speechwordt5} shows the overall pipeline, which is trained under a combination of $\mathcal{L}_{\text {wNLL}}^\mathbf{X}$, $\mathcal{L}_{\text {wNLL}}^\mathbf{Y}$, $\mathcal{L}_{\text{wc}}$ and $\mathcal{L}_{\text{wf}}$ as:
\begin{equation}\label{eqn:loss_jstti_e2e_ref}
\mathcal{L}_{\text{JSTTI-E2E-Refinement}} = \mathcal{L}_{\text {wNLL}}^\mathbf{X}+\tau\mathcal{L}_{\text {wNLL}}^\mathbf{Y} + \gamma_1 \mathcal{L}_{\text{wc}} + \gamma_2 \mathcal{L}_{\text{wf}}
\end{equation}

To obtain even more accurate word boundaries, we apply XLS-R word boundary self-training again, but this time using the unsupervised word boundaries extracted with our \textit{GradSeg + wav2bnd + JSTTI E2E-refinement} method as targets for frame-level XLS-R fine-tuning. In this iteration, we also train from a larger checkpoint with 1B parameters. We extract a new set of word-level discrete speech token sequences for re-training the JSTTI Transformer encoder, and denote this setting as \emph{GradSeg + wav2bnd + JSTTI E2E-refinement + wav2bnd-large}.

\subsection{Word-level Pseudo-text Self-training}\label{subsec:self_train_trans}
Inspired by previous work on phone-level unsupervised ASR \citep{Chen2019,Baevski2021-wav2vec-u,Liu2022-wav2vecu2.0}, we apply \emph{self-training on top of pseudo-transcripts}. Specifically, we fine-tune a pre-trained HuBERT-Large checkpoint under the word-level Connectionist Temporal Classification (CTC) loss, and instead of using word-level ground-truth transcripts, we use the word-level pseudo-transcripts predicted by the JSTTI model. We use greedy decoding to obtain the final predicted transcripts from the fine-tuned HuBERT-Large model.

\section{Experimental Methods}\label{sec:method}

Experiments described in this paper are synthetic unsupervised ASR experiments, designed to explore the behavior of the proposed algorithms on a series of tasks with a variety of difficulty levels.  None of the synthetic tasks described in this article constitute fully natural, deployable unsupervised ASR, but the experimental tasks we describe are designed to gradually approach that goal, and to clarify attributes of model behavior that need to be carefully managed in our pursuit of eventual in-the-wild deployment for real-world unsupervised ASR.  Section~\ref{subsec:curation} describes a set of synthetically curated speech corpora, designed to permit a careful study of the relationship between vocabulary size and WER.   Section~\ref{subsec:settings} describes hyperparameter settings used in the training and testing phases of all experiments, while Section~\ref{subsec:baselines} describes two baseline systems, one designed for whole-word unsupervised ASR using the PUSM criterion (positional unigram and skipgram matching) and another designed for character-based unsupervised ASR, a direct alternative to whole-word unsupervised ASR for letter-based languages, using the REBORN reinforcement-learning-based unsupervised training pipeline.

\subsection{Curating the Corpus}\label{subsec:curation}
\begin{figure}[H]
  \centering
  \includegraphics[width=0.8\linewidth,trim={0 0 0 0},clip]{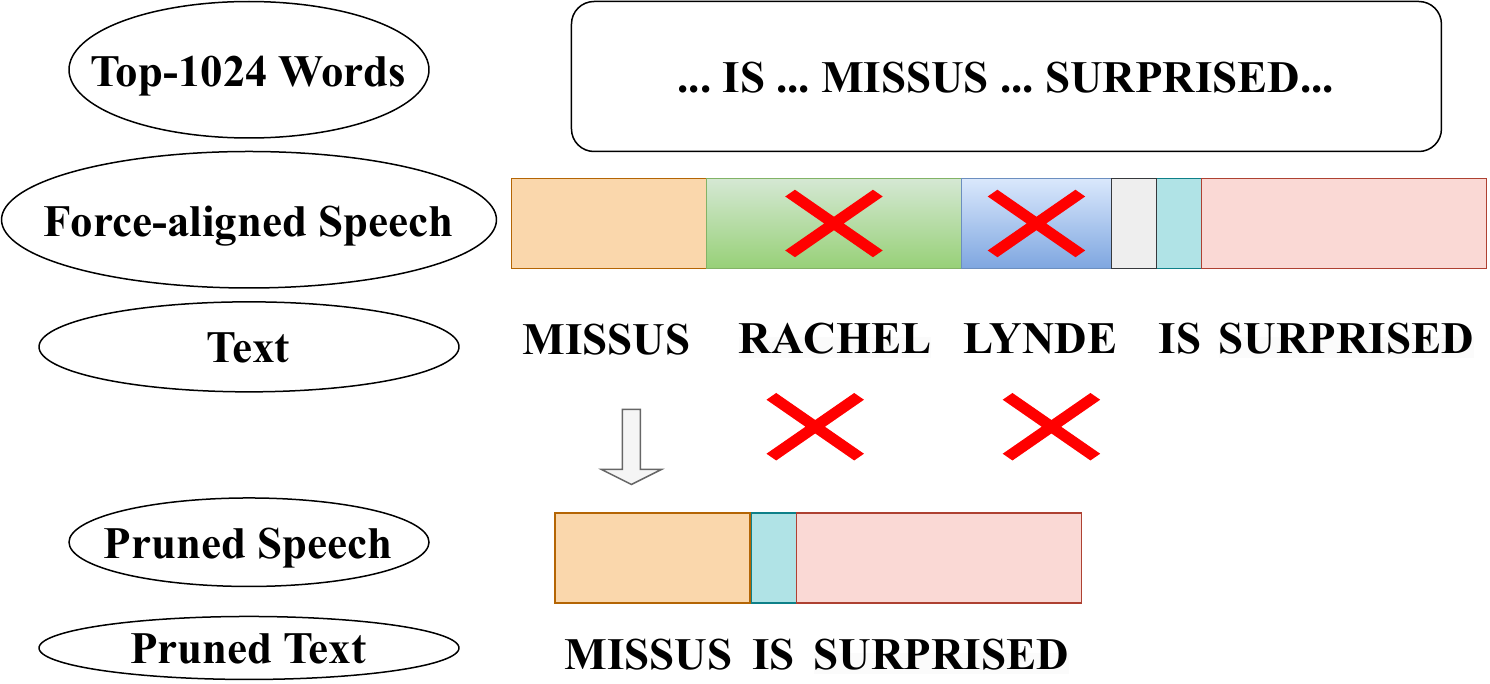}
  \caption{Curation process for our 1024-word corpus. we keep only the top 1024 words in the clean subset of LibriSpeech using forced alignments and respectively pre-process the transcripts.}
  \label{fig:curation}
\end{figure}

We conduct experiments on the clean subset of the LibriSpeech corpus \citep{Panayotov15-LibriSpeech}. Due to the extended tail distribution of English words in the LibriSpeech corpus, we simplify the problem by \emph{pruning the corpus so that only the top-K high-frequency words exist} in both the unpaired speech and text. The curation process for $K=1024$, the setting we explore in Sections \ref{sec:result} and \ref{sec:segmentation}, is shown in Figure \ref{fig:curation}, where we concatenate force-aligned speech segments corresponding to high-frequency words. In Section \ref{sec:extend_vocab}, we repeat the same process for $K=2048$ and $K=4096$ to test our proposed word-level unsupervised ASR model across different vocabulary sizes.

The unpaired text corpus is constructed from the corresponding speech transcripts by removing low-frequency words. This is an idealist setting where unpaired speech and text distributions are ``exact matches'' of each other, given a perfect modality converter, i.e., where the divergence in Eq. (\ref{data_assumption}) is zero, and $\mathbf{P} \left(Y \right) = \mathbf{P^{\prime}} \left(Y\right)$. In future work, we will explore truly non-parallel settings.

\subsection{Experiment Settings}
\label{subsec:settings}

All our experiments are carried out on the curated $K$-word LibriSpeech-clean subset as described in Section \ref{subsec:curation}. Except for Section \ref{sec:extend_vocab}, where we specifically examine $K=2048$ and $K=4096$, we otherwise focus only on the case where $K = 1024$. We keep 90\% of the speech utterances and their corresponding transcript, with the pairing relationship broken up for unsupervised training, and the rest serve as the evaluation set for reporting our results unless denoted otherwise. The speakers in the training and evaluation set are chosen to be distinct. 

All JSTTI models use a 2-layer Transformer encoder architecture, with a model dimension of 768, a position-wise feedforward dimension of 3072, and 12 attention heads. All JSTTI models are trained under an Adam optimizer with an initial learning rate of $0.0002$, initial warm-up, and polynomial decay. All JSTTI models trained without E2E-refinement use a uniform weight of $1$ between $\mathcal{L}_{\text {wNLL}}^\mathbf{X}$  and $\mathcal{L}_{\text {wNLL}}^\mathbf{Y}$, and we set $\lambda = 0.5$ in both Eq. (\ref{eqn:speech_mle_enc}) and Eq. (\ref{eqn:text_mle_enc}). The JSTTI models with E2E-refinement include an additional CNN segmenter, for which we use five 1D convolution layers, with kernel sizes 11, 9, 7, 5, and 3 from layer one to layer five, respectively, and ReLU activation after each layer. We set $\gamma_1 = 500$ and $\gamma_2 = 500,000$ in Eq. (\ref{eqn:loss_jstti_e2e_ref}), and we increase $\tau$ to 10 during the \textit{JSTTI E2E-refinement} stage.

As an early exploration, we use paired speech-text data for recording checkpoints. This set of paired data, which we further denote as the validation set, amounts to 10\% of the entire training split.  For the JSTTI Transformer encoder model, we directly calculate the average validation word error rates (WER) over all the validation utterances to log the best model checkpoint. The validation WERs are calculated by feeding the text output layer with speech encoder representation from the second encoder layer. This can be viewed as a surrogate supervised metric, as the evaluation routine uses the speech representation from the first encoder layer instead. We leave a detailed exploration of unsupervised validation metrics for word-level unsupervised ASR to future work. We further note that all reported results are calculated on the evaluation, \emph{not} the validation set.

We apply early stopping once the validation metrics converge. For JSTTI models, this usually takes less than 7000 epochs if we train from scratch. However, since we keep the speech quantization model unchanged (for a fixed vocabulary size setting) in Step 3 of the data pre-processing step (c.f Section \ref{subsec:extract_word_level_features}), we can initialize a JSTTI model trained with more accurate word boundaries with a checkpoint previously trained on top of less accurate word boundaries. Due to this initialization, all JSTTI models, except the first one trained for a given vocabulary size, converge much faster, usually less than 1600 epochs. For JSTTI E2E-refinement (c.f. Section \ref{subsec:unsup_bnd_and_refinement}), we stop before the 700$^{\text{th}}$ epoch.

\subsection{Baselines}
\label{subsec:baselines}

\cite{wang-etal-2023-unsup-speech2sign} have studied the word-level unsupervised ASR task with the PUSM loss under an ideal assumption that word-level forced alignments are available for pooling word-level features. Here, to compare with our proposed JSTTI models, we train several PUSM models using the same discrete speech token sequences used for training the JSTTI models on the same 1024-word corpus curated as in Section \ref{subsec:curation}. This extends the PUSM models to cases with only unsupervised word boundaries for pooling word-level features. We further modify the update strategy of the original distribution matching criterion to allow us to perform accurate unigram and skipgram matching on a large effective batch without running into GPU memory issues, given that the model parameters and the position unigram and skipgram parameters fit. \ref{appendix:modified_pusm} explains how we modify this and showcases that the modification results in a stronger baseline. All PUSM models are trained under an Adam optimizer with a learning rate of $0.4$, with no learning rate decay. For the PUSM models, we directly calculate the validation WER to log the model checkpoint over 3000 epochs.

To further showcase that our JSTTI model yields lower WERs than any existing training algorithm when a pronunciation model is \emph{not} available for learning an unsupervised ASR, we train a REBORN unsupervised ASR system using English characters as output units (denoted as \emph{REBORN-char}) on our curated 1024-word corpus under the same ideally paired speech-text setting described in Section \ref{subsec:curation}.  Except for the choice of output/text units, the REBORN-char model is trained in the same fashion explained in Section \ref{background:wav2vecu_REBORN} and in~\cite{Tseng2024-reborn}. Starting from an initial wav2vec-U model trained with characters as text units, we perform three iterations of REBORN training, where each iteration involves updating the segmentation model and the phone prediction model in two stages. To keep the comparison with our JSTTI-ENC model fair, we calculate the character error rates on the validation set as the validation metric for both stages of a REBORN training iteration. 

In addition to the REBORN-char model, we train a phone-level REBORN model (REBORN-phn). While not directly comparable to our JSTTI model, which is G2P-free, the phone-level wav2vec-U + REBORN model establishes another top-line result for unsupervised ASR on the same 1024-word corpus had a G2P system been available.

For both REBORN-char and REBORN-phn, We apply HMM self-training after obtaining the pseudo-transcripts from the final REBORN iteration. We subsequently use the decoded word pseudo-targets from the HMM to fine-tune a pre-trained HuBERT-Large checkpoint under the same word-level CTC loss described in Section \ref{subsec:self_train_trans}. 

While we could potentially establish another baseline by training a whole-word unsupervised ASR system under the REBORN pipeline, we find it impossible for a wav2vec-U system to converge on word-level targets even as we provide word-level speech features pooled with forced alignments as input. A similar issue has been observed in \cite{wang-etal-2023-unsup-speech2sign} when using word-level targets in their forced-alignment-based unsupervised ASR and unsupervised speech-to-sign tasks. As the wav2vec-U system was initially designed to work with phone-level text, its specific design considerations may not extend to the whole-word scenarios.

\section{Results}\label{sec:result}

This section describes key results of this work, as tested using the smallest synthetic speech and text corpora, with 1024 distinct word types represented.  Three categories of model structure variables are systematically varied: segmentation, training criterion, and self-training refinement.  The best segmentation is achieved by applying four segmentation algorithms, one after the other, each one teaching the next: \emph{GradSeg} initial segmentations are used to train \emph{wav2bnd}, which is then used to train a JSTTI end-to-end system including CNN segmenter and Transformer encoder, and segmentations obtained from the CNN segmenter within are used to train a \emph{wav2bnd-large} segmenter.  The best training criterion is JSTTI.  The best post-processing strategy uses JSTTI to train a HuBERT-Large model using the word-level CTC pseudo-text training criterion.  The performance of whole-word unsupervised ASR does not outperform previously published phone-based unsupervised ASR, but the proposed algorithms for whole-word unsupervised ASR outperform, by a large margin, the application of any prior model to the G2P-free unsupervised ASR task. 
Table~\ref{tab:hubert_word_error_rates} shows these results, which are narrated in the remainder of this section.

\begin{table}[H]
\caption{\textbf{Main Results:}
WER of unsupervised ASR on \textbf{a vocabulary of
1024 words}.
\textbf{Segmentation,} fully unsupervised: word boundaries from VG-HuBERT~\citep{peng2022-vghubert},GS=GradSeg~\citep{fuchs2023-gradseg}, w2b=wav2bnd~\citep{Algayres23-xlsr-wav2bnd}, w2bL=wav2bnd-Large, and the proposed JE2E=JSTTI with end-to-end
training of segmenter and recognizer; character-level boundaries refined by REBORN-char~\citep{Tseng2024-reborn}.
Partially supervised:
Whole-word Forced Alignment using ground-truth transcriptions,
REBORN-phn using a ground-truth G2P.
\textbf{Training criteria}: whole-word JSTTI, whole-word PUSM~\citep{wang-etal-2023-unsup-speech2sign}, REBORN with character or phone targets ~\citep{Tseng2024-reborn}. \textbf{Student} during self-training: 
HMM or HuBL=HuBERT-Large or HMM + HuBL.
Lowest WER of any fully unsupervised, G2P-free, model 
is shown in bold.  Lowest WER of any partially supervised
model is underlined.}\label{tab:hubert_word_error_rates}
\centering
\begin{tabular}{lllc}
\toprule
Segmentation & Training & Student & WER($\downarrow$) \\
\midrule
\midrule
VG-HuBERT & PUSM & None & 58.73 \\
VG-HuBERT & JSTTI & None & 47.99 \\
\midrule
\midrule
GS & PUSM & None & 50.84 \\
GS+w2b & PUSM & None & 43.08 \\
GS+w2b+JE2E & PUSM & None & 33.71 \\
GS+w2b+JE2E+w2bL  & PUSM & None & 31.91  \\
GS+w2b+JE2E & PUSM & HuBL & 29.40 \\
GS+w2b+JE2E+w2bL & PUSM & HuBL & 28.87 \\
\midrule
\midrule
REBORN-char & REBORN & HMM & 35.49 \\
REBORN-char & REBORN & HMM + HuBL & 26.26 \\
\midrule
\midrule
GS & JSTTI & None & 44.76 \\
GS+JE2E & JSTTI & None & 42.02 \\
GS+w2b  & JSTTI & None & 38.08 \\
GS+w2b+w2bL & JSTTI & None & 36.43 \\
GS+w2b+JE2E & JSTTI & None & 29.39 \\
GS+w2b+JE2E+w2bL & JSTTI & None & 26.51 \\
GS+w2b+JE2E & JSTTI & HuBL & 21.65 \\
GS+w2b+JE2E+w2bL & JSTTI & HuBL & \textbf{20.68} \\
\midrule
\midrule
Forced Alignment  & PUSM & None & 20.89  \\
Forced Alignment & PUSM & HuBL & 18.33 \\
Forced Alignment & JSTTI & None & 18.06 \\
Forced Alignment & JSTTI & HuBL & 12.99 \\
\midrule
REBORN-phn & REBORN & HMM & 9.04 \\
REBORN-phn  & REBORN &HMM + HuBL & \underline{5.30}\\
\bottomrule
\end{tabular}
\end{table}

Table \ref{tab:hubert_word_error_rates} demonstrates that off-the-shelf unsupervised word segmenters such as VG-HuBERT and GradSeg, when used for training and evaluation of the JSTTI models, yield very high word error rates, but that our proposed iterative refinement training pipeline is able to achieve significantly lower word error rates.  For whole-word unsupervised ASR, regardless of the 
ASR training criterion or self-training student model, 
the best results are achieved by combining all three steps to extract word-level speech tokens (initial \textit{wav2bnd} training, followed by \textit{JSTTI E2E-refinement}, followed by another run of \textit{wav2bnd-large}), as described in Section \ref{subsec:unsup_bnd_and_refinement}.\footnote{We use \textit{sclite} and \textit{sc\_stats} to compare the transcripts from the fully unsupervised whole-word JSTTI models as we successively improve the segmentation of the speech tokens from \textit{GS} to \textit{GS+w2b}, and then to \textit{GS+w2b+JE2E} and finally to \textit{GS+w2b+JE2E+w2bL}. We conclude that the improvement between the transcripts from each pair of successive steps are all significant with $p<0.001$. Further, after applying HuBL pseudo-text self-training, \textit{GS+w2b+JE2E+w2bL} is still significantly better than \textit{GS+w2b+JE2E} with $p<0.001$.}

Table~\ref{tab:hubert_word_error_rates} further demonstrates the effectiveness of pseudo-text self-training (c.f. Section \ref{subsec:self_train_trans}). We start from a pre-trained HuBERT-Large checkpoint with 300M parameters and fine-tune the checkpoint under the CTC loss directly calculated on top of word token pseudo-targets (denoted as word-CTC). We repeat the same self-training routine on top of pseudo-transcripts obtained from three JSTTI-ENC models in Table \ref{tab:hubert_word_error_rates} that are most insightful, namely, the model trained with speech tokens extracted from forced alignments (top-line), the model trained with speech tokens extracted from \textit{GradSeg + wav2bnd + JSTTI E2E-refinement + wav2bnd-large} and the model trained with speech tokens extracted from \textit{GradSeg + wav2bnd + JSTTI E2E-refinement}. 
As expected, pseudo-text self-training further reduces the WERs of all three selected settings. With self-training, we obtain a final WER of 20.68\% without resorting to forced alignments. 

Comparing the WERs of our newly proposed JSTTI models and the results of the (modified) PUSM models in Table \ref{tab:hubert_word_error_rates}, we see that the unsupervised word-level ASR system trained under our JSTTI encoder model yields significantly lower WERs, which we attribute to the intermediate encoder layer's capability to capture a shared speech-text hidden space emerging from unimodal masked reconstruction tasks under a shared encoder architecture.  Table \ref{tab:hubert_word_error_rates} further shows that pseudo-text word-CTC self-training increases the gap between JSTTI and PUSM models. This means that the pseudo-transcripts provided by the JSTTI model exhibit transcription errors that are easier to correct under comparable settings.\footnote{We use \textit{sclite} and \textit{sc\_stats} to compare the transcripts from the best fully unsupervised whole-word JSTTI model after HuBL pseudo-text self-training (row 18) and the best fully unsupervised whole-word PUSM model after HuBL pseudo-text self-training (row 8), and conclude that the transcripts from the JSTTI model are indeed significantly better with $p<0.001$.}

Table \ref{tab:hubert_word_error_rates} shows the WERs of REBORN-char and REBORN-phn models after HMM self-training (HMM) and then followed by HuBERT-Large self-training (HMM + HuBL).  Comparing with the best results in Table \ref{tab:hubert_word_error_rates} obtained with unsupervised word boundaries, we see that the unsupervised WERs from the REBORN-char models are significantly worse than those from our word-level JSTTI models, both before and after HuBERT-Large self-training. Our JSTTI-based word-level unsupervised ASR optimizes the unsupervised learning criterion in Eq. (\ref{uns_criterion}) on the whole-word level, using unsupervised word boundaries that are easier to segment from speech. In contrast, the REBORN training iterations (before any self-training) optimize Eq. (\ref{uns_criterion}) on top of character targets with vague boundaries that are harder to segment. Even after HMM self-training, the pseudo-transcripts of the REBORN-char model contain more errors than those from our best-performing JSTTI model trained with unsupervised word boundaries, making it harder for the HuBERT-Large model to correct those errors during word-level CTC self-training.\footnote{We use \textit{sclite} and \textit{sc\_stats} to compare the transcripts from the best fully unsupervised whole-word JSTTI model after HuBL pseudo-text self-training (row 18) and the fully unsupervised REBORN-char model after HuBL pseudo-text self-training (row 10), and conclude that the transcripts from the JSTTI model are indeed significantly better with $p<0.001$.}

On the other hand, the REBORN-phn models perform significantly better than any unsupervised ASR models presented so far that do not use a G2P, even those top-lines using partially supervised word boundaries from forced alignments. Given a perfect G2P, phone transcripts almost unambiguously specify how the speaker should pronounce the text. Hence, the mutual information between phonemicized text distribution and the distribution of speech utterances is much higher than the mutual information between other higher-level text forms and speech. This makes optimizing Eq. (\ref{uns_criterion}) more straightforward for learning an unsupervised ASR. Furthermore, learning unsupervised phone boundaries is easier than learning word boundaries \citep{Bhati2022-scpc,CuervoL2022-hcpc}. However, training an accurate G2P or collecting an accurate phone lexicon may not be possible for all languages except for a few high-resource ones. Hence, it is imperative to investigate unsupervised ASR in a G2P-free setting, and our proposed whole-word JSTTI model is the most performant choice in this setting.

\section{Relationship Between WER and Segmentation Performance}
\label{sec:segmentation}

Section \ref{sec:result} demonstrated a large variation in WER as a function of the type of algorithm used to segment the audio into word-sized units.  This section further explores the relationship between segmentation accuracy and whole-word unsupervised recognition error rate.
To evaluate the quality of the word boundaries, we use the standard boundary token F1 score \citep{Dunbar2017} with a tolerance of 20ms, together with token precision and token recall. A token in an utterance is only marked as correctly discovered when both of its discovered boundaries are within the tolerance windows of a word's true start and end boundaries in the time-aligned transcription.

\begin{table}[H]
\caption{Boundary Token Precision (T-P) / Recall (T-R) / F-1 (T-F1) in percentage (\%) (with 20ms tolerance) for the unsupervised word boundaries obtained and used to pool word-level speech sequences for training JSTTI models. The WERs are copied from the corresponding JSTTI models (without a student for self-training) in Table \ref{tab:hubert_word_error_rates}. Rows 2-5 correspond to the iterative boundary refinement routine in Section \ref{subsec:unsup_bnd_and_refinement}, and rows 6 and 7 serve as ablations. The best result is shown in bold.}\label{tab:hubert_word_boundary_metrics}
\centering
\begin{tabular}{llcccc}
\toprule
Segmentation & Training & WER($\downarrow$) & T-P($\uparrow$) & T-R($\uparrow$) & T-F1($\uparrow$) \\
\midrule
\midrule
VG-HuBERT & JSTTI  & 47.99 & 24.94 & 21.18 & 22.91 \\
\midrule
\midrule
GS & JSTTI & 44.76 & 36.21& 36.51 & 36.36\\
GS+w2b  & JSTTI & 38.08 & 44.59 & 44.11 & 44.35\\
GS+w2b+JE2E & JSTTI  & 29.39 & 63.17 & 63.56 & 63.36 \\
GS+w2b+JE2E+w2bL & JSTTI & \textbf{26.51} & \textbf{64.47} & \textbf{64.67} & \textbf{64.57} \\
\midrule
\midrule
GS+w2b+w2bL & JSTTI  & 36.43 & 44.74 & 44.97 & 44.85 \\
\midrule
\midrule
GS+JE2E & JSTTI &  42.02 & 37.94 & 39.03 & 38.48 \\
\bottomrule
\end{tabular}
\end{table}
By comparing the WERs in Table \ref{tab:hubert_word_error_rates} and the word boundary metrics in Table \ref{tab:hubert_word_boundary_metrics}, we see that more accurate word boundaries usually lead to reduced WERs, especially as we gradually improve the word boundaries obtained from GradSeg with \textit{wav2bnd}, \textit{JSTTI E2E-refinement} and \textit{wav2bnd-large} (c.f. Section \ref{subsec:unsup_bnd_and_refinement}). The boundary metrics in rows 3-5 compared to those in rows 6 and 7 allow us to further verify that obtaining better word boundaries with XLS-R boundary self-training (\textit{wav2bnd}) before our proposed JSTTI E2E-refinement routine is important, as jointly finetuning the CNN segmenter and the JSTTI model pre-trained using GradSeg boundaries (\textit{GradSeg + JSTTI E2E-refinement}) does not improve the boundary metrics as significantly. We also see that simply doing another run of boundary XLS-R self-training on top of the boundary predictions from \textit{GradSeg + wav2bnd} (\textit{GradSeg + wav2bnd + wav2bnd-large}) does not lead to a big improvement for the boundary metrics. We conclude that our proposed iterative boundary refinement routine allows us to extract word-level segmental information more accurately from frame-level speech features, which is crucial in terms of reducing the WERs of our word-level unsupervised JSTTI models when we only have access to unsupervised word boundaries.

We further wish to examine the following scenario pertaining to a JSTTI model trained with unsupervised word boundaries and the quality of word segmentation during inference: if the evaluation data are more accurately segmented than the training data, such as when a more accurate unsupervised word segmenter becomes available in the future, can a trained JSTTI model take advantage of the improved segmentation? Table \ref{tab:old_model_new_seq_1} shows WERs achieved by models trained using discrete speech sequences obtained from worse boundaries (four different word segmentation methods as previously listed in Table \ref{tab:hubert_word_error_rates}) but evaluated using speech token sequences obtained from better boundaries (i.e., from the discrete speech sequences segmented with the unsupervised word segmenter denoted as \textit{GradSeg + wav2bnd + JSTTI E2E-refinement + wav2bnd-large}).

\begin{table}[H]
\caption{WERs when evaluating the JSTTI models trained using speech segmentation from a specific unsupervised word segmenter with more accurately segmented speech tokens (GS+w2b+JE2E+w2bL=\textit{GradSeg + wav2bnd + JSTTI E2E-refinement + wav2bnd-large} in all cases). The last row is copied from Table \ref{tab:hubert_word_error_rates} for comparison.}\label{tab:old_model_new_seq_1}
\centering
\begin{tabular}{lcc}
\toprule
Segmentation Used During Training & WER($\downarrow$) \\
\midrule
\midrule
GS & 28.29 \\
\hline
GS+w2b & 28.00\\
\hline
GS+w2b+JE2E & 26.85 \\
\hline
GS+w2b+JE2E+w2bL  & 26.51 \\
\bottomrule
\end{tabular}
\end{table}

We see that just by replacing word-level speech tokens used for evaluation with those obtained with more accurate word boundaries, the WERs of the best and the worst JSTTI models are within 6-7\% of each other. We conclude that JSTTI models are not too sensitive to bad word boundaries during training.

\section{Discussion: Larger Whole-word Vocabularies}\label{sec:extend_vocab}

Section \ref{subsec:curation} describes curation of a corpus with a fixed-$K$ vocabulary size from the clean subset of LibriSpeech \citep{Panayotov15-LibriSpeech}. While most of our experiments focus on $K=1024$, this section curates two additional settings with $K=2048$ and $K=4096$ on the same clean subset of LibriSpeech. The \emph{GradSeg} and \emph{wav2bnd} segmenters are re-trained for these two additional vocabulary settings. We extract speech token sequences from mean-pooled word-level speech features following Section \ref{subsec:extract_word_level_features}, where we re-train the k-means quantizer for the $K=2048$ vocabulary size by setting the cluster number to $k=2048$ and similarly by setting the cluster number to $k=4096$ for the $K=4096$ vocabulary setting. We keep the hyperparameters for training the JSTTI model the same as those used for the $K=1024$ setting, except that during the JSTTI E2E-refinement stage (on top of \textit{GradSeg + wav2bnd}), we increase the word duration parameter $R$ in Eq. (\ref{eqn:wf_loss}) for $K=2048$ and then $K=4096$, as the average duration of a word is longer in these two settings. We then extract the word boundaries from the CNN segmenters in the E2E JSTTI models and separately perform \textit{wav2bnd-large} boundary refinement runs for the utterances in $K=2048$ and $K=4096$. In Tables \ref{tab:2048_from_scratch_wer} and \ref{tab:4096_from_scratch_wer}, we display the WERs on the evaluation utterances using the JSTTI models trained for $K=2048$ and $K=4096$ under the iterative refinement routine (c.f. Section \ref{subsec:unsup_bnd_and_refinement}) together with the top lines obtained using forced alignments.

\begin{table}[H]
\caption{WERs for the JSTTI models trained using discrete speech token sequences
extracted with different word segmentation methods when the \textbf{vocabulary size} $\mathbf{K=2048}$. The top line obtained with forced alignments has been
underlined, and the best result obtained with unsupervised word boundaries is shown in
bold. \textbf{Segmentation}=word segmentation algorithm. \textbf{Student}=student learner in self-training.}\label{tab:2048_from_scratch_wer}
\centering
\begin{tabular}{llc}
\toprule
Segmentation & Student & WER($\downarrow$) \\
\midrule
\midrule
GS+w2b & None & 35.48\\
\hline
GS+w2b+JE2E & None & 28.46 \\
\hline
GS+w2b+JE2E+w2bL  & None & 26.47 \\
\hline
GS+w2b+JE2E+w2bL  & HubL & \textbf{20.21} \\
\midrule
\midrule
Forced Alignment & None & 16.57 \\
\hline
Forced Alignment & HuBL & \underline{12.01} \\
\bottomrule
\end{tabular}
\end{table}

\begin{table}[H]
\caption{WERs for the JSTTI models trained using discrete speech token sequences
extracted with different word segmentation methods when the \textbf{vocabulary size}  $\mathbf{K=4096}$. The top-line obtained with forced alignments has been
underlined, and the best result obtained with unsupervised word boundaries is shown in
bold. All results are obtained directly from the hypotheses of corresponding whole-word JSTTI models (i.e., \textbf{Student} = None)}\label{tab:4096_from_scratch_wer}
\centering
\begin{tabular}{lcc}
\toprule
Segmentation & Student & WER($\downarrow$) \\
\midrule
\midrule
GS+w2b & None & 38.97\\
\hline
GS+w2b+JE2E & None & 32.40 \\
\hline
GS+w2b+JE2E+w2bL  & None & \textbf{31.60} \\
\midrule
\midrule
Forced Alignment & None & \underline{19.45} \\
\bottomrule
\end{tabular}
\end{table}

We see that when increasing the vocabulary size of our curated corpus from $K=1024$ to $K=2048$, we do not experience any performance degradation in terms of WER; most of the time, we even get some performance improvement, likely because the inclusion of additional words makes the utterances contextually more natural. 
Pseudo-text self-training (c.f. Section \ref{subsec:self_train_trans}) provides additional 
benefits when applied to either the best fully 
unsupervised approach (GS+w2b+JE2E+w2bL)
or when applied to the partially supervised approach
(Forced Alignment).
Unfortunately, similarly low WER is not observed for the
$K=4096$ case. While both the \textit{JSTTI E2E-refinement} routine and the subsequent \textit{wav2bnd-large} boundary refinement bring about positive reductions in WERs, the effects are relatively more subtle than their counterparts for $K=1024$ and $K=2048$. This results in a substantially higher WER when unsupervised word boundaries are used for the $K=4096$ setting. We also see a WER degradation for the case when word-level forced alignments are used, further showing that as we further increase the vocabulary size beyond $K=2048$, optimizing Eq. (\ref{uns_criterion}) on the whole-word level becomes more challenging for the JSTTI model. Hence, for $K=4096$, we explore if we could mitigate the performance degradation from the whole-word JSTTI model before we apply pseudo-text self-training.

Suppose that we have access to the best unsupervised word segmenter (GradSeg + wav2bnd + JSTTI E2E-refinement + wav2bnd-large) trained for the $K=1024$ vocabulary setting, together with the corresponding JSTTI model and the k-means model using for quantizing the speech features in the 1024-word corpus. In that case, we can efficiently adapt the trained JSTTI model to the $K=4096$ vocabulary setting: all of the encoder parameters, together with the parameters within the text encoder pre-net and text encoder post-net corresponding to the top 1024 words with the highest frequency could be directly copied from the JSTTI model trained on the 1024-word corpus. We find that simply tuning the prominence threshold for the \textit{wav2bnd-large} model trained on the 1024-word corpus allows us to segment the words in the 4096-word corpus accurately. To avoid completely re-learning the speech token representations when we perform the above initialization, we modify the k-means clustering routine when training it on the 4096-word corpus, and Algorithm \ref{algo:fckmeans}, which is one variant of FC-Kmeans \citep{Ay2023-fckmeans}, explains the modification.

\begin{algorithm}[H]
\caption{Fixed-centroids k-means on 4096-word corpus}
\label{algo:fckmeans}
\textbf{Input:} 
\begin{itemize}
    \item $\{f_1, f_2, f_3, \dots, f_n\}$: A collection of word-level speech features from the 4096-word corpus.
    \item $\{c_1, c_2, c_3, \dots, c_{1024}\}$: Set of word-level cluster centroids trained on the 1024-word corpus.
\end{itemize}
\textbf{Output:} 
\begin{itemize}
    \item $\{C_1, C_2, C_3, \dots, C_{4096}\}$: Set of word-level cluster centroids trained on the 4096-word corpus, with $C_i = c_i$ for $i = 1, \dots, 1024$.
\end{itemize}

\begin{algorithmic}[1]
\State \textbf{Phase-I:} Train a k-means model with $k = 4096$ on $\{f_1, f_2, \dots, f_n\}$, resulting in cluster centroids $\{\mu_1, \mu_2, \dots, \mu_{4096}\}$.
\State \textbf{Phase-II:}
\For{$i = 1, \dots, 4096$}
    \State Calculate $D_i = \frac{1}{1024} \sum_{j=1}^{1024} \left\| \mu_i - c_j \right\|^2$
\EndFor
\State Initialize $C_1, C_2, \dots, C_{1024} \gets c_1, c_2, \dots, c_{1024}$.
\State Initialize $C_{1025}, C_{1026}, \dots, C_{4096} \gets \mu_j\text{'s with the top-3072 } D_j\text{'s}$.
\State Update k-means cluster centroids but fixing the first 1024 centroids:
\Repeat
    \For{$i = 1, \dots, 4096$}
        \State Set $G_i \gets \varnothing$
    \EndFor
    \For{each $f \in \{f_1, f_2, \dots, f_n\}$}
        \State $i \gets \operatorname{argmin}_{i = 1, \dots, k} \left\| f - C_i \right\|^2$
        \State Add $f$ to $G_i$
    \EndFor
    \For{$i = 1025, \dots, 4096$}
        \State $C_i \gets \frac{1}{|G_i|} \sum_{f \in G_i} f$
    \EndFor
\Until{convergence}
\end{algorithmic}
\end{algorithm}
The rest of the parameters in the JSTTI model for $K=4096$ that are not in the previously trained JSTTI model for $K=1024$ (e.g., those input and output embeddings corresponding to new/unseen speech and text tokens) are randomly initialized. Table \ref{tab:4096_from_1024_wer} reports the results, where the top-line models trained with Forced Alignment are additionally initialized with the fully unsupervised checkpoint in row 1. Comparing Table \ref{tab:4096_from_1024_wer} and Table \ref{tab:4096_from_scratch_wer}, we see that progressively fine-tuning the JSTTI model from the JSTTI model already trained for a smaller vocabulary size yields superior results, and that pseudo-text self-training using a HuBERT-Large (HuBL) student further improves WER. Comparing Table \ref{tab:4096_from_1024_wer}, Table \ref{tab:2048_from_scratch_wer}, and Table \ref{tab:hubert_word_error_rates}, we conclude that this progressive training method allows us to extend the vocabulary size to $K=4096$ with just a small performance penalty compared to $K=1024$ and $K=2048$.

\begin{table}[H]
\caption{Improved WERs for the JSTTI models trained on the 4096-word corpus. The JSTTI model in the first row is initialized with the best JSTTI model trained on the 1024-word corpus (GS+w2b+JE2E+w2bL-1024, the same segmenter used for obtaining the best fully unsupervised result in Table \ref{tab:hubert_word_error_rates}). The best unsupervised word segmenter trained on the 1024-word corpus is used to segment and obtain word-level speech tokens for the 4096-word corpus (c.f. Section \ref{subsec:extract_word_level_features}). The top-line results obtained with Forced Alignment on the 4096-word corpus are further fine-tuned from the fully unsupervised JSTTI models.}\label{tab:4096_from_1024_wer}
\centering
\begin{tabular}{llc}
\toprule
Segmentation & Student & WER($\downarrow$) \\
\midrule
\midrule
GS+w2b+JE2E+w2bL-1024  & None & 28.82 \\ & HuBL & \textbf{22.87}\\
\midrule
\midrule
Forced Alignment & None & 17.57 \\
& HubL & \underline{13.02} \\
\bottomrule
\end{tabular}
\end{table}

\begin{figure}[H]
  \centering
  \includegraphics[width=\linewidth,trim={0 0 0 0},clip]{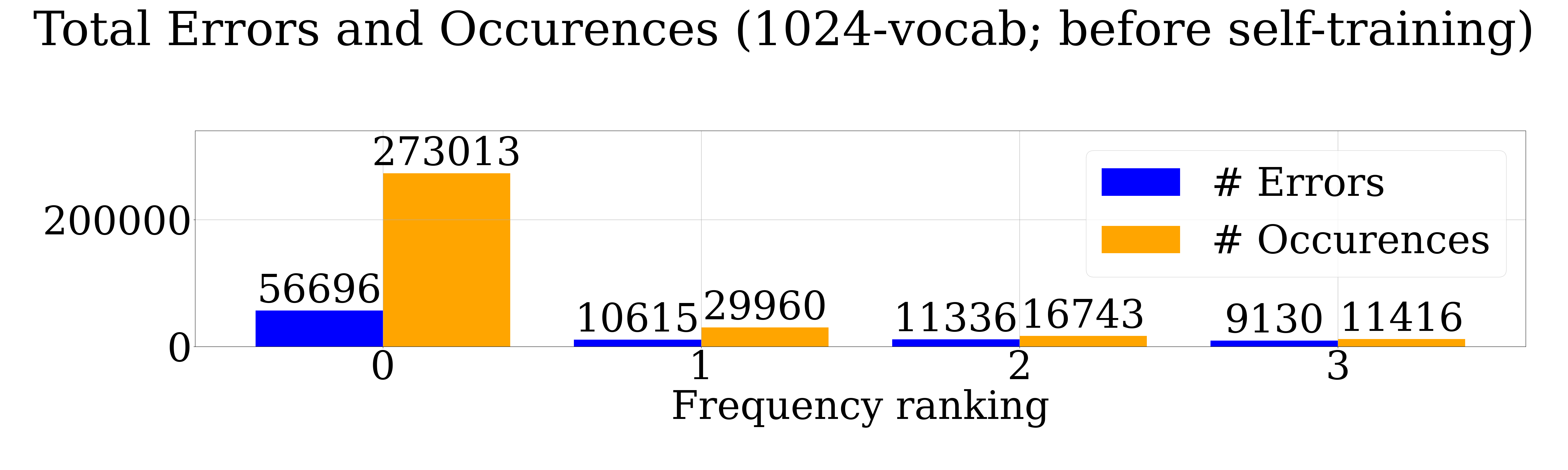}
  \caption{Number of errors and number of total occurrences associated with each bin of words for the 1024-word corpus before self-training.}
  \label{fig:1024_best_unsup_before_st}
\end{figure}

\begin{figure}[H]
  \centering
  \includegraphics[width=\linewidth,trim={0 0 0 0},clip]{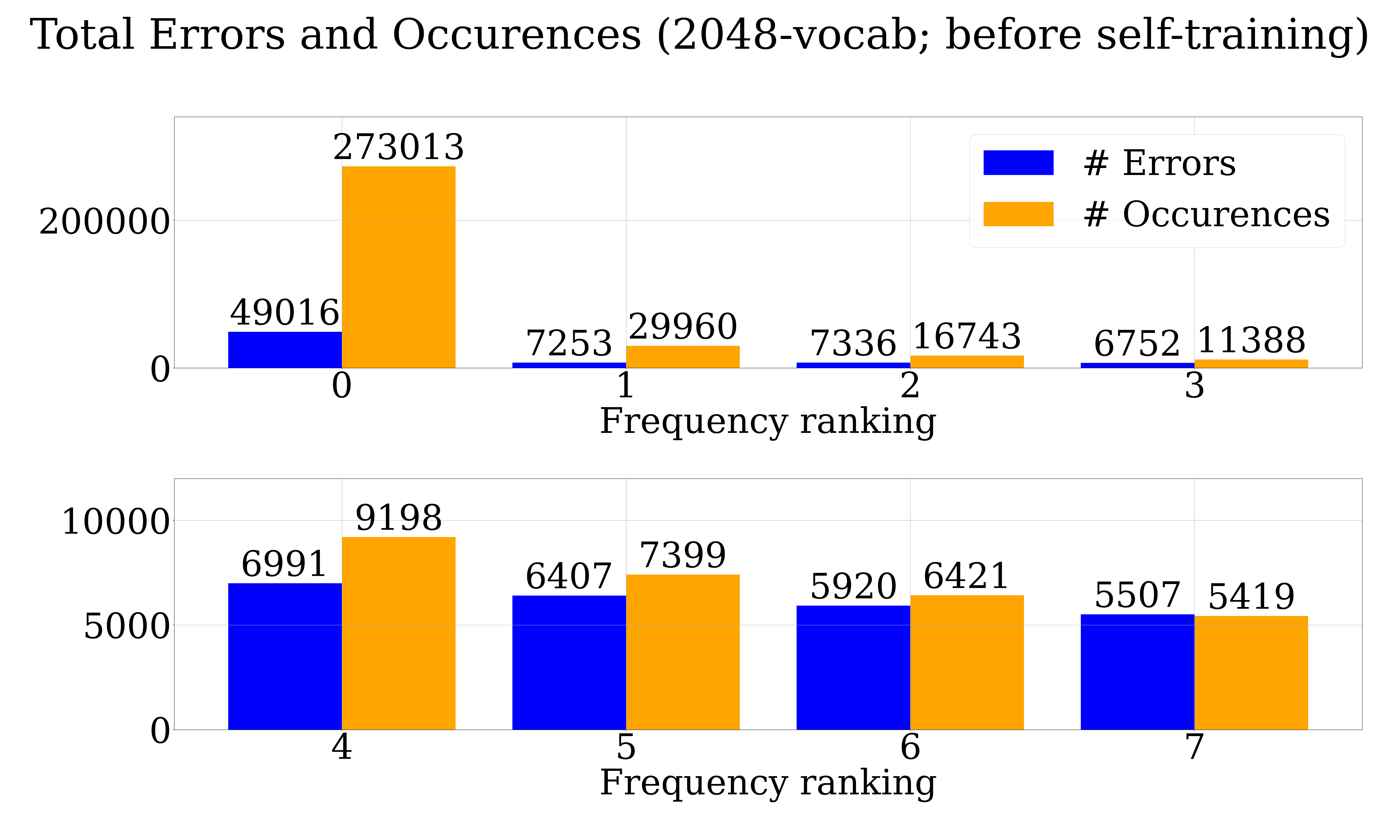}
  \caption{Number of errors and number of total occurrences associated with each bin of words for the 2048-word corpus before self-training.}
  \label{fig:2048_best_unsup_before_st}
\end{figure}

\begin{figure}[H]
  \centering
  \includegraphics[width=\linewidth,trim={0 0 0 0},clip]{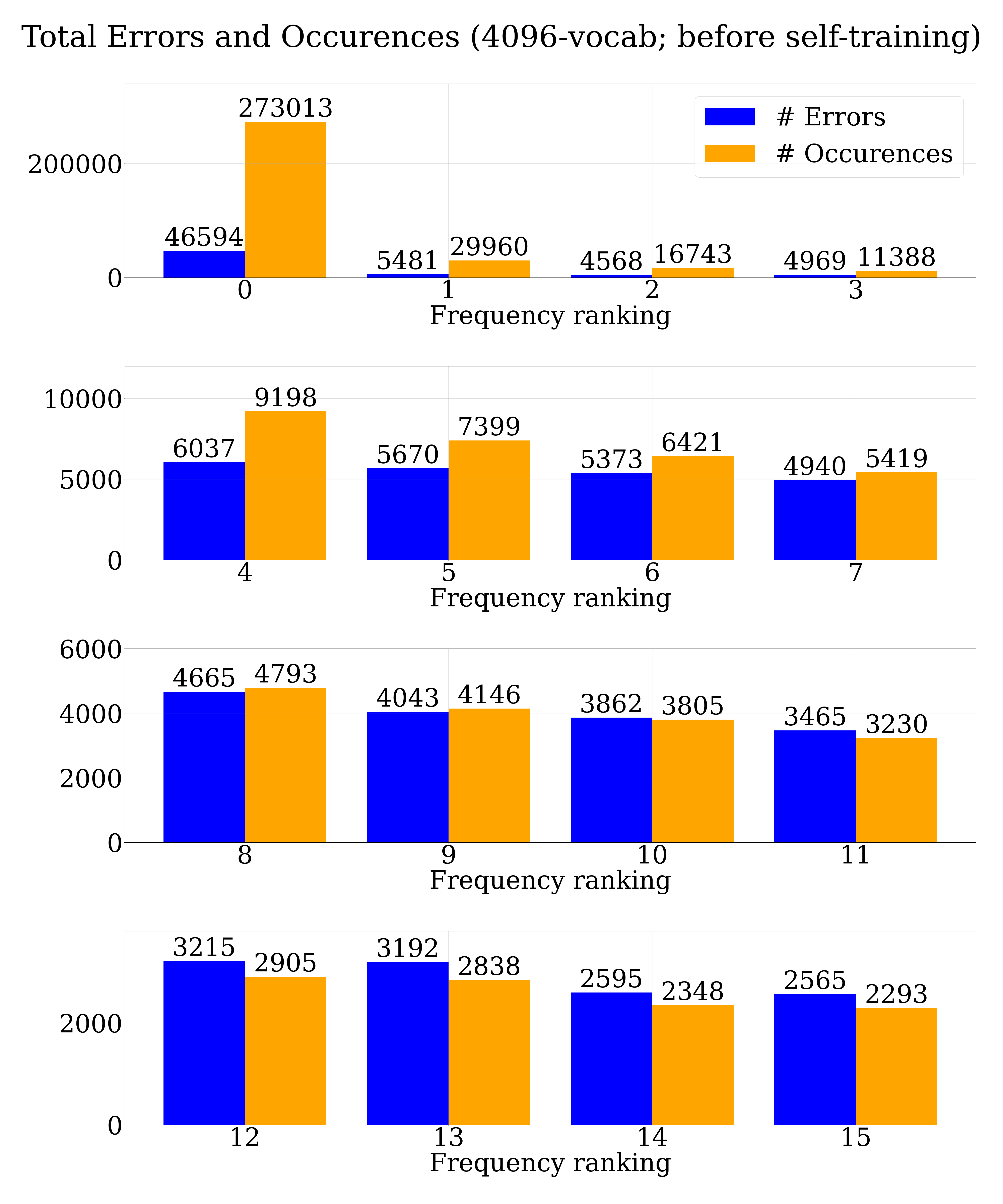}
  \caption{Number of errors and number of total occurrences associated with each bin of words for the 4096-word corpus before self-training.}
  \label{fig:4096_best_unsup_before_st}
\end{figure}

To further understand how well high, mid, and low-frequency words are recognized in each vocabulary setting ($K=1024$, $K=2048$, and $K=4096$), we perform an error analysis by plotting the total number of errors associated with a word against the total number of occurrences of a word in the transcript of the evaluation set. After aligning the ground-truth transcript and the unsupervised model's hypothesis, we attribute substitution errors and deletion errors to the word in the reference transcript, while insertion errors are attributed to the word in the unsupervised hypothesis transcript. We sort the words by occurrence counts and group every 256 words into one bin. Hence, the first bin (with index 0) contains the 256 words that occur most frequently in the transcripts, and the last bin contains the 256 words that occur least frequently in the transcripts. Figures \ref{fig:1024_best_unsup_before_st}, \ref{fig:2048_best_unsup_before_st}, and \ref{fig:4096_best_unsup_before_st} show the error analysis performed for $K=1024$, $K=2048$, and $K=4096$, respectively, for the best JSTTI model trained using unsupervised word boundaries before pseudo-text self-training.

First, we see that the error distribution is very uneven among groups of words with different occurrence frequencies. Word groups with a higher occurrence frequency usually have a lower overall error rate for all three vocabulary sizes. In contrast, low-frequency words barely get recognized correctly, especially for those in the 2048-word and the 4096-word corpora, where the error counts for words in the later word bins even exceed the occurrence counts. Second, comparing the error patterns between the 1024-word corpus and the 2048-word corpus, together with those between the 2048-word corpus and the 4096-word corpus, we see that the error rates for the top 1024 words in the 2048-word corpus, as well as the error rates for the top 2048 words in the 4096-word corpus, significantly drop compared to the same group of words in the 1024-word corpus and the 2048-word corpus, respectively. As we add more words to the curated corpus, we retain more contextual information within the original utterances and text, and our JSTTI criterion is able to take advantage of the richer contextual information to reduce the error rates in both cases.

\begin{figure}[H]
  \centering
  \includegraphics[width=\linewidth,trim={0 0 0 0},clip]{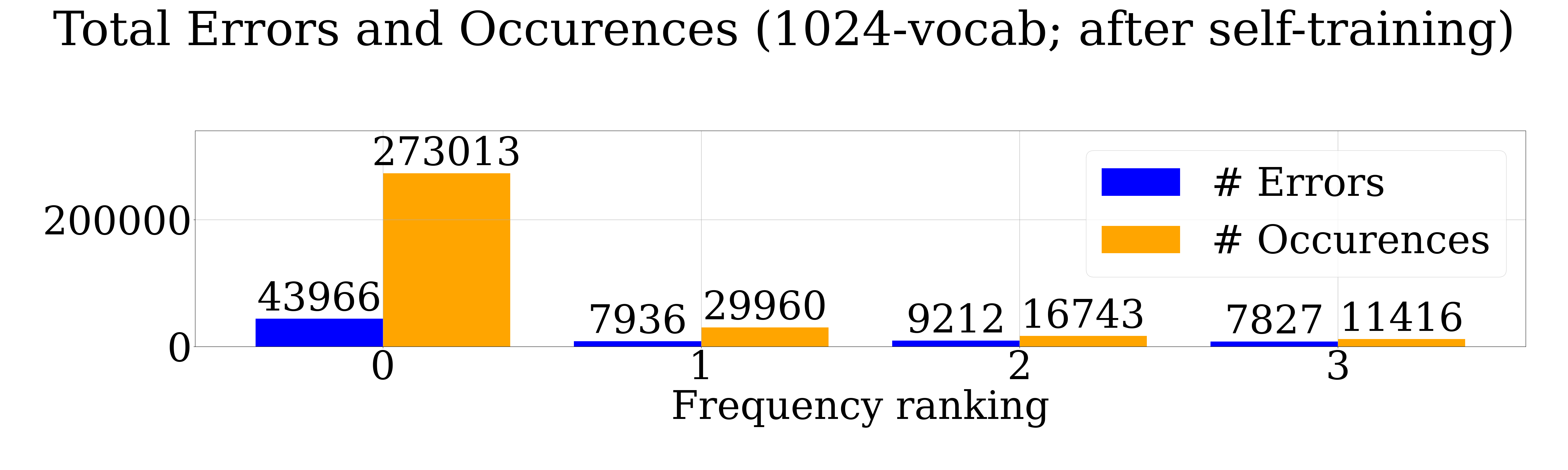}
  \caption{Number of errors and number of total occurrences associated with each bin of words for the 1024-word corpus after self-training.}
  \label{fig:1024_best_unsup_after_st}
\end{figure}

\begin{figure}[H]
  \centering
  \includegraphics[width=\linewidth,trim={0 0 0 0},clip]{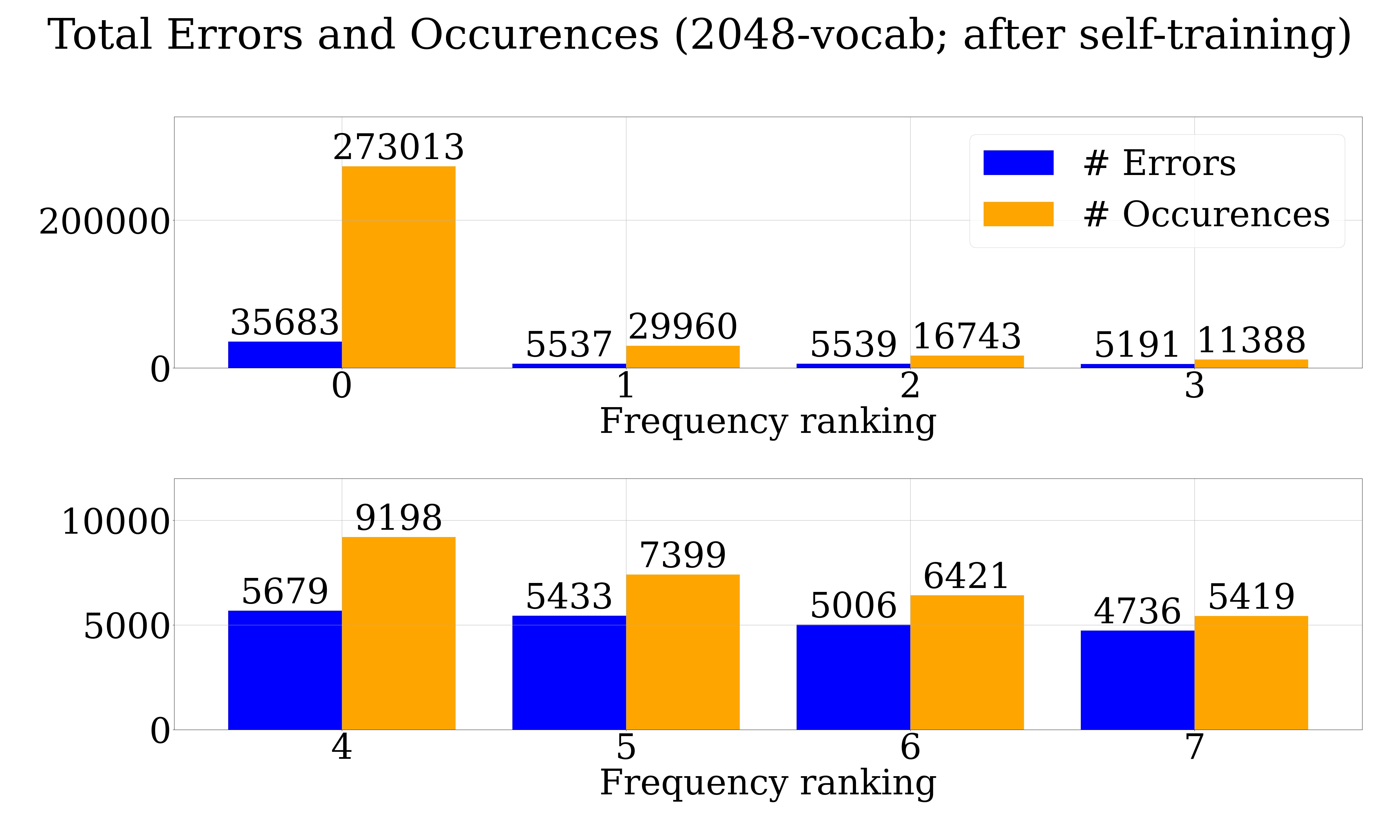}
  \caption{Number of errors and number of total occurrences associated with each bin of words for the 2048-word corpus after self-training.}
  \label{fig:2048_best_unsup_after_st}
\end{figure}

\begin{figure}[H]
  \centering
  \includegraphics[width=\linewidth,trim={0 0 0 0},clip]{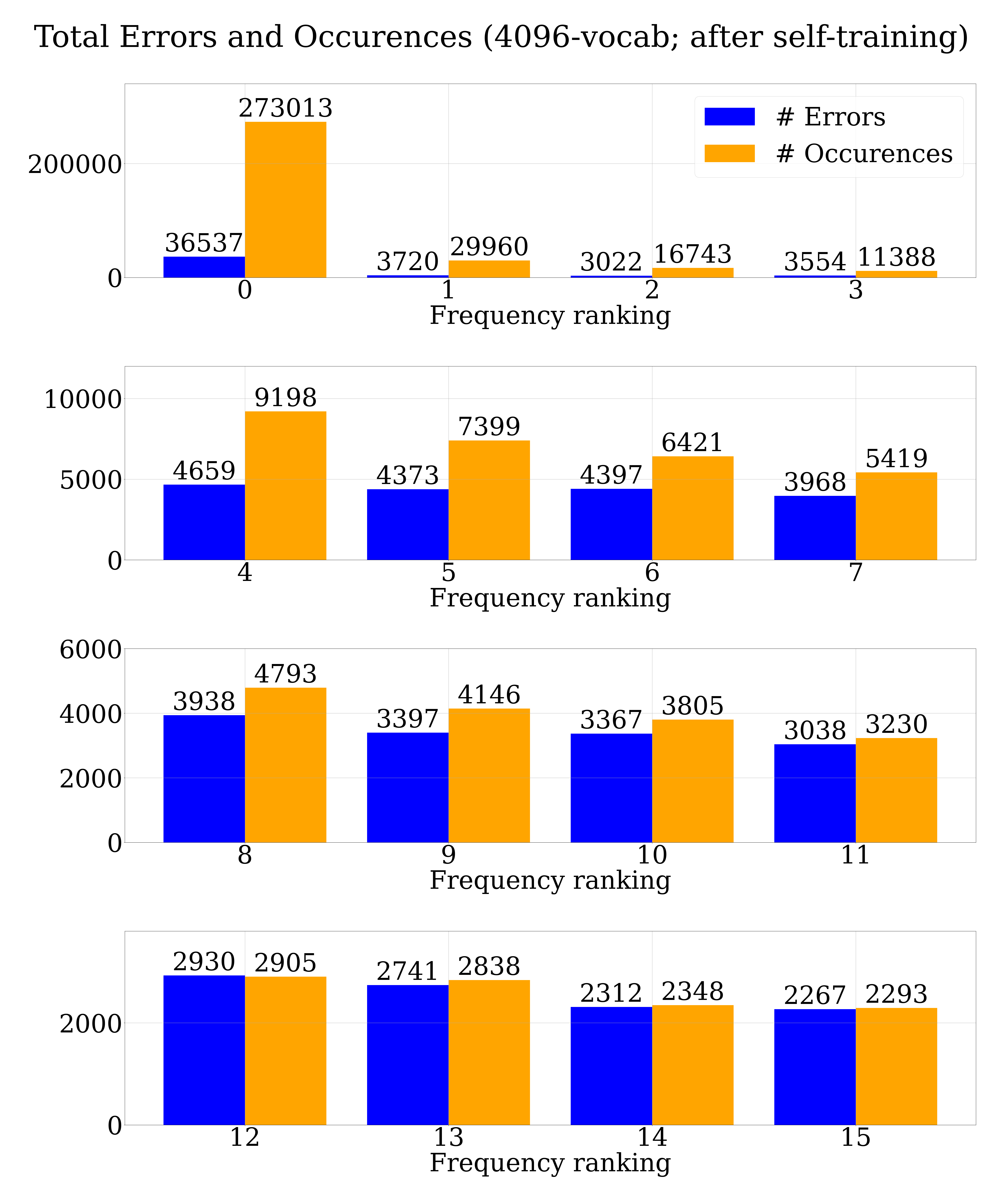}
  \caption{Number of errors and number of total occurrences associated with each bin of words for the 4096-word corpus after self-training.}
  \label{fig:4096_best_unsup_after_st}
\end{figure}

Figures \ref{fig:1024_best_unsup_after_st}, \ref{fig:2048_best_unsup_after_st}, and \ref{fig:4096_best_unsup_after_st} show the error analysis after pseudo-text self-training (c.f. Section \ref{subsec:self_train_trans}) is applied. This simple pseudo-text self-training routine generally reduces word error rates across all frequency bins; however, the effect is more pronounced for words with high occurrence frequencies and milder for words with low occurrence frequencies. For example, the first four word bins (bins 0-3) for the 4096-word corpus see a 24\% error rate reduction, while the last four bins (bins 12-15) only see an 11\% error rate reduction. After self-training, we still incur fewer errors for the top 1024 words in the 2048-word corpus and for the top 2048 words in the 4096-word corpus, compared to the same group of words in the 1024-word corpus and the 2048-word corpus, respectively, as self-training with a pre-trained speech foundation model could further take advantage of the richer contextual information (within the pseudo-transcripts provided by JSTTI models) in the larger-vocabulary settings.

\section{Conclusion}
\label{sec:conclusion}
This work is the first step toward solving the word-level unsupervised ASR task. We curated simplified corpora for this new task and proposed solving the unsupervised ASR problem with the joint speech-text token-infilling (JSTTI) Transformer. Our model’s performance strongly correlates with word segmentation performance: as we iteratively refine the word-level segmental structure, we obtain lower word error rates. Our new model consistently beat two strong baseline models, a position-unigram and skipgram matching (PUSM) model trained on word targets and a REBORN-refined wav2vec-U model trained on character targets. While we develop our method on the 1024-word corpus, it extends seamlessly to the 2048-word corpus, and with careful initialization, we only see minor performance degradation for the 4096-word corpus. In future work, we plan to develop efficient universal unsupervised model selection metrics, test our methods in truly unpaired settings, and improve the recognition performance of rare words.

\appendix
\section{Modified Position-unigram and Skipgram Matching With Arbitrarily Large Batch Size}\label{appendix:modified_pusm}
To obtain a more solid baseline in Table \ref{tab:hubert_word_error_rates} for the word-level unsupervised ASR task, we modify the update strategy for the PUSM loss to perform position-unigram and skipgram loss calculation (c.f. Section \ref{background:pusm}) on an arbitrarily large batch. Our \emph{modified update strategy for PUSM} works by iterating between Algorithms \ref{aggregate_uniksip:first_phase} and \ref{aggregate_uniksip:second_phase}:

\begin{algorithm}[H]
\caption{Unigram and Skipgram Aggregation with Gradient Update (First Phase)}\label{aggregate_uniksip:first_phase}
\begin{algorithmic}[1]
\State \textbf{Parameters:} Forward batch size $B$; number of batches for aggregation $F$; max skipgram skip-size $K$
\State \textbf{Input:} Batches of speech $X^{B_f}$, for $f = 1$ \text{to} $F$; batches of text $Y^{B_f}$, for $f = 1$ \text{to} $F$; generator parameters $G$ (fixed)
\State \textbf{Output:} Aggregated position unigrams $U_{G_x}$ and $U_y$; aggregated skipgrams $S_{G_x}$ and $S_y$
\State Initialize position unigrams $U_{G_x} \leftarrow 0$ and $U_y \leftarrow 0$, both of size $T \times \left|\mathcal{Y}\right|$, for saving $\hat{P}_{X_t} G$ and $\hat{P}_{Y_t}$, for $1 \leq t \leq T$
\State Initialize skipgrams $S_{G_x} \leftarrow 0 $ and $S_y \leftarrow 0$, both of size $K \times \left|\mathcal{Y}\right| \times \left|\mathcal{Y}\right|$, for saving $G^{\top} \hat{P}_k^{X X^{\prime}} G$ and $\hat{P}_k^{Y Y^{\prime}}$, for $1 \leq k \leq K$

\For{$f = 1$ \textbf{to} $F$}
    \State Obtain a batch of speech sequence $X^{B_f}$ and a batch of text sequence $Y^{B_f}$
    \For{each position $t = 1$ \textbf{to} $T$}
        \State Aggregate position unigrams:
        \[
        U_{G_x}\left[t, :\right] \gets U_{G_x}\left[t, :\right] + \operatorname{sg}\left(\hat{P}_{X^{B_f}_t} G\right)
        \] \label{eqn:aggregate_unigram}
        \State
        \[
        U_y\left[t, :\right] \gets U_y\left[t, :\right] + \hat{P}_{Y^{B_f}_t}
        \]
    \EndFor
    \For{each skip-size $k = 1$ \textbf{to} $K$}
        \State Aggregate skipgrams:
        \[
        S_{G_x}\left[k, :, :\right] \gets S_{G_x}\left[k, :, :\right] + \operatorname{sg}\left(G^{\top} \hat{P}_k^{X^{B_f} {X^{B_f}}^{\prime}} G \right)
        \] \label{eqn:aggregate_skipgram}
        \State
        \[
        S_{y}\left[k, :, :\right] \gets S_{y}\left[k, :, :\right] + \hat{P}_k^{Y^{B_f} {Y^{B_f}}^{\prime}}
        \]
    \EndFor
\EndFor
\end{algorithmic}
\end{algorithm}

\begin{algorithm}[H]
\caption{Unigram and Skipgram Aggregation with Gradient Update (Second Phase)}\label{aggregate_uniksip:second_phase}
\begin{algorithmic}[1]
\State \textbf{Parameters:} Forward batch size $B$; number of batches for aggregation $F$; max Skipgram skip-size $K$
\State \textbf{Input:} Batches of speech $X^{B_f}$, for $f = 1$ \text{to} $F$; batches of text $Y^{B_f}$, for $f = 1$ \text{to} $F$; generator parameters $G$ (before gradient update); aggregated position unigrams $U_{G_x}$ and $U_y$; aggregated skipgrams $S_{G_x}$ and $S_y$
\State \textbf{Output:} Generator parameters $G$ (after gradient update)
\State Initialize gradient accumulator: $\Delta G \leftarrow 0$, for collecting gradients across all $F$ batches
\For{$f = 1$ \textbf{to} $F$}
    \State Obtain a batch of speech sequence $X^{B_f}$ and a batch of text sequence $Y^{B_f}$
    \State Calculate unigram loss:
    \[
    \mathcal{L}^{B_f}_{\mathrm{pos}}(G) = \sum_{t=1}^T \left\lVert U_{G_x}\left[t, :\right] + \hat{P}_{X^{B_f}_t} G - \operatorname{sg}\left(\hat{P}_{X^{B_f}_t} G\right) - U_{y}\left[t, :\right] \right\rVert_1
    \]
    \State Calculate skipgram loss:
    \begin{align*}
\mathcal{L}^{B_f}_{\mathrm{skip}}(G) &= \sum_{k=1}^K \left\lVert S_{G_x}\left[k, :, :\right] + G^{\top} \hat{P}_k^{X^{B_f} {X^{B_f}}^{\prime}} G \right. \\
&\quad \left. - \operatorname{sg}\left(G^{\top} \hat{P}_k^{X^{B_f} {X^{B_f}}^{\prime}} G\right) - S_{y}\left[k, :, :\right] \right\rVert_1
\end{align*}
\State Aggregate Gradient: 
\[\Delta G \leftarrow \Delta G + \frac{\partial \mathcal{L}^{B_f}_{\mathrm{pos}}(G)}{\partial G} + \frac{\partial \mathcal{L}^{B_f}_{\mathrm{skip}}(G)}{\partial G}
\]
\EndFor
 \State Perform gradient update to $G$: 
 \[
    G \leftarrow G - \gamma \Delta G
 \]
\end{algorithmic}
\end{algorithm}
As the stop-gradient operator $\operatorname{sg}\left(\cdot\right)$ is used in Lines \ref{eqn:aggregate_unigram} and \ref{eqn:aggregate_skipgram} in Algorithm \ref{aggregate_uniksip:first_phase}, we no longer store the gradient over the position unigram and skipgram losses for each sample. Instead, we only store the summed position unigram and skipgram counts. The gradient update step in Algorithm \ref{aggregate_uniksip:second_phase} is equivalent to using a batch size of $B \times F$ directly. For example, we can see that the skipgram loss over a large batch of $B \times F$ samples decomposes as:

\begin{align}
    \frac{\partial \mathcal{L}^{B \times F}_{\mathrm{skip}}(G)}{\partial G_{xy}}  &=  \frac{\partial \sum_{k=1}^K\left\lVert \sum_{f=1}^{F}G^{\top} \hat{P}_k^{X^{B_f} {X^{B_f}}^{\prime}} G - S_{y}\left[k, :, :\right] \right\rVert_1}{\partial G_{xy}} \nonumber \\
    &=  \sum_{k=1}^K \operatorname{Tr}\left(\frac{\partial\left\lVert \sum_{f=1}^{F}G^{\top} \hat{P}_k^{X^{B_f} {X^{B_f}}^{\prime}} G - S_{y}\left[k, :, :\right] \right\rVert_1}{\partial \sum_{f=1}^{F}G^{\top} \hat{P}_k^{X^{B_f} {X^{B_f}}^{\prime}} G} \right . \nonumber \\ &\quad\quad\quad\mkern9mu \left .\frac{\partial \sum_{f=1}^{F}G^{\top}  \hat{P}_k^{X^{B_f} {X^{B_f}}^{\prime}\top}G}{\partial G_{xy}}\right)\nonumber \\
    &=\sum_{k=1}^K\sum_{f=1}^{F} \operatorname{Tr}\left( \frac{\partial \left\lVert S_{G_{x}\_p\_m}\left[k, :, :\right]- S_{y}\left[k, :, :\right ]\right\rVert_1}{\partial S_{G_{x}\_p\_m}[k, :, :]} \frac{\partial S_{G_{x}\_p\_m}\left[k, :, :\right]^\top}{\partial G_{xy}}\right) \nonumber \\
    &=\sum_{f =1}^{F} \frac{\partial \mathcal{L}^{B_f}_{\mathrm{skip}}}{\partial G_{xy}}
\end{align} 
where $\operatorname{Tr}(X) :=\sum_{i}X_{ii}$ is the trace of matrix $X$ and
\begin{equation}
      S_{G_{x}\_p\_m}\left[k, :, :\right] := S_{G_x}\left[k, :, :\right] + G^{\top} \hat{P}_k^{X^{B_f} {X^{B_f}}^{\prime}} G - 
    \operatorname{sg}\left(G^{\top} \hat{P}_k^{X^{B_f} {X^{B_f}}^{\prime}} G\right).  
\end{equation}
In our implementation, for simplicity, we set the effective batch size $B \times F$ to roughly the size of the entire speech corpus. In all our experiments, we sum the position-dependent unigram loss and the skipgram loss (both with a weight of 1), and we set $K = 4$ in Eq. (\ref{eqn:skipgram}) and Algorithms \ref{aggregate_uniksip:first_phase} and \ref{aggregate_uniksip:second_phase}.

Our two-step update strategy allows for the calculation of the gradient over an arbitrarily large batch size, which creates a stronger baseline for the PUSM model. To show this, we compare three training settings:
\begin{itemize}
    \item A fixed batch size of $6,000$, chosen so that the empirical position unigram and the empirical skipgram calculated for every speech sample in the batch, as well as all the associated gradient graphs, lie just below the total GPU memory of a 32GB V100 GPU (denoted as WER-6k);
    \item A fixed batch size of $6,000$, but only one single update is made after accumulating the gradient across $20$ batches (denoted WER-6k-20); note that this setting is not as effective as using an effective batch size of $12,000$, as accumulating the gradients of the L1 norms (Eqs. (\ref{eqn:pos_unigram}) and (\ref{eqn:skipgram})) across multiple batches is not equivalent to accumulating the position-unigram and skipgram statistics across batches;
    \item The proposed modified training pipeline that allows the use of a true effective batch size of around $120,000$, as we separate position unigram + skipgram estimation and gradient update into two phases (denoted WER-120k).
    \end{itemize}
We show the comparison in Table \ref{tab:modified_uniskip}, where each row corresponds to how we segment the word-level speech tokens for training the PUSM models.

\begin{table}[H]
\caption{WERs for PUSM models trained under the modified update strategy against two baselines. The best result for each word boundary setting is shown in bold. All results are obtained without pseudo-text self-training (i.e., \textbf{Student} = None). }\label{tab:modified_uniskip}
\centering
\begin{tabular}{lccc}
\toprule
Word Segmentation & WER-6k($\downarrow$) & WER-6k-20($\downarrow$) & WER-120k($\downarrow$)\\
\midrule
\midrule
GS+w2b & 52.33 & 51.27 &  \textbf{44.18} \\
\hline
GS+w2b+JE2E & 41.29 & 40.62 &  \textbf{34.08}\\
\hline
GS+w2b+JE2E+w2bL & 39.60 & 38.75 & \textbf{32.11} \\
\midrule
\midrule
Forced Alignment & 24.46 & 23.91 & \textbf{20.89}\\
\bottomrule
\end{tabular}
\end{table}
    
We see that the \textit{WER-120k} strategy leads to a consistent 14-18\% relative reduction in WERs over using a fixed batch size of $6000$, showing that having a large effective batch size is crucial to the PUSM method for word-level unsupervised ASR. On the other hand, \textit{WER-6k-20} offers limited performance gain, showing that simple gradient accumulation is still limited by the significant estimation errors of position-unigram and skipgram statistics on small batches.

\section{Additional Analyses and Ablations}
We perform additional analyses and ablations for the JSTTI-based word-
level unsupervised ASR, pertaining to the choice of input features,
the choice of model architecture for optimizing the JSTTI criterion, regularization sensitivity for the JSTTI E2E-refinement stage (c.f. Section \ref{subsec:unsup_bnd_and_refinement}), sensitivity to the word-level k-means speech quantization model (c.f. Step 3 of Section \ref{subsec:extract_word_level_features}), and an analysis of why obtaining the text output from the intermediate layer of the JSTTI-ENC model is reasonable. We carry out all analyses and ablations on the 1024-word corpus.

\subsection{Ablation: VG-HuBERT features vs HuBERT-Large features for Word-level Unsupervised ASR}
We would like to understand if speech features from a visually grounded word discovery model trained with weakly labeled speech-image pairs offers any benefit for the word-level unsupervised ASR task. Table \ref{tab:vghubert_word_error_rates} reports the WERs when we switch to frame-level features extracted from the 10$^{\text{th}}$ layer of a public VG-HuBERT$_{3}$ checkpoint \citep{peng2022-vghubert}. Comparing the rows in Table \ref{tab:vghubert_word_error_rates} with the relevant rows in Table \ref{tab:hubert_word_error_rates}, we see that word-level features based on VG-HuBERT are less effective than those based on HuBERT-Large, regardless of what word boundaries we use to pool the word-level features.
Despite the performance lag in all settings, applying \textit{JSTTI E2E-refinement} with VG-HuBERT features, together with the subsequent iteration of \textit{wav2bnd-large} training (c.f. Section \ref{subsec:unsup_bnd_and_refinement}), still reduces the WER significantly. We additionally note that \textit{JSTTI E2E-refinement} with VG-HuBERT frame-level features did not improve the word token F1: the F1 score dropped to 35.59\% after \textit{JSTTI E2E-refinement} and only rose to 42.86\% after \textit{wav2bnd-large}. This is expected as the VG-HuBERT model's contextualization tends to push word identity information toward the temporal center of each word \citep{peng2022-vghubert}. 

\begin{table}[H]
\caption{WERs obtained with JSTTI-ENC using discrete speech token sequences extracted with different word segmentation methods, where the discrete speech tokens are pooled and quantized from frame-level VG-HuBERT features. Rows 2-5 correspond to the iterative boundary refinement routine in Section \ref{subsec:unsup_bnd_and_refinement}. The top-line obtained with forced alignments has been underlined, and the best result obtained with unsupervised word boundaries is shown in bold. All results are obtained without pseudo-text self-training (i.e., \textbf{Student} = None).}\label{tab:vghubert_word_error_rates}
\centering
\begin{tabular}{lc}
\toprule
Word Segmentation Method & WER($\downarrow$) \\
\midrule
\midrule
VG-HuBERT & 50.26\\
\midrule
\midrule
GS & 47.47 \\
\hline
GS+w2b  & 41.31 \\
\hline
GS+w2b+JE2E & 35.52 \\
\hline
GS+w2b+JE2E+w2bL & \textbf{31.55}  \\
\midrule
\midrule
Forced Alignment & \underline{22.06} \\
\bottomrule
\end{tabular}
\end{table}

\subsection{Ablation: Encoder-only vs Encoder-decoder Model for Word-level Unsupervised ASR}
To understand if the specific architecture design affects word-level unsupervised ASR performance under the JSTTI task, we compare the JSTTI-ENC model to a one-layer encoder-decoder model with a similar number of parameters. For slightly better convergence, we modify the masking strategy in Section \ref{subsec:model_training} for the encoder-decoder model: first, we replace every selected token span for masking with a single $<\!\text{MASK}\!>$ token, and similarly for every selected token span for random replacement; second, we add insertion noise for the remainder of the masking budget. During training, the encoder-decoder optimizes the combined reconstruction loss using the original speech and text sequences as targets. Using paired validation data to record model checkpoints, we calculate the average teacher-forcing speech-to-text accuracy. During inference, to avoid decoder hallucination, we develop an encoder state matching method to construct a speech-to-text embedding alignment and use the alignment to obtain text predictions during inference. Algorithms \ref{algo:speech_text_dictionary} and \ref{algo:speech_text_dictionary_inference} explain the process of obtaining the alignment and performing inference, which is inspired by comparable methods in unsupervised word translation \citep{Lample2018-word-trans-non-parallel}.

\begin{algorithm}
\caption{Speech-Text Embedding Alignment}\label{algo:speech_text_dictionary}
\begin{algorithmic}[1]
\State \textbf{Input:} Unpaired speech sequences $\mathbf{S}^{TRAINING}$ from the training set, unpaired text sequences $\mathbf{T}^{TRAINING}$ from the training set
\State \textbf{Output:} All pairwise cosine distances $d_{s,t}$ between the embeddings of speech token $s$ and text token $t$
    \State Initialize speech dictionary $D_s \in \mathbb{R}^{M \times k} \gets 0$, and text dictionary $D_t \in \mathbb{R}^{M \times K} \gets 0$, where $M$ is the encoder feature dimension, $k$ is the speech token dictionary size, and $K$ is the text token dictionary size.
    \State For each speech token $s \in [k]$ and each text token $t \in [K]$, obtain their total counts $N_s$ and $N_t$.
    \For {$s_1, \cdots ,s_L = \mathbf{s}^{TRAINING} \in \mathbf{S}^{TRAINING}$}
        \State \[ S\_EMBED_{1}, \cdots, S\_EMBED_{L} \gets \operatorname{ENC}(s_1, \cdots, s_L) \]
        \For{$l \in 1, \cdots, L$}
            \State $D_s[:, s_l] \gets D_s[:, s_l] +  \frac{1}{N_{s_l}} S\_EMBED_{l}$
        \EndFor
    \EndFor
    \For {$t_1, \cdots ,t_L = \mathbf{t}^{TRAINING} \in \mathbf{T}^{TRAINING}$}
        \State \[ T\_EMBED_{1}, \cdots, T\_EMBED_{L} \gets \operatorname{ENC}(t_1, \cdots, t_L) \]
        \For{$l \in 1, \cdots, L$}
            \State $D_t[:, t_l] \gets D_t[:, t_l] +  \frac{1}{N_{t_l}} T\_EMBED_{l}$
        \EndFor
    \EndFor
    \For{$s \in [k]$}
        \For{$t \in [K]$}
            \State $d_{s,t} \gets \operatorname{cos-sim}(D_s[:, s], D_t[:, t])$
        \EndFor
    \EndFor
\end{algorithmic}
\end{algorithm}

\begin{algorithm}
\caption{Unsupervised ASR Inference Using Speech-Text Embedding Alignment}\label{algo:speech_text_dictionary_inference}
\begin{algorithmic}[1]
\State \textbf{Input:} A speech utterance sequence $s_1, \cdots, s_L = \mathbf{s}$ from the evaluation set, where each $s_l$ is a discrete speech token; all pairwise cosine distances $d_{s,t}$ between the embeddings of
speech token s and text token t
\State \textbf{Output:} The predicted transcript for $\mathbf{s}$: $\hat{t}_1, \cdots, \hat{t}_L = \hat{\mathbf{t}}$
    \State $\hat{\mathbf{t}} \gets \varnothing$
    \For{$s_l \in s_1, \cdots, s_L$}
        \State \[ \hat{t} \gets \operatorname{argmin}_{t} d_{s_l, t} \]
        \State \[\hat{\mathbf{t}} \gets [\hat{\mathbf{t}}, \hat{t}]\]
    \EndFor
\end{algorithmic}
\end{algorithm}


We report the ablations in Table \ref{tab:enc_vs_enc_dec_word_error_rates} and compare them with the established main results in Table \ref{tab:hubert_word_error_rates} for each word segmentation method used to obtain discrete speech tokens. Note that the basic JSTTI encoder-decoder model can be similarly augmented with \textit{JSTTI E2E-refinement} and \textit{wav2bnd-large} refinement (c.f. Section \ref{subsec:unsup_bnd_and_refinement}), and the corresponding results are additionally reported in Table \ref{tab:enc_vs_enc_dec_word_error_rates}. Running \textit{JSTTI E2E-refinement} and \textit{wav2bnd-large} refinement with the encoder-decoder model gives us a total of two additional sets of word boundaries, and we use these two additional sets of word boundaries to extract word-level speech tokens and train two additional encoder-only JSTTI models in Table \ref{tab:enc_vs_enc_dec_e2e_bnd_word_error_rates}.

\begin{table}[H]
\caption{WERs obtained with the JSTTI-ENC-DEC model under different word segmentation methods. Rows 2-5 correspond to the iterative boundary refinement routine in Section \ref{subsec:unsup_bnd_and_refinement}. The top-line obtained with forced alignments has been underlined, and the best result obtained with unsupervised word boundaries is shown in bold. All results are obtained without pseudo-text self-training (i.e., \textbf{Student} = None).}\label{tab:enc_vs_enc_dec_word_error_rates}
\centering
\begin{tabular}{lc}
\toprule
Word Segmentation Method & WER($\downarrow$) \\
\midrule
\midrule
VG-HuBERT  & 56.14 \\
\hline
GS & 53.42\\
\hline
GS+w2b  & 46.03\\
\hline
GS+w2b+JE2E & 39.23\\
\hline
GS+w2b+JE2E+w2bL & \textbf{33.97} \\
\midrule
\midrule
Forced Alignment & \underline{26.98}  \\
\bottomrule
\end{tabular}
\end{table}







\begin{table}[H]
\caption{WERs obtained with the JSTTI-ENC using JSTTI-E2E refined boundaries from the segmenter of the JSTTI-ENC-DEC model, together with boundaries obtained after subsequent \textit{wav2bnd-large}. All results are obtained without pseudo-text self-training (i.e., \textbf{Student} = None).}\label{tab:enc_vs_enc_dec_e2e_bnd_word_error_rates}
\centering
\begin{tabular}{m{80mm}l}
\toprule
Word Segmentation Method & WER($\downarrow$) \\
\midrule
\midrule
GS+w2b+JE2E \newline (JE2E Segmenter From JSTTI-ENC-DEC) & 28.96 \tablefootnote{As the result in row 1 of this table is obtained by \emph{fixing} the CNN segmenter imported from the JSTTI-ENC-DEC model, for a fair comparison, we also further fine-tune the JSTTI-ENC checkpoint obtained for row 15 of Table \ref{tab:hubert_word_error_rates} by \emph{fixing} the CNN segmenter, and obtained a WER of 28.26\%}\\
\hline
GS+w2b+JE2E+w2bL \newline (JE2E Segmenter From JSTTI-ENC-DEC)  & 26.47 \\
\bottomrule
\end{tabular}
\end{table}

Table \ref{tab:enc_vs_enc_dec_word_error_rates} shows that the JSTTI-ENC-DEC model combined with the encoder state matching method generally gives worse error rates. We believe this is because the encoder-only model directly applies the text output layer to the first encoder layer and could better utilize contextualized representation during evaluation, while the encoder state matching method used for evaluating the encoder-decoder model cannot.

However, by comparing the WERs in row 1 of Table \ref{tab:enc_vs_enc_dec_e2e_bnd_word_error_rates} to row 15 in Table \ref{tab:hubert_word_error_rates}, we see that both the encoder-decoder architecture and the encoder-only architecture are almost equally capable of refining the initial \textit{GradSeg + wav2bnd} word boundaries via the JSTTI E2E-refinement routine. This continues to hold as we further apply \textit{wav2bnd-large} refinement, when we compare row 2 in Table \ref{tab:enc_vs_enc_dec_e2e_bnd_word_error_rates} to row 16 in Table \ref{tab:hubert_word_error_rates}.










\subsection{Ablation: Regularization Strength for JSTTI E2E-Refinement}\label{subsec:e2e_hyperparamter}
We add the word count loss and the word frequency loss (c.f. Eqs. (\ref{eqn:wc_loss}), (\ref{eqn:wf_loss}) and (\ref{eqn:loss_jstti_e2e_ref})) for our JSTTI E2E-refinement routine (c.f. Section \ref{subsec:unsup_bnd_and_refinement}) to improve the word-level segmental structure. We repeat the refinement routine with three distinct sets of regularization strengths to show that adding this regularization and setting large enough weights in the loss are crucial. We denote them as \textit{small regularization} with $\gamma_1 = 5, \gamma_2 = 5,000$, \textit{medium regularization} with $\gamma_1 = 500, \gamma_2 = 500,000$ and \textit{strong regularization} with $\gamma_1 = 50,000, \gamma_2 = 50,000,000$. The \textit{medium regularization} setting here corresponds to what is reported for the setting \textit{GradSeg + wav2bnd + JSTTI E2E-refinement} in Tables \ref{tab:hubert_word_error_rates} and \ref{tab:hubert_word_boundary_metrics}. The three runs are reported in Table \ref{tab:regularization_e2e_error_rates}.

\begin{table}[H]
\caption{WERs for the JSTTI-ENC model under the JSTTI E2E-refinement step, with three different regularization strengths applied. The JSTTI-ENC model and CNN segmenter within are separately pre-trained and initialized according toSection \ref{subsec:unsup_bnd_and_refinement} before JSTTI E2E-refinement. The best results are shown in bold. All results are obtained without pseudo-text self-training (i.e., \textbf{Student} = None).}\label{tab:regularization_e2e_error_rates}
\centering
\begin{tabular}{m{50mm}cc}
\toprule
Regularization Strength & WER($\downarrow$) & Token F-1($\uparrow$)\\
\midrule
\midrule
Small Regularization & 33.25 &  58.27\\ 
\hline
Medium Regularization &  29.39 & \textbf{63.36}\\
\hline
Strong Regularization  & \textbf{29.18} & 63.31 \\ 
\bottomrule
\end{tabular}
\end{table}


From Table \ref{tab:regularization_e2e_error_rates}, we see that setting the regularization coefficients $\gamma_1$ and $\gamma_2$ too small harms performance but setting them too large is not greatly detrimental.

\subsection{Ablation: Choice of Word-level Clusters for Speech Quantization}\label{subsec:choice_word_clusters}
Throughout our experiments, we used different sets of word boundaries for temporal pooling of the word-level feature vectors but kept the cluster centroids fixed across the speech token extraction process (c.f. Section \ref{subsec:extract_word_level_features}; step 3 of the feature extraction process). These cluster centroids were obtained by training a k-means model on top of word-level speech feature vectors mean-pooled using the word boundaries predicted by VG-HuBERT. We denote cluster centroids obtained this way as VG-HuBERT-induced cluster centroids. Doing so saved training time as we could fine-tune a previous JSTTI-ENC model checkpoint once better-quality speech token sequences become available. To show that this choice is indeed near-optimal, we re-cluster the speech features mean-pooled using word-level forced alignments to obtain a new k-means quantizer, re-train the JSTTI-ENC model using speech token sequences quantized by this k-means quantizer, and report the results in Table \ref{tab:vghubert_bnd_clus_error_rates_ablation}. 

\begin{table}[H]
\caption{WERs obtained with two different JSTTI-ENC models when forced alignments are used to pool word-level speech features. \textit{FA BND \& FA CLUS} denotes the setting where forced aligned boundaries are used for both pooling word-level features and then obtaining the word-level k-means quantizer, while \textit{FA BND \& VG-HuBERT CLUS} denotes the setting where the k-means quantizer is trained under word-level features pooled using unsupervised word boundaries provided by VG-HuBERT (as in Section \ref{subsec:extract_word_level_features}). All results are obtained without pseudo-text self-training (i.e., \textbf{Student} = None).}\label{tab:vghubert_bnd_clus_error_rates_ablation}
\centering
\begin{tabular}{lc}
\toprule
Settings & WER($\downarrow$)\\
\midrule
\midrule
FA BND \& FA CLUS & 17.86 \\
\hline
FA BND \& VG-HUBERT CLUS & 18.06\\
\bottomrule
\end{tabular}
\end{table}

Table \ref{tab:vghubert_bnd_clus_error_rates_ablation}
shows that we only get a slight improvement if we were to switch away from using VG-HuBERT-induced cluster centroids, even under the top-line setting with forced alignment boundaries available. This shows that fixing the k-means quantization model to the one trained with word-level features pooled from VG-HuBERT unsupervised word boundaries (Step 3 of Section \ref{subsec:extract_word_level_features}) is reasonable.

\subsection{Analysis: JSTTI-ENC Model Layer-wise Performance}\label{subsec:model_layerwise}
We try to understand why the model's first encoder layer offers better contextualized speech representations during the speech-to-text evaluation stage of the JSTTI-ENC model, even though the representation from the second layer is used during the training stage (under joint speech-to-speech token infilling and text-to-text token infilling). We show the evaluation set WERs and training set WERs when speech representations from either the first or the second encoder layer are fed into the text output layer for the word-level unsupervised ASR task, under different word boundary settings. We report the results under five different boundary settings in Tables \ref{tab:l1_vs_l2_eval} and \ref{tab:l1_vs_l2_train}.

\begin{table}[H]
\caption{WERs for the JSTTI-ENC models when we feed either layer one or layer two speech features from the evaluation set to the text output layer (WER-L1-Eval and WER-L2-Eval, respectively). The better results within each row are shown in bold. All results are obtained without pseudo-text self-training (i.e., \textbf{Student} = None).}\label{tab:l1_vs_l2_eval}
\centering
\begin{tabular}{lcc}
\toprule
Word Segmentation Method & WER-L1-EVAL($\downarrow$) & WER-L2-EVAL($\downarrow$) \\
\midrule
\midrule
GS & \textbf{44.76} & 54.59\\
\hline
GS+w2b & \textbf{38.08} & 48.02\\
\hline
GS+w2b+JE2E\tablefootnote{We further fine-tune the JSTTI-ENC checkpoint obtained for row 15 of Table \ref{tab:hubert_word_error_rates} by \emph{fixing} the CNN segmenter, and hence the WER is lower than previously reported.}& \textbf{28.26} &  39.02\\
\hline
GS+w2b+JE2E+w2bL  & \textbf{26.51} &  37.57\\
\midrule
\midrule
Forced Alignment & \textbf{18.06} & 27.76 \\
\bottomrule
\end{tabular}
\end{table}

\begin{table}[H]
\caption{WERs for the JSTTI-ENC models when we feed either layer one or layer two speech features from the training set to the text output layer (WER-L1-Train and WER-L2-Train, respectively). The better results within each row are shown in bold. All results are obtained without pseudo-text self-training (i.e., \textbf{Student} = None).}\label{tab:l1_vs_l2_train}
\centering
\begin{tabular}{lcc}
\toprule
Word Segmentation Method & WER-L1-TRAIN($\downarrow$) & WER-L2-TRAIN($\downarrow$) \\
\midrule
\midrule
GS & \textbf{43.55} & 48.41\\
\hline
GS+w2b & \textbf{36.69} & 41.54\\
\hline
GS+w2b+JE2E & \textbf{26.61} & 30.89 \\
\hline
GS+w2b+JE2E+w2bL  & \textbf{24.92} &  29.11\\
\midrule
\midrule
Forced Alignment & \textbf{16.58} & 19.89 \\
\bottomrule
\end{tabular}
\end{table}

From Tables \ref{tab:l1_vs_l2_eval} and \ref{tab:l1_vs_l2_train}, we see that feeding the contextualized speech representations from the first encoder layer into the text output layer consistently shows significantly better performance on both the training set and the evaluation set. Comparing the respective entries among the two tables, we see that using the second encoder layer's representations exhibits significantly worse over-fitting on the training set, while the amount of over-fitting is more subdued if the representations from the first layer are used. We hypothesize that after JSTTI training, the first encoder layer learns a more universal representation that is shared between the speech and text modalities, while the representations from the second layer tend to be modality-specific as they are directly connected with the modality-specific output layers when optimizing the JSTTI criterion.

To further showcase that the first encoder layer indeed learns a shared speech-text representation, we report the cross-modal word purity (WP), cross-modal cluster purity (CP), and the cross-modal word-normalized mutual information (WNMI), similarly defined as in \citep{Hsu2022-hubert} to measure the quality of the discrete representation of layer-wise features. To calculate these metrics, we first extract contextualized word-level speech features and text features on the entire training set of discrete word-level speech token sequences and word-level text from either layer one or layer two of the JSTTI-ENC model. We then perform k-means clustering on the extracted contextualized \emph{speech} representations (with $k=1024$) and obtain a set of cluster centroids. Finally, we quantize the extracted \emph{text} representations with this set of cluster centroids. Denote $Y_t$ as a word label in text and $L_t$ as the corresponding k-means label obtained from the above routine. We define the empirical joint probability $\hat{P}_{Y L}$, as
\begin{align}
    \hat{P}_{Y L}(i, j)=\frac{\sum_{t=1}^T\left[Y_t=i \wedge L_t=j\right]}{T}
\end{align}
where $T$ is the total number of extracted word-level representations in the training set. From the joint probability, we can compute the mostly likely word label $Y^*(j)$ for each k-means label $j$, and the most likely k-means label $L^*(i)$ for each word label $i$ as:
\begin{align}
L^*(i) & =\arg \max _j \hat{P}_{Y L}(i, j)\\
Y^*(j) & =\arg \max _i \hat{P}_{Y L}(i, j) .
\end{align}
We can define word purity (WP), cluster purity (CP), and word-normalized mutual information (WNMI) as:
\begin{align}
\mathrm{WP} &= \mathbb{E}_{\hat{P}_{L}(j)}\left[\hat{P}_{Y \mid L}\left(Y^*(j) \mid j\right)\right]\\
\mathrm{CP} &= \mathbb{E}_{\hat{P}_{Y}(i)}\left[\hat{P}_{L \mid Y}\left(L^*(i) \mid i\right)\right]\\
\mathrm{WNMI} &= \frac{I(Y ; L)}{H(Y)} =-\frac{\sum_i \sum_j \hat{P}_{Y L}(i, j) \log \frac{\hat{P}_{Y L}(i, j)}{\hat{P}_Y(i) \hat{P}_L(j)}}{\sum_i \hat{P}_Y(i) \log \hat{P}_Y(i)}
\end{align}
where we denote the marginal distribution over k-means label $j$ as $\hat{P}_L(j)$, the marginal distribution over word label $i$ as $\hat{P}_Y(i)$, the conditional probability of a word label $i$ given a k-means label $j$ as $\hat{P}_{Y \mid L}(i \mid j)$, the conditional probability of a k-means label $j$ given a word label $i$ as $\hat{P}_{L \mid Y}(j \mid i)$, the mutual information between $Y$ and $L$ as $I(Y ; L)$ and the entropy of $Y$ as $H(Y)$. WP can be interpreted as the word accuracy if we transcribe each k-means class with its most likely word label. CP is the counterpart of WP and measures the average probability that the same word is labeled as the same k-means class. WNMI is an information-theoretic metric that measures the percentage of uncertainty eliminated for a word label $Y$ after observing its k-means label $L$ \citep{Hsu2022-hubert}. We only report these metrics in Table \ref{tab:l1_vs_l2_train_cp_wp_wnmi} for the setting trained with word-level speech token sequences obtained under the boundary setting denoted as \textit{GradSeg + wav2bnd + JSTTI E2E-refinement + wav2bnd-large} (c.f. Section \ref{subsec:unsup_bnd_and_refinement}), as results obtained under other settings are similar. 
\begin{table}[H]
\caption{WP, CP, and WNMI calculated with cross-modal quantization on top of layer one (L1) and layer one (L2) features extracted with JSTTI-ENC model on the training set. The better results among the two layers are shown in bold.}\label{tab:l1_vs_l2_train_cp_wp_wnmi}
\centering
\begin{tabular}{lccc}
\toprule
Layer & WP($\uparrow$) & CP($\uparrow$) 
 & CWMI($\uparrow$) \\
 \midrule
 \midrule
L1 & \textbf{0.831} & \textbf{0.550} & \textbf{0.855}\\
\hline
L2 & 0.581 & 0.378 & 0.577\\
\bottomrule
\end{tabular}
\end{table}
Table \ref{tab:l1_vs_l2_train_cp_wp_wnmi} shows better cross-modal WP, CP, and WNMI for layer one rather than layer two, again solidifying the hypothesis that layer one learns a shared representation between speech and text modalities, while the representations from layer 2 are less shared across the two modalities.

\section*{CRediT authorship contribution statement}
\textbf{Junrui Ni:} Conceptualization,  Methodology, Software, Validation, Formal analysis, Investigation, Data Curation, Writing - Original Draft, Writing - Review \& Editing, Visualization. \textbf{Liming Wang:} Conceptualization, Methodology, Software, Writing - Review \& Editing, Visualization. \textbf{Yang Zhang:} Conceptualization, Resources, Supervision. \textbf{Kaizhi Qian:} Conceptualization, Resources, Supervision.  \textbf{Heting Gao:}  Writing - Review \& Editing, Visualization. \textbf{Mark Hasegawa-Johnson:} Conceptualization, Methodology, Resources, Writing - Review \& Editing, Supervision, Project administration, Funding acquisition. \textbf{Chang D. Yoo:} Methodology, Writing - Review \& Editing, Supervision, Project administration, Funding acquisition. 

\section*{Declaration of competing interest}
The authors declared that they had no competing interests that could appear to influence the work reported in this article.

\section*{Data availability}
The experiment is based on LibriSpeech~\citep{Panayotov15-LibriSpeech}, and the forced alignment used to curate the corpora in this study can be found in \url{https://github.com/CorentinJ/librispeech-alignments}. All code relevant to this study has been uploaded to \url{https://github.com/JeromeNi/wholeword-uasr-jstti}, with full instructions on replicating the results.

\section*{Acknowledgments}
Funding: This research was supported in part by a grant (No. 116499) from the Institute for Information \& Communication Technology Promotion (IITP).  All findings and opinions are those of the authors and are not endorsed by IITP.

During the preparation of this work the authors used ChatGPT in order to convert algorithm descriptions into LaTeX formats so that the descriptions are more readable. After using this tool/service, the authors reviewed and edited the content as needed and take full responsibility for the content of the publication.

\bibliographystyle{elsarticle-harv} 
\bibliography{cas-refs}

\begin{thebibliography}{43}
\expandafter\ifx\csname natexlab\endcsname\relax\def\natexlab#1{#1}\fi
\providecommand{\url}[1]{\texttt{#1}}
\providecommand{\href}[2]{#2}
\providecommand{\path}[1]{#1}
\providecommand{\DOIprefix}{doi:}
\providecommand{\ArXivprefix}{arXiv:}
\providecommand{\URLprefix}{URL: }
\providecommand{\Pubmedprefix}{pmid:}
\providecommand{\doi}[1]{\href{http://dx.doi.org/#1}{\path{#1}}}
\providecommand{\Pubmed}[1]{\href{pmid:#1}{\path{#1}}}
\providecommand{\bibinfo}[2]{#2}
\ifx\xfnm\relax \def\xfnm[#1]{\unskip,\space#1}\fi
\bibitem[{Algayres et~al.(2023)Algayres, Diego{-}Simon, Sagot and Dupoux}]{Algayres23-xlsr-wav2bnd}
\bibinfo{author}{Algayres, R.}, \bibinfo{author}{Diego{-}Simon, P.}, \bibinfo{author}{Sagot, B.}, \bibinfo{author}{Dupoux, E.}, \bibinfo{year}{2023}.
\newblock \bibinfo{title}{{XLS-R} fine-tuning on noisy word boundaries for unsupervised speech segmentation into words}, in: \bibinfo{editor}{Bouamor, H.}, \bibinfo{editor}{Pino, J.}, \bibinfo{editor}{Bali, K.} (Eds.), \bibinfo{booktitle}{Findings of the Association for Computational Linguistics: {EMNLP} 2023, Singapore, December 6-10, 2023}, \bibinfo{publisher}{Association for Computational Linguistics}. pp. \bibinfo{pages}{12103--12112}.
\newblock \URLprefix \url{https://doi.org/10.18653/v1/2023.findings-emnlp.810}, \DOIprefix\doi{10.18653/V1/2023.FINDINGS-EMNLP.810}.
\bibitem[{Algayres et~al.(2022)Algayres, Ricoul, Karadayi, Lauren{\c{c}}on, Zaiem, Mohamed, Sagot and Dupoux}]{algayres-etal-2022-dp}
\bibinfo{author}{Algayres, R.}, \bibinfo{author}{Ricoul, T.}, \bibinfo{author}{Karadayi, J.}, \bibinfo{author}{Lauren{\c{c}}on, H.}, \bibinfo{author}{Zaiem, S.}, \bibinfo{author}{Mohamed, A.}, \bibinfo{author}{Sagot, B.}, \bibinfo{author}{Dupoux, E.}, \bibinfo{year}{2022}.
\newblock \bibinfo{title}{{DP}-parse: Finding word boundaries from raw speech with an instance lexicon}.
\newblock \bibinfo{journal}{Transactions of the Association for Computational Linguistics} \bibinfo{volume}{10}, \bibinfo{pages}{1051--1065}.
\newblock \URLprefix \url{https://aclanthology.org/2022.tacl-1.61}, \DOIprefix\doi{10.1162/tacl_a_00505}.
\bibitem[{Ao et~al.(2022)Ao, Wang, Zhou, Wang, Ren, Wu, Liu, Ko, Li, Zhang, Wei, Qian, Li and Wei}]{Ao2022-speecht5}
\bibinfo{author}{Ao, J.}, \bibinfo{author}{Wang, R.}, \bibinfo{author}{Zhou, L.}, \bibinfo{author}{Wang, C.}, \bibinfo{author}{Ren, S.}, \bibinfo{author}{Wu, Y.}, \bibinfo{author}{Liu, S.}, \bibinfo{author}{Ko, T.}, \bibinfo{author}{Li, Q.}, \bibinfo{author}{Zhang, Y.}, \bibinfo{author}{Wei, Z.}, \bibinfo{author}{Qian, Y.}, \bibinfo{author}{Li, J.}, \bibinfo{author}{Wei, F.}, \bibinfo{year}{2022}.
\newblock \bibinfo{title}{{SpeechT5}: Unified-modal encoder-decoder pre-training for spoken language processing}, in: \bibinfo{editor}{Muresan, S.}, \bibinfo{editor}{Nakov, P.}, \bibinfo{editor}{Villavicencio, A.} (Eds.), \bibinfo{booktitle}{Proceedings of the 60th Annual Meeting of the Association for Computational Linguistics (Volume 1: Long Papers), {ACL} 2022, Dublin, Ireland, May 22-27, 2022}, \bibinfo{publisher}{Association for Computational Linguistics}. pp. \bibinfo{pages}{5723--5738}.
\newblock \URLprefix \url{https://doi.org/10.18653/v1/2022.acl-long.393}, \DOIprefix\doi{10.18653/V1/2022.ACL-LONG.393}.
\bibitem[{Artetxe et~al.(2018)Artetxe, Labaka, Agirre and Cho}]{Artetxe18-unsupnmt}
\bibinfo{author}{Artetxe, M.}, \bibinfo{author}{Labaka, G.}, \bibinfo{author}{Agirre, E.}, \bibinfo{author}{Cho, K.}, \bibinfo{year}{2018}.
\newblock \bibinfo{title}{Unsupervised neural machine translation}, in: \bibinfo{booktitle}{6th International Conference on Learning Representations, {ICLR} 2018, Vancouver, BC, Canada, April 30 - May 3, 2018, Conference Track Proceedings}, \bibinfo{publisher}{OpenReview.net}.
\newblock \URLprefix \url{https://openreview.net/forum?id=Sy2ogebAW}.
\bibitem[{Ay et~al.(2023)Ay, Özbakır, Kulluk, Gülmez, Öztürk and Özer}]{Ay2023-fckmeans}
\bibinfo{author}{Ay, M.}, \bibinfo{author}{Özbakır, L.}, \bibinfo{author}{Kulluk, S.}, \bibinfo{author}{Gülmez, B.}, \bibinfo{author}{Öztürk, G.}, \bibinfo{author}{Özer, S.}, \bibinfo{year}{2023}.
\newblock \bibinfo{title}{{FC-Kmeans}: Fixed-centered {K-means} algorithm}.
\newblock \bibinfo{journal}{Expert Systems with Applications} \bibinfo{volume}{211}, \bibinfo{pages}{118656}.
\newblock \URLprefix \url{https://www.sciencedirect.com/science/article/pii/S0957417422016979}, \DOIprefix\doi{https://doi.org/10.1016/j.eswa.2022.118656}.
\bibitem[{Babu et~al.(2022)Babu, Wang, Tjandra, Lakhotia, Xu, Goyal, Singh, von Platen, Saraf, Pino, Baevski, Conneau and Auli}]{Babu2022-xlsrv2}
\bibinfo{author}{Babu, A.}, \bibinfo{author}{Wang, C.}, \bibinfo{author}{Tjandra, A.}, \bibinfo{author}{Lakhotia, K.}, \bibinfo{author}{Xu, Q.}, \bibinfo{author}{Goyal, N.}, \bibinfo{author}{Singh, K.}, \bibinfo{author}{von Platen, P.}, \bibinfo{author}{Saraf, Y.}, \bibinfo{author}{Pino, J.}, \bibinfo{author}{Baevski, A.}, \bibinfo{author}{Conneau, A.}, \bibinfo{author}{Auli, M.}, \bibinfo{year}{2022}.
\newblock \bibinfo{title}{{XLS-R:} self-supervised cross-lingual speech representation learning at scale}, in: \bibinfo{editor}{Ko, H.}, \bibinfo{editor}{Hansen, J.H.L.} (Eds.), \bibinfo{booktitle}{Interspeech 2022, 23rd Annual Conference of the International Speech Communication Association, Incheon, Korea, 18-22 September 2022}, \bibinfo{publisher}{{ISCA}}. pp. \bibinfo{pages}{2278--2282}.
\newblock \URLprefix \url{https://doi.org/10.21437/Interspeech.2022-143}, \DOIprefix\doi{10.21437/INTERSPEECH.2022-143}.
\bibitem[{Baevski et~al.(2021)Baevski, Hsu, Conneau and Auli}]{Baevski2021-wav2vec-u}
\bibinfo{author}{Baevski, A.}, \bibinfo{author}{Hsu, W.}, \bibinfo{author}{Conneau, A.}, \bibinfo{author}{Auli, M.}, \bibinfo{year}{2021}.
\newblock \bibinfo{title}{Unsupervised speech recognition}, in: \bibinfo{editor}{Ranzato, M.}, \bibinfo{editor}{Beygelzimer, A.}, \bibinfo{editor}{Dauphin, Y.N.}, \bibinfo{editor}{Liang, P.}, \bibinfo{editor}{Vaughan, J.W.} (Eds.), \bibinfo{booktitle}{Advances in Neural Information Processing Systems 34: Annual Conference on Neural Information Processing Systems 2021, NeurIPS 2021, December 6-14, 2021, virtual}, pp. \bibinfo{pages}{27826--27839}.
\newblock \URLprefix \url{https://papers.nips.cc/paper_files/paper/2021/hash/ea159dc9788ffac311592613b7f71fbb-Abstract.html}.
\bibitem[{Baevski et~al.(2020)Baevski, Zhou, Mohamed and Auli}]{Baevski2020-wav2vec2}
\bibinfo{author}{Baevski, A.}, \bibinfo{author}{Zhou, Y.}, \bibinfo{author}{Mohamed, A.}, \bibinfo{author}{Auli, M.}, \bibinfo{year}{2020}.
\newblock \bibinfo{title}{wav2vec 2.0: {A} framework for self-supervised learning of speech representations}, in: \bibinfo{editor}{Larochelle, H.}, \bibinfo{editor}{Ranzato, M.}, \bibinfo{editor}{Hadsell, R.}, \bibinfo{editor}{Balcan, M.}, \bibinfo{editor}{Lin, H.} (Eds.), \bibinfo{booktitle}{Advances in Neural Information Processing Systems 33: Annual Conference on Neural Information Processing Systems 2020, NeurIPS 2020, December 6-12, 2020, virtual}, pp. \bibinfo{pages}{12449--12460}.
\newblock \URLprefix \url{https://proceedings.neurips.cc/paper/2020/hash/92d1e1eb1cd6f9fba3227870bb6d7f07-Abstract.html}.
\bibitem[{Bapna et~al.(2021)Bapna, Chung, Wu, Gulati, Jia, Clark, Johnson, Riesa, Conneau and Zhang}]{bapna2021-slam}
\bibinfo{author}{Bapna, A.}, \bibinfo{author}{Chung, Y.}, \bibinfo{author}{Wu, N.}, \bibinfo{author}{Gulati, A.}, \bibinfo{author}{Jia, Y.}, \bibinfo{author}{Clark, J.H.}, \bibinfo{author}{Johnson, M.}, \bibinfo{author}{Riesa, J.}, \bibinfo{author}{Conneau, A.}, \bibinfo{author}{Zhang, Y.}, \bibinfo{year}{2021}.
\newblock \bibinfo{title}{{SLAM:} {A} unified encoder for speech and language modeling via speech-text joint pre-training}.
\newblock \bibinfo{journal}{CoRR} \bibinfo{volume}{abs/2110.10329}.
\newblock \URLprefix \url{https://arxiv.org/abs/2110.10329}, \href{http://arxiv.org/abs/2110.10329}{{\tt arXiv:2110.10329}}.
\bibitem[{Bengio(2012)}]{Bengio2012-deeptransfer}
\bibinfo{author}{Bengio, Y.}, \bibinfo{year}{2012}.
\newblock \bibinfo{title}{Deep learning of representations for unsupervised and transfer learning}, in: \bibinfo{editor}{Guyon, I.}, \bibinfo{editor}{Dror, G.}, \bibinfo{editor}{Lemaire, V.}, \bibinfo{editor}{Taylor, G.W.}, \bibinfo{editor}{Silver, D.L.} (Eds.), \bibinfo{booktitle}{Unsupervised and Transfer Learning - Workshop held at {ICML} 2011, Bellevue, Washington, USA, July 2, 2011}, \bibinfo{publisher}{JMLR.org}. pp. \bibinfo{pages}{17--36}.
\newblock \URLprefix \url{http://proceedings.mlr.press/v27/bengio12a.html}, \DOIprefix\doi{10.5555/3045796.3045800}.
\bibitem[{Bhati et~al.(2022)Bhati, Villalba, Zelasko, Moro{-}Vel{\'{a}}zquez and Dehak}]{Bhati2022-scpc}
\bibinfo{author}{Bhati, S.}, \bibinfo{author}{Villalba, J.}, \bibinfo{author}{Zelasko, P.}, \bibinfo{author}{Moro{-}Vel{\'{a}}zquez, L.}, \bibinfo{author}{Dehak, N.}, \bibinfo{year}{2022}.
\newblock \bibinfo{title}{Unsupervised speech segmentation and variable rate representation learning using segmental contrastive predictive coding}.
\newblock \bibinfo{journal}{{IEEE} {ACM} Trans. Audio Speech Lang. Process.} \bibinfo{volume}{30}, \bibinfo{pages}{2002--2014}.
\newblock \URLprefix \url{https://doi.org/10.1109/TASLP.2022.3180684}, \DOIprefix\doi{10.1109/TASLP.2022.3180684}.
\bibitem[{Chen et~al.(2019)Chen, Tsai, Liu, Lee and Lee}]{Chen2019}
\bibinfo{author}{Chen, K.}, \bibinfo{author}{Tsai, C.}, \bibinfo{author}{Liu, D.}, \bibinfo{author}{Lee, H.}, \bibinfo{author}{Lee, L.}, \bibinfo{year}{2019}.
\newblock \bibinfo{title}{Completely unsupervised phoneme recognition by a generative adversarial network harmonized with iteratively refined hidden markov models}, in: \bibinfo{editor}{Kubin, G.}, \bibinfo{editor}{Kacic, Z.} (Eds.), \bibinfo{booktitle}{Interspeech 2019, 20th Annual Conference of the International Speech Communication Association, Graz, Austria, 15-19 September 2019}, \bibinfo{publisher}{{ISCA}}. pp. \bibinfo{pages}{1856--1860}.
\newblock \URLprefix \url{https://doi.org/10.21437/Interspeech.2019-2068}, \DOIprefix\doi{10.21437/INTERSPEECH.2019-2068}.
\bibitem[{Chen et~al.(2022a)Chen, Wang, Chen, Wu, Liu, Chen, Li, Kanda, Yoshioka, Xiao, Wu, Zhou, Ren, Qian, Qian, Wu, Zeng, Yu and Wei}]{chen2022-wavlm}
\bibinfo{author}{Chen, S.}, \bibinfo{author}{Wang, C.}, \bibinfo{author}{Chen, Z.}, \bibinfo{author}{Wu, Y.}, \bibinfo{author}{Liu, S.}, \bibinfo{author}{Chen, Z.}, \bibinfo{author}{Li, J.}, \bibinfo{author}{Kanda, N.}, \bibinfo{author}{Yoshioka, T.}, \bibinfo{author}{Xiao, X.}, \bibinfo{author}{Wu, J.}, \bibinfo{author}{Zhou, L.}, \bibinfo{author}{Ren, S.}, \bibinfo{author}{Qian, Y.}, \bibinfo{author}{Qian, Y.}, \bibinfo{author}{Wu, J.}, \bibinfo{author}{Zeng, M.}, \bibinfo{author}{Yu, X.}, \bibinfo{author}{Wei, F.}, \bibinfo{year}{2022}a.
\newblock \bibinfo{title}{{WavLM}: Large-scale self-supervised pre-training for full stack speech processing}.
\newblock \bibinfo{journal}{{IEEE} J. Sel. Top. Signal Process.} \bibinfo{volume}{16}, \bibinfo{pages}{1505--1518}.
\newblock \URLprefix \url{https://doi.org/10.1109/JSTSP.2022.3188113}, \DOIprefix\doi{10.1109/JSTSP.2022.3188113}.
\bibitem[{Chen et~al.(2022b)Chen, Zhang, Rosenberg, Ramabhadran, Moreno, Bapna and Zen}]{chen2022-maestro}
\bibinfo{author}{Chen, Z.}, \bibinfo{author}{Zhang, Y.}, \bibinfo{author}{Rosenberg, A.}, \bibinfo{author}{Ramabhadran, B.}, \bibinfo{author}{Moreno, P.J.}, \bibinfo{author}{Bapna, A.}, \bibinfo{author}{Zen, H.}, \bibinfo{year}{2022}b.
\newblock \bibinfo{title}{{MAESTRO:} matched speech text representations through modality matching}, in: \bibinfo{editor}{Ko, H.}, \bibinfo{editor}{Hansen, J.H.L.} (Eds.), \bibinfo{booktitle}{Interspeech 2022, 23rd Annual Conference of the International Speech Communication Association, Incheon, Korea, 18-22 September 2022}, \bibinfo{publisher}{{ISCA}}. pp. \bibinfo{pages}{4093--4097}.
\newblock \URLprefix \url{https://doi.org/10.21437/Interspeech.2022-10937}, \DOIprefix\doi{10.21437/INTERSPEECH.2022-10937}.
\bibitem[{Conneau et~al.(2020)Conneau, Baevski, Collobert, Mohamed and Auli}]{conneau2020-xlsr53}
\bibinfo{author}{Conneau, A.}, \bibinfo{author}{Baevski, A.}, \bibinfo{author}{Collobert, R.}, \bibinfo{author}{Mohamed, A.}, \bibinfo{author}{Auli, M.}, \bibinfo{year}{2020}.
\newblock \bibinfo{title}{Unsupervised cross-lingual representation learning for speech recognition}.
\newblock \bibinfo{journal}{CoRR} \bibinfo{volume}{abs/2006.13979}.
\newblock \URLprefix \url{https://arxiv.org/abs/2006.13979}, \href{http://arxiv.org/abs/2006.13979}{{\tt arXiv:2006.13979}}.
\bibitem[{Cuervo et~al.(2022a)Cuervo, Grabias, Chorowski, Ciesielski, Lancucki, Rychlikowski and Marxer}]{Cuervo2022-macpc}
\bibinfo{author}{Cuervo, S.}, \bibinfo{author}{Grabias, M.}, \bibinfo{author}{Chorowski, J.}, \bibinfo{author}{Ciesielski, G.}, \bibinfo{author}{Lancucki, A.}, \bibinfo{author}{Rychlikowski, P.}, \bibinfo{author}{Marxer, R.}, \bibinfo{year}{2022}a.
\newblock \bibinfo{title}{Contrastive prediction strategies for unsupervised segmentation and categorization of phonemes and words}, in: \bibinfo{booktitle}{{IEEE} International Conference on Acoustics, Speech and Signal Processing, {ICASSP} 2022, Virtual and Singapore, 23-27 May 2022}, \bibinfo{publisher}{{IEEE}}. pp. \bibinfo{pages}{3189--3193}.
\newblock \URLprefix \url{https://doi.org/10.1109/ICASSP43922.2022.9746102}, \DOIprefix\doi{10.1109/ICASSP43922.2022.9746102}.
\bibitem[{Cuervo et~al.(2022b)Cuervo, Lancucki, Marxer, Rychlikowski and Chorowski}]{CuervoL2022-hcpc}
\bibinfo{author}{Cuervo, S.}, \bibinfo{author}{Lancucki, A.}, \bibinfo{author}{Marxer, R.}, \bibinfo{author}{Rychlikowski, P.}, \bibinfo{author}{Chorowski, J.}, \bibinfo{year}{2022}b.
\newblock \bibinfo{title}{Variable-rate hierarchical {CPC} leads to acoustic unit discovery in speech}, in: \bibinfo{editor}{Koyejo, S.}, \bibinfo{editor}{Mohamed, S.}, \bibinfo{editor}{Agarwal, A.}, \bibinfo{editor}{Belgrave, D.}, \bibinfo{editor}{Cho, K.}, \bibinfo{editor}{Oh, A.} (Eds.), \bibinfo{booktitle}{Advances in Neural Information Processing Systems 35: Annual Conference on Neural Information Processing Systems 2022, NeurIPS 2022, New Orleans, LA, USA, November 28 - December 9, 2022}.
\newblock \URLprefix \url{http://papers.nips.cc/paper_files/paper/2022/hash/e2b0a30ea6a67cba58134e57348afb91-Abstract-Conference.html}.
\bibitem[{Dunbar et~al.(2017)Dunbar, Cao, Benjumea, Karadayi, Bernard, Besacier, Anguera and Dupoux}]{Dunbar2017}
\bibinfo{author}{Dunbar, E.}, \bibinfo{author}{Cao, X.}, \bibinfo{author}{Benjumea, J.}, \bibinfo{author}{Karadayi, J.}, \bibinfo{author}{Bernard, M.}, \bibinfo{author}{Besacier, L.}, \bibinfo{author}{Anguera, X.}, \bibinfo{author}{Dupoux, E.}, \bibinfo{year}{2017}.
\newblock \bibinfo{title}{The zero resource speech challenge 2017}.
\newblock \bibinfo{journal}{CoRR} \bibinfo{volume}{abs/1712.04313}.
\newblock \URLprefix \url{http://arxiv.org/abs/1712.04313}.
\bibitem[{Fuchs and Hoshen(2023)}]{fuchs2023-gradseg}
\bibinfo{author}{Fuchs, T.S.}, \bibinfo{author}{Hoshen, Y.}, \bibinfo{year}{2023}.
\newblock \bibinfo{title}{Unsupervised word segmentation using temporal gradient pseudo-labels}, in: \bibinfo{booktitle}{{IEEE} International Conference on Acoustics, Speech and Signal Processing {ICASSP} 2023, Rhodes Island, Greece, June 4-10, 2023}, \bibinfo{publisher}{{IEEE}}. pp. \bibinfo{pages}{1--5}.
\newblock \URLprefix \url{https://doi.org/10.1109/ICASSP49357.2023.10095363}, \DOIprefix\doi{10.1109/ICASSP49357.2023.10095363}.
\bibitem[{Gao et~al.(2023)Gao, Shi, Chuang, Garc{\'{\i}}a, Lee, Watanabe and Khudanpur}]{Gao2023-euro}
\bibinfo{author}{Gao, D.}, \bibinfo{author}{Shi, J.}, \bibinfo{author}{Chuang, S.}, \bibinfo{author}{Garc{\'{\i}}a, L.P.}, \bibinfo{author}{Lee, H.}, \bibinfo{author}{Watanabe, S.}, \bibinfo{author}{Khudanpur, S.}, \bibinfo{year}{2023}.
\newblock \bibinfo{title}{Euro: {ESPnet} unsupervised {ASR} open-source toolkit}, in: \bibinfo{booktitle}{{IEEE} International Conference on Acoustics, Speech and Signal Processing {ICASSP} 2023, Rhodes Island, Greece, June 4-10, 2023}, \bibinfo{publisher}{{IEEE}}. pp. \bibinfo{pages}{1--5}.
\newblock \URLprefix \url{https://doi.org/10.1109/ICASSP49357.2023.10096977}, \DOIprefix\doi{10.1109/ICASSP49357.2023.10096977}.
\bibitem[{Graves et~al.(2006)Graves, Fern{\'{a}}ndez, Gomez and Schmidhuber}]{Graves06-CTC}
\bibinfo{author}{Graves, A.}, \bibinfo{author}{Fern{\'{a}}ndez, S.}, \bibinfo{author}{Gomez, F.J.}, \bibinfo{author}{Schmidhuber, J.}, \bibinfo{year}{2006}.
\newblock \bibinfo{title}{Connectionist temporal classification: labelling unsegmented sequence data with recurrent neural networks}, in: \bibinfo{booktitle}{ICML}, pp. \bibinfo{pages}{369--376}.
\newblock \URLprefix \url{https://doi.org/10.1145/1143844.1143891}, \DOIprefix\doi{10.1145/1143844.1143891}.
\bibitem[{Hasegawa{-}Johnson et~al.(2020)Hasegawa{-}Johnson, Rolston, Goudeseune, Levow and Kirchhoff}]{MHJ-languagenet-g2p}
\bibinfo{author}{Hasegawa{-}Johnson, M.}, \bibinfo{author}{Rolston, L.}, \bibinfo{author}{Goudeseune, C.}, \bibinfo{author}{Levow, G.}, \bibinfo{author}{Kirchhoff, K.}, \bibinfo{year}{2020}.
\newblock \bibinfo{title}{Grapheme-to-phoneme transduction for cross-language {ASR}}, in: \bibinfo{editor}{Anke, L.E.}, \bibinfo{editor}{Mart{\'{\i}}n{-}Vide, C.}, \bibinfo{editor}{Spasic, I.} (Eds.), \bibinfo{booktitle}{Statistical Language and Speech Processing - 8th International Conference, {SLSP} 2020, Cardiff, UK, October 14-16, 2020, Proceedings}, \bibinfo{publisher}{Springer}. pp. \bibinfo{pages}{3--19}.
\newblock \URLprefix \url{https://doi.org/10.1007/978-3-030-59430-5_1}, \DOIprefix\doi{10.1007/978-3-030-59430-5_1}.
\bibitem[{Hsu et~al.(2021)Hsu, Bolte, Tsai, Lakhotia, Salakhutdinov and Mohamed}]{Hsu2022-hubert}
\bibinfo{author}{Hsu, W.}, \bibinfo{author}{Bolte, B.}, \bibinfo{author}{Tsai, Y.H.}, \bibinfo{author}{Lakhotia, K.}, \bibinfo{author}{Salakhutdinov, R.}, \bibinfo{author}{Mohamed, A.}, \bibinfo{year}{2021}.
\newblock \bibinfo{title}{{HuBERT}: Self-supervised speech representation learning by masked prediction of hidden units}.
\newblock \bibinfo{journal}{{IEEE} {ACM} Trans. Audio Speech Lang. Process.} \bibinfo{volume}{29}, \bibinfo{pages}{3451--3460}.
\newblock \URLprefix \url{https://doi.org/10.1109/TASLP.2021.3122291}, \DOIprefix\doi{10.1109/TASLP.2021.3122291}.
\bibitem[{Kamper(2023)}]{Kamper2023-wordsegdp}
\bibinfo{author}{Kamper, H.}, \bibinfo{year}{2023}.
\newblock \bibinfo{title}{Word segmentation on discovered phone units with dynamic programming and self-supervised scoring}.
\newblock \bibinfo{journal}{{IEEE} {ACM} Trans. Audio Speech Lang. Process.} \bibinfo{volume}{31}, \bibinfo{pages}{684--694}.
\newblock \URLprefix \url{https://doi.org/10.1109/TASLP.2022.3229264}, \DOIprefix\doi{10.1109/TASLP.2022.3229264}.
\bibitem[{Lample et~al.(2018a)Lample, Conneau, Denoyer and Ranzato}]{Lample2018-unsupmtmono}
\bibinfo{author}{Lample, G.}, \bibinfo{author}{Conneau, A.}, \bibinfo{author}{Denoyer, L.}, \bibinfo{author}{Ranzato, M.}, \bibinfo{year}{2018}a.
\newblock \bibinfo{title}{Unsupervised machine translation using monolingual corpora only}, in: \bibinfo{booktitle}{6th International Conference on Learning Representations, {ICLR} 2018, Vancouver, BC, Canada, April 30 - May 3, 2018, Conference Track Proceedings}, \bibinfo{publisher}{OpenReview.net}.
\newblock \URLprefix \url{https://openreview.net/forum?id=rkYTTf-AZ}.
\bibitem[{Lample et~al.(2018b)Lample, Conneau, Ranzato, Denoyer and J{\'{e}}gou}]{Lample2018-word-trans-non-parallel}
\bibinfo{author}{Lample, G.}, \bibinfo{author}{Conneau, A.}, \bibinfo{author}{Ranzato, M.}, \bibinfo{author}{Denoyer, L.}, \bibinfo{author}{J{\'{e}}gou, H.}, \bibinfo{year}{2018}b.
\newblock \bibinfo{title}{Word translation without parallel data}, in: \bibinfo{booktitle}{6th International Conference on Learning Representations, {ICLR} 2018, Vancouver, BC, Canada, April 30 - May 3, 2018, Conference Track Proceedings}, \bibinfo{publisher}{OpenReview.net}.
\newblock \URLprefix \url{https://openreview.net/forum?id=H196sainb}.
\bibitem[{Lewis et~al.(2020)Lewis, Liu, Goyal, Ghazvininejad, Mohamed, Levy, Stoyanov and Zettlemoyer}]{lewis-etal-2020-bart}
\bibinfo{author}{Lewis, M.}, \bibinfo{author}{Liu, Y.}, \bibinfo{author}{Goyal, N.}, \bibinfo{author}{Ghazvininejad, M.}, \bibinfo{author}{Mohamed, A.}, \bibinfo{author}{Levy, O.}, \bibinfo{author}{Stoyanov, V.}, \bibinfo{author}{Zettlemoyer, L.}, \bibinfo{year}{2020}.
\newblock \bibinfo{title}{{BART:} denoising sequence-to-sequence pre-training for natural language generation, translation, and comprehension}, in: \bibinfo{editor}{Jurafsky, D.}, \bibinfo{editor}{Chai, J.}, \bibinfo{editor}{Schluter, N.}, \bibinfo{editor}{Tetreault, J.R.} (Eds.), \bibinfo{booktitle}{Proceedings of the 58th Annual Meeting of the Association for Computational Linguistics, {ACL} 2020, Online, July 5-10, 2020}, \bibinfo{publisher}{Association for Computational Linguistics}. pp. \bibinfo{pages}{7871--7880}.
\newblock \URLprefix \url{https://doi.org/10.18653/v1/2020.acl-main.703}, \DOIprefix\doi{10.18653/V1/2020.ACL-MAIN.703}.
\bibitem[{Liu et~al.(2022)Liu, Hsu, Auli and Baevski}]{Liu2022-wav2vecu2.0}
\bibinfo{author}{Liu, A.H.}, \bibinfo{author}{Hsu, W.}, \bibinfo{author}{Auli, M.}, \bibinfo{author}{Baevski, A.}, \bibinfo{year}{2022}.
\newblock \bibinfo{title}{Towards end-to-end unsupervised speech recognition}, in: \bibinfo{booktitle}{{IEEE} Spoken Language Technology Workshop, {SLT} 2022, Doha, Qatar, January 9-12, 2023}, \bibinfo{publisher}{{IEEE}}. pp. \bibinfo{pages}{221--228}.
\newblock \URLprefix \url{https://doi.org/10.1109/SLT54892.2023.10023187}, \DOIprefix\doi{10.1109/SLT54892.2023.10023187}.
\bibitem[{Liu et~al.(2018)Liu, Chen, Lee and Lee}]{Liu18-mapping-relation-gan}
\bibinfo{author}{Liu, D.}, \bibinfo{author}{Chen, K.}, \bibinfo{author}{Lee, H.}, \bibinfo{author}{Lee, L.}, \bibinfo{year}{2018}.
\newblock \bibinfo{title}{Completely unsupervised phoneme recognition by adversarially learning mapping relationships from audio embeddings}, in: \bibinfo{editor}{Yegnanarayana, B.} (Ed.), \bibinfo{booktitle}{Interspeech 2018, 19th Annual Conference of the International Speech Communication Association, Hyderabad, India, 2-6 September 2018}, \bibinfo{publisher}{{ISCA}}. pp. \bibinfo{pages}{3748--3752}.
\newblock \URLprefix \url{https://doi.org/10.21437/Interspeech.2018-1800}, \DOIprefix\doi{10.21437/INTERSPEECH.2018-1800}.
\bibitem[{Ni et~al.(2022)Ni, Wang, Gao, Qian, Zhang, Chang and Hasegawa{-}Johnson}]{Ni-unsuptts-interspeech2022}
\bibinfo{author}{Ni, J.}, \bibinfo{author}{Wang, L.}, \bibinfo{author}{Gao, H.}, \bibinfo{author}{Qian, K.}, \bibinfo{author}{Zhang, Y.}, \bibinfo{author}{Chang, S.}, \bibinfo{author}{Hasegawa{-}Johnson, M.}, \bibinfo{year}{2022}.
\newblock \bibinfo{title}{Unsupervised text-to-speech synthesis by unsupervised automatic speech recognition}, in: \bibinfo{editor}{Ko, H.}, \bibinfo{editor}{Hansen, J.H.L.} (Eds.), \bibinfo{booktitle}{Interspeech 2022, 23rd Annual Conference of the International Speech Communication Association, Incheon, Korea, 18-22 September 2022}, \bibinfo{publisher}{{ISCA}}. pp. \bibinfo{pages}{461--465}.
\newblock \URLprefix \url{https://doi.org/10.21437/Interspeech.2022-816}, \DOIprefix\doi{10.21437/INTERSPEECH.2022-816}.
\bibitem[{van~den Oord et~al.(2018)van~den Oord, Li and Vinyals}]{vandenoord2018-cpc}
\bibinfo{author}{van~den Oord, A.}, \bibinfo{author}{Li, Y.}, \bibinfo{author}{Vinyals, O.}, \bibinfo{year}{2018}.
\newblock \bibinfo{title}{Representation learning with contrastive predictive coding}.
\newblock \bibinfo{journal}{CoRR} \bibinfo{volume}{abs/1807.03748}.
\newblock \URLprefix \url{http://arxiv.org/abs/1807.03748}, \href{http://arxiv.org/abs/1807.03748}{{\tt arXiv:1807.03748}}.
\bibitem[{Panayotov et~al.(2015)Panayotov, Chen, Povey and Khudanpur}]{Panayotov15-LibriSpeech}
\bibinfo{author}{Panayotov, V.}, \bibinfo{author}{Chen, G.}, \bibinfo{author}{Povey, D.}, \bibinfo{author}{Khudanpur, S.}, \bibinfo{year}{2015}.
\newblock \bibinfo{title}{Librispeech: An {ASR} corpus based on public domain audio books}, in: \bibinfo{booktitle}{ICASSP}, pp. \bibinfo{pages}{5206--5210}.
\newblock \URLprefix \url{https://doi.org/10.1109/ICASSP.2015.7178964}, \DOIprefix\doi{10.1109/ICASSP.2015.7178964}.
\bibitem[{Pasad et~al.(2024)Pasad, Chien, Settle and Livescu}]{Pasad2024-what-do-ssl-speech-words}
\bibinfo{author}{Pasad, A.}, \bibinfo{author}{Chien, C.}, \bibinfo{author}{Settle, S.}, \bibinfo{author}{Livescu, K.}, \bibinfo{year}{2024}.
\newblock \bibinfo{title}{What do self-supervised speech models know about words?}
\newblock \bibinfo{journal}{Trans. Assoc. Comput. Linguistics} \bibinfo{volume}{12}, \bibinfo{pages}{372--391}.
\newblock \URLprefix \url{https://doi.org/10.1162/tacl_a_00656}, \DOIprefix\doi{10.1162/TACL_A_00656}.
\bibitem[{Pasad et~al.(2023)Pasad, Shi and Livescu}]{pasad2022-comp-layer-analysis}
\bibinfo{author}{Pasad, A.}, \bibinfo{author}{Shi, B.}, \bibinfo{author}{Livescu, K.}, \bibinfo{year}{2023}.
\newblock \bibinfo{title}{Comparative layer-wise analysis of self-supervised speech models}, in: \bibinfo{booktitle}{{IEEE} International Conference on Acoustics, Speech and Signal Processing {ICASSP} 2023, Rhodes Island, Greece, June 4-10, 2023}, \bibinfo{publisher}{{IEEE}}. pp. \bibinfo{pages}{1--5}.
\newblock \URLprefix \url{https://doi.org/10.1109/ICASSP49357.2023.10096149}, \DOIprefix\doi{10.1109/ICASSP49357.2023.10096149}.
\bibitem[{Peng and Harwath(2022)}]{peng2022-vghubert}
\bibinfo{author}{Peng, P.}, \bibinfo{author}{Harwath, D.}, \bibinfo{year}{2022}.
\newblock \bibinfo{title}{Word discovery in visually grounded, self-supervised speech models}, in: \bibinfo{editor}{Ko, H.}, \bibinfo{editor}{Hansen, J.H.L.} (Eds.), \bibinfo{booktitle}{Interspeech 2022, 23rd Annual Conference of the International Speech Communication Association, Incheon, Korea, 18-22 September 2022}, \bibinfo{publisher}{{ISCA}}. pp. \bibinfo{pages}{2823--2827}.
\newblock \URLprefix \url{https://doi.org/10.21437/Interspeech.2022-10652}, \DOIprefix\doi{10.21437/INTERSPEECH.2022-10652}.
\bibitem[{Polyak et~al.(2021)Polyak, Adi, Copet, Kharitonov, Lakhotia, Hsu, Mohamed and Dupoux}]{polyak21-speech-resynthesis}
\bibinfo{author}{Polyak, A.}, \bibinfo{author}{Adi, Y.}, \bibinfo{author}{Copet, J.}, \bibinfo{author}{Kharitonov, E.}, \bibinfo{author}{Lakhotia, K.}, \bibinfo{author}{Hsu, W.}, \bibinfo{author}{Mohamed, A.}, \bibinfo{author}{Dupoux, E.}, \bibinfo{year}{2021}.
\newblock \bibinfo{title}{Speech resynthesis from discrete disentangled self-supervised representations}, in: \bibinfo{editor}{Hermansky, H.}, \bibinfo{editor}{Cernock{\'{y}}, H.}, \bibinfo{editor}{Burget, L.}, \bibinfo{editor}{Lamel, L.}, \bibinfo{editor}{Scharenborg, O.}, \bibinfo{editor}{Motl{\'{\i}}cek, P.} (Eds.), \bibinfo{booktitle}{Interspeech 2021, 22nd Annual Conference of the International Speech Communication Association, Brno, Czechia, 30 August - 3 September 2021}, \bibinfo{publisher}{{ISCA}}. pp. \bibinfo{pages}{3615--3619}.
\newblock \URLprefix \url{https://doi.org/10.21437/Interspeech.2021-475}, \DOIprefix\doi{10.21437/INTERSPEECH.2021-475}.
\bibitem[{Tseng et~al.(2024)Tseng, Hu, Chiang, Tseng, Lee, Lee and Sun}]{Tseng2024-reborn}
\bibinfo{author}{Tseng, L.}, \bibinfo{author}{Hu, E.}, \bibinfo{author}{Chiang, D.C.}, \bibinfo{author}{Tseng, Y.}, \bibinfo{author}{Lee, H.}, \bibinfo{author}{Lee, L.}, \bibinfo{author}{Sun, S.}, \bibinfo{year}{2024}.
\newblock \bibinfo{title}{{REBORN:} reinforcement-learned boundary segmentation with iterative training for unsupervised {ASR}}.
\newblock \bibinfo{journal}{CoRR} \bibinfo{volume}{abs/2402.03988}.
\newblock \URLprefix \url{https://doi.org/10.48550/arXiv.2402.03988}, \DOIprefix\doi{10.48550/ARXIV.2402.03988}, \href{http://arxiv.org/abs/2402.03988}{{\tt arXiv:2402.03988}}.
\bibitem[{Wang et~al.(2024)Wang, Hasegawa{-}Johnson and Yoo}]{wang2024unsupervised}
\bibinfo{author}{Wang, L.}, \bibinfo{author}{Hasegawa{-}Johnson, M.}, \bibinfo{author}{Yoo, C.D.}, \bibinfo{year}{2024}.
\newblock \bibinfo{title}{Unsupervised speech recognition with n-skipgram and positional unigram matching}, in: \bibinfo{booktitle}{{IEEE} International Conference on Acoustics, Speech and Signal Processing, {ICASSP} 2024, Seoul, Republic of Korea, April 14-19, 2024}, \bibinfo{publisher}{{IEEE}}. pp. \bibinfo{pages}{10936--10940}.
\newblock \URLprefix \url{https://doi.org/10.1109/ICASSP48485.2024.10446327}, \DOIprefix\doi{10.1109/ICASSP48485.2024.10446327}.
\bibitem[{Wang et~al.(2023)Wang, Ni, Gao, Li, Chang, Fan, Wu, Hasegawa{-}Johnson and Yoo}]{wang-etal-2023-unsup-speech2sign}
\bibinfo{author}{Wang, L.}, \bibinfo{author}{Ni, J.}, \bibinfo{author}{Gao, H.}, \bibinfo{author}{Li, J.}, \bibinfo{author}{Chang, K.C.}, \bibinfo{author}{Fan, X.}, \bibinfo{author}{Wu, J.}, \bibinfo{author}{Hasegawa{-}Johnson, M.}, \bibinfo{author}{Yoo, C.D.}, \bibinfo{year}{2023}.
\newblock \bibinfo{title}{Listen, decipher and sign: Toward unsupervised speech-to-sign language recognition}, in: \bibinfo{editor}{Rogers, A.}, \bibinfo{editor}{Boyd{-}Graber, J.L.}, \bibinfo{editor}{Okazaki, N.} (Eds.), \bibinfo{booktitle}{Findings of the Association for Computational Linguistics: {ACL} 2023, Toronto, Canada, July 9-14, 2023}, \bibinfo{publisher}{Association for Computational Linguistics}. pp. \bibinfo{pages}{6785--6800}.
\newblock \URLprefix \url{https://doi.org/10.18653/v1/2023.findings-acl.424}, \DOIprefix\doi{10.18653/V1/2023.FINDINGS-ACL.424}.
\bibitem[{Williams(1992)}]{williams1992-reinforce}
\bibinfo{author}{Williams, R.J.}, \bibinfo{year}{1992}.
\newblock \bibinfo{title}{Simple statistical gradient-following algorithms for connectionist reinforcement learning}.
\newblock \bibinfo{journal}{Mach. Learn.} \bibinfo{volume}{8}, \bibinfo{pages}{229--256}.
\newblock \URLprefix \url{https://doi.org/10.1007/BF00992696}, \DOIprefix\doi{10.1007/BF00992696}.
\bibitem[{Yeh et~al.(2019)Yeh, Chen, Yu and Yu}]{Yeh2019-unsup-asr}
\bibinfo{author}{Yeh, C.}, \bibinfo{author}{Chen, J.}, \bibinfo{author}{Yu, C.}, \bibinfo{author}{Yu, D.}, \bibinfo{year}{2019}.
\newblock \bibinfo{title}{Unsupervised speech recognition via segmental empirical output distribution matching}, in: \bibinfo{booktitle}{7th International Conference on Learning Representations, {ICLR} 2019, New Orleans, LA, USA, May 6-9, 2019}, \bibinfo{publisher}{OpenReview.net}.
\newblock \URLprefix \url{https://openreview.net/forum?id=Bylmkh05KX}.
\bibitem[{Zhang et~al.(2024)Zhang, Chen, Zhou, Wu, Ren, Liu, Yao, Gong, Dai, Li and Wei}]{zhang2024-speechlm}
\bibinfo{author}{Zhang, Z.}, \bibinfo{author}{Chen, S.}, \bibinfo{author}{Zhou, L.}, \bibinfo{author}{Wu, Y.}, \bibinfo{author}{Ren, S.}, \bibinfo{author}{Liu, S.}, \bibinfo{author}{Yao, Z.}, \bibinfo{author}{Gong, X.}, \bibinfo{author}{Dai, L.}, \bibinfo{author}{Li, J.}, \bibinfo{author}{Wei, F.}, \bibinfo{year}{2024}.
\newblock \bibinfo{title}{{SpeechLM}: Enhanced speech pre-training with unpaired textual data}.
\newblock \bibinfo{journal}{{IEEE} {ACM} Trans. Audio Speech Lang. Process.} \bibinfo{volume}{32}, \bibinfo{pages}{2177--2187}.
\newblock \URLprefix \url{https://doi.org/10.1109/TASLP.2024.3379877}, \DOIprefix\doi{10.1109/TASLP.2024.3379877}.
\bibitem[{Zhang et~al.(2022)Zhang, Zhou, Ao, Liu, Dai, Li and Wei}]{zhang2022-speechut}
\bibinfo{author}{Zhang, Z.}, \bibinfo{author}{Zhou, L.}, \bibinfo{author}{Ao, J.}, \bibinfo{author}{Liu, S.}, \bibinfo{author}{Dai, L.}, \bibinfo{author}{Li, J.}, \bibinfo{author}{Wei, F.}, \bibinfo{year}{2022}.
\newblock \bibinfo{title}{{SpeechUT}: Bridging speech and text with hidden-unit for encoder-decoder based speech-text pre-training}, in: \bibinfo{editor}{Goldberg, Y.}, \bibinfo{editor}{Kozareva, Z.}, \bibinfo{editor}{Zhang, Y.} (Eds.), \bibinfo{booktitle}{Proceedings of the 2022 Conference on Empirical Methods in Natural Language Processing, {EMNLP} 2022, Abu Dhabi, United Arab Emirates, December 7-11, 2022}, \bibinfo{publisher}{Association for Computational Linguistics}. pp. \bibinfo{pages}{1663--1676}.
\newblock \URLprefix \url{https://doi.org/10.18653/v1/2022.emnlp-main.108}, \DOIprefix\doi{10.18653/V1/2022.EMNLP-MAIN.108}.

\end{thebibliography}






\end{document}